\documentclass{article}

\usepackage[preprint]{neurips_2026}

\usepackage[utf8]{inputenc} 
\usepackage[T1]{fontenc}    
\usepackage{hyperref}       
\usepackage{url}            
\usepackage{booktabs}       
\usepackage{amsfonts}       
\usepackage{nicefrac}       
\usepackage{microtype}      
\usepackage{xcolor}         
\usepackage{xspace}
\usepackage{amsmath}
\usepackage{graphicx}
\usepackage{enumitem} 
\usepackage{tabularx}
\usepackage{amssymb}
\usepackage{etoc}
\usepackage{titletoc} 
\usepackage{booktabs}   
\usepackage{multirow}   
\usepackage{array}      
\usepackage{tabularx}
\usepackage{booktabs}
\newcolumntype{Y}{>{\centering\arraybackslash}X} 
\usepackage{subcaption}
\usepackage{xcolor}
\definecolor{affective}{HTML}{BEDBFF}
\definecolor{memory}{HTML}{FFA1AD}
\definecolor{intent}{HTML}{7BF1A8}
\usepackage{pgf-pie}
\usepackage{tikz}

\definecolor{mPink}{HTML}{FFB7B2}
\definecolor{mBlue}{HTML}{B2CEFE}
\definecolor{mGreen}{HTML}{B2F2BB}
\definecolor{mYellow}{HTML}{FFFFB2}
\definecolor{mPurple}{HTML}{E2B2FE}
\definecolor{mOrange}{HTML}{FFDAC1}
\definecolor{mMint}{HTML}{B2EBE0}
\definecolor{mLavender}{HTML}{D1D1F5}
\definecolor{mRose}{HTML}{FFC4D6}
\definecolor{mPeach}{HTML}{FFCCB6}
\definecolor{mLemon}{HTML}{FFF5BA}
\definecolor{mSky}{HTML}{A7DBF2}
\definecolor{mSage}{HTML}{C9E4DE}
\definecolor{mGrey}{HTML}{E0E0E0}

\newcommand{\name}{EgoIntrospect\xspace}
\title{\name: An Egocentric Dataset and Benchmark for User-Centric Internal State Reasoning}

\author{%
Zeyu Wang$^{1*}$, Chang Liu$^{1,8*}$, Eduardus Tjitrahardja$^{1}$, Yuntao Wang$^{1\ddagger}$, 
Borislav Pavlov$^{1}$, \\ \bfseries Fangfei Gou$^{1}$, Jose Manuel Davila$^{1}$, Dai Shi$^{2}$, Ran Xu$^{1}$, Yue Pan$^{3}$, Jiayi Tan$^{3}$, Shuting Chang$^{2}$, \\ \bfseries Qi Wang$^{2}$, Jinzhao Li$^{1}$, Jiacheng Hua$^{1}$, 
Yifei Huang$^{4}$, 
Jingwei Sun$^{5}$, Yu Zhang$^{5}$, Liuxin Zhang$^{5}$,\\ \bfseries  
Guocai Yao$^{6}$, Jia Jia$^{1}$, Yin Li$^{7}$, Qianying Wang$^{5}$, Yuanchun Shi$^{1}$, Miao Liu$^{1\ddagger}$ \\
$^{1}$Tsinghua University, $^{2}$Tongji University \\
$^{3}$Renmin University of China, $^{4}$The University of Tokyo \\
$^{5}$Lenovo Group, $^{6}$Peking University \\
$^{7}$University of Wisconsin--Madison,$^{8}$Shanghai Qi Zhi Institute \\
}
\begin{document}

\maketitle
\begin{abstract}
Despite extensive efforts on egocentric video datasets and benchmarks, understanding users' internal states, crucial for enabling seamless AI assistant experiences, remains largely overlooked. 
In this work, we introduce \textbf{\name}, the first egocentric dataset captured in user-driven scenarios with self-annotations that explicitly reveal users' interactive intentions with AI assistants. 
\name was collected using a cross-device setup, providing synchronized video, audio, gaze, motion, physiological signals. It consists of 180 hours of recordings from 60 subjects, with an average clip duration of 3h.
Leveraging \name, we formalize a suite of tasks centered on user internal states, including \emph{affective experience}, \emph{interactive intent}, and \emph{cognitive memory}. 
We further process the annotations to construct benchmarks that evaluate the ability of modern multimodal large language models (MLLMs) to reason about users’ internal states from egocentric observations. 
Experiments on our benchmark suggest that existing MLLMs struggle to effectively leverage multimodal signals to infer users’ subjective internal states.
The dataset and annotations will be made publicly available to advance research in egocentric vision and wearable AI assistants. Project page: \url{https://ego-introspect.github.io/}
\end{abstract}

\section{Introduction}

\begin{figure*}
  \centering
  \includegraphics[width=0.98\textwidth]{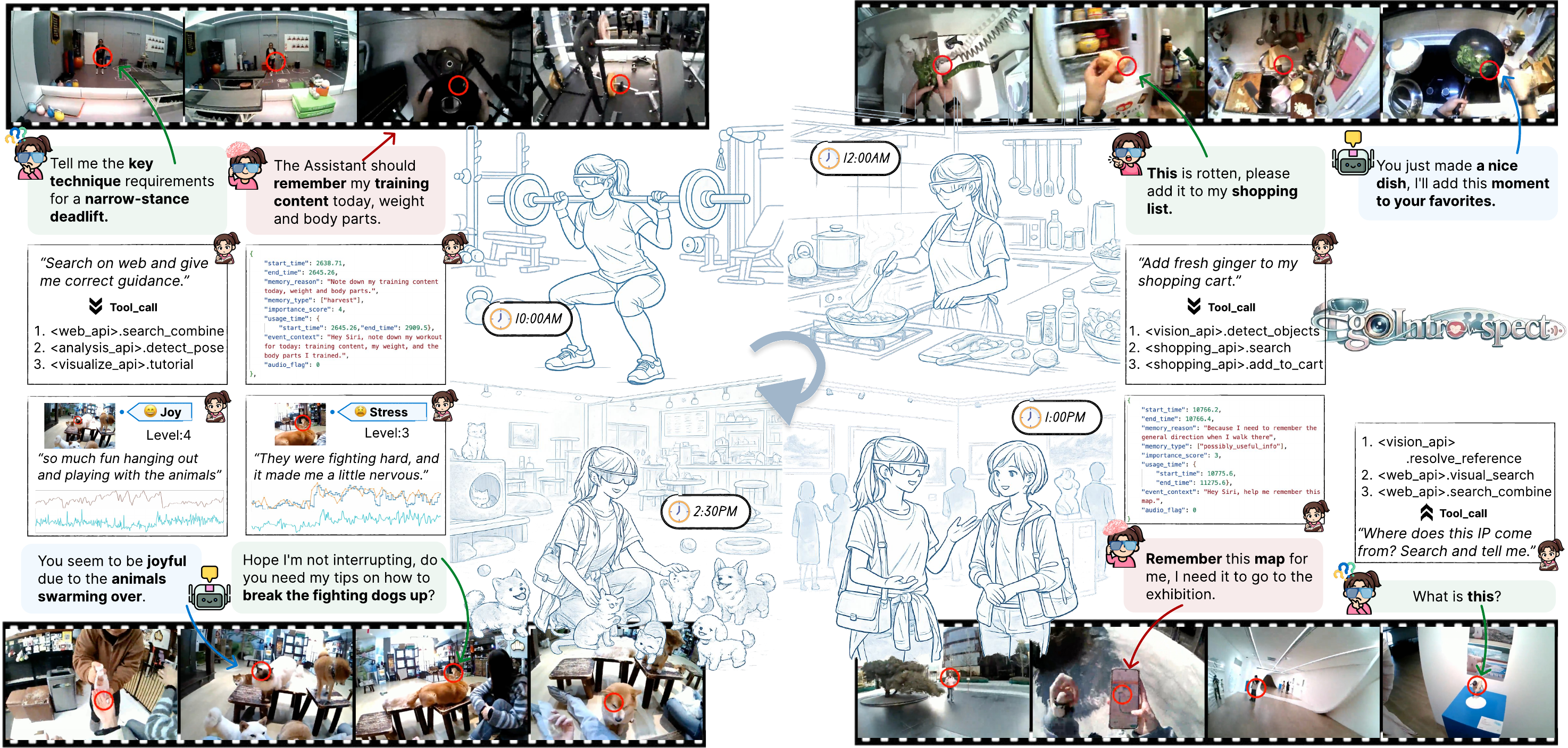}
  \vspace{-0.5em}
  \caption{\emph{Visual examples of \name for understanding user internal states}. We illustrate the daily usage of smart-glass AI assistants through three core dimensions:
(1) \textcolor{affective}{Affective Experience}: recognizing salient moments and inferring the user's emotional states (e.g., joy or stress);
(2) \textcolor{intent}{Request Intent}: tackling complex, context-grounded queries and initiating proactive assistance (e.g., gym guidance or shopping list management); and
(3) \textcolor{memory}{Cognitive Memory}: modeling memory reachability and the practical need for short-/long-term memory assistance (e.g., navigation guidance or training schedule).
Our dataset synchronizes egocentric video, audio, gaze, and physiological signals, bridging first-person observations with the user’s internal cognitive processes.}
\vspace{-1.5em}
  \label{fig:teaser}
\end{figure*}

Recent advances in wearable computing~\cite{meta_rayban_smart_glasses_2023,engel2023project} and multimodal large language models (MLLMs)~\cite{Yin_2024,liu2023llava,achiam2023gpt,comanici2025gemini} have enabled a new generation of intelligent, context-aware virtual assistants, capable of operating over multimodal egocentric data and responding to various user initiated queries. Prior works on egocentric MLLMs~\cite{lin2022egocentric,wang2023lifelongmemory,khirodkar2023ego,ye2024mm,xue2023learning,Zhang_2023_ICCV,Xu_2024_CVPR} have been largely task-driven, focusing on perception and reasoning about the \textit{external} observations, including understanding objects, scenes, and the camera wearer’s interactions with their environment (e.g., actions).
However, as illustrated in Fig.~\ref{fig:teaser}, real-world interactions with AI assistants during daily routines often extend beyond these capabilities. Many user queries are instead centered on \textit{internal} states, such as affective status, request intent grounded in visual context, and the memorability of first-person visual experiences.


We argue that the egocentric AI assistants should adopt a user-driven design, requiring models to infer user-centric internal states from continuous inputs. Concretely, affective awareness enables \textit{personalized, context-sensitive responses}, request intent prediction enables \textit{proactive assistance beyond reactive query answering}, and memorability estimation supports \textit{selective recall for long video understanding}. Together, these capabilities constitute a critical yet missing component of user-centric AI assistants that can adapt, anticipate, and collaborate with users in everyday settings.

However, the progress towards this goal is bottlenecked by data and benchmark. While recent work has presented large-scale datasets and benchmarks for objects, actions and scene understanding in egocentric vision~\cite{grauman2022ego4d,Grauman_2024_CVPR,damen2018scaling,yang2025egolife}, and a small body of work has begun to explore affective experience in this setting~\cite{lin20243}, prior work remain limited in two key aspects. \textit{First}, they typically focus on a narrow slice of internal state (e.g., emotion), without capturing the broader spectrum of user-centric cognition. \textit{Second}, they predominantly rely on video and audio, overlooking physiological signals that provide complementary and often more direct measurements of a user's internal processes.

To bridge the gaps, we introduce \textbf{\name}, \textit{the first multimodal egocentric dataset and benchmark designed for understanding user internal states in a naturalistic, daily-life setting}. \name is collected using a minimally invasive, wearable platform including everyday smart glasses~\cite{baumann2023neon}, Apple Watch, a physiological watch~\cite{raloway_s2_nd} and smart ring~\cite{tang2025dataset}, capturing synchronized egocentric video, audio, gaze, motion, physiological and environmental signal. Our dataset comprises over 180 hours of recordings from 60 participants. The key property of \name is that all annotations are provided by users both in-situ and post-hoc, enabling more reliable capture of their affective states, request intent, and perceived memorability.


Building on this data, we establish a suite of benchmark tasks that systematically evaluate models' ability to infer and reason about user internal states from multi-modal egocentric data, including:
\vspace{-0.3em}
\begin{itemize}[nosep,leftmargin=10pt]
    \item \emph{Affective Experience}: probes whether models can identify salient moments users wish to record and infer their emotional status.
    \item \emph{Request Intent}: evaluates whether models can infer user intent by retrieving appropriate external knowledge for a request, and delivering proactive assistance with accurate timing and content.
    \item \emph{Cognitive Memory}: assesses whether models can estimate what users can vividly remember, as well as what requires memory assistance for future recall.
\end{itemize}
It is worth noting that several of the above tasks are inherently subjective, posing challenges for reliable evaluation. We tackle these challenges through carefully designed data collection and annotation protocols that mitigate noise, along with principled task formulations that encourage consistency and predictability. In the main paper, we elaborate data collection and annotation protocols, and benchmark design. Detailed experiment setup and results are deferred to the Appendix.


\section{Related Works}

\subsection{Egocentric Datasets and Benchmarks}
Recent years have seen a surge in egocentric datasets~\cite{damen2022rescaling,grauman2022ego4d,sigurdsson2018charades,jia2020lemma,Huang_2024_CVPR,Grauman_2024_CVPR,yang2025egolife,pan2023aria}. Ego4D and Ego-Exo4D~\cite{grauman2022ego4d,Grauman_2024_CVPR} encompass a vast array of activities. Recently, EgoLife~\cite{yang2025egolife} collects long egocentric daily life logs, but is limited to a small set of subjects and staged environments. Another related line of work collects physiological signals alongside egocentric data for emotion analysis in lab-controlled environments~\cite{jammot2025egoemotion,park2020k} or for energy expenditure monitoring~\cite{nakamura2017jointly}. Compared to prior efforts, \textit{\name} records multi-sensory long-form streams and adopts a free-form capture protocol defined by participants. And to address the inherent subjectivity, we follow established practices in human-computer interaction~\cite{doherty2018construal,lukka2024measuring} by adopting both in-situ triggering and retrospective self-annotation, mitigating known biases such as peak-end effects, and salience-driven recall.

Recently, a few egocentric VQA benchmarks have been proposed to evaluate models’ ability to reason about first-person behaviors~\cite{Manigrasso_2026_WACV,Datta_2022_CVPR,barmann2022did,jia2022egotaskqa,yuan2025eoc,Huang_2025_CVPR,benita2022assistq,chandrasegaran2024hourvideo,cheng2024egothink,chen2026egoplan,mangalam2023egoschema,peng2025eye,zhou2025egotextvqa}. These recent benchmarks are largely task-driven, where questions are generated from Ego4D or Ego-Exo4D to probe model capabilities in reasoning~\cite{mangalam2023egoschema}, grounding~\cite{Huang_2025_CVPR}, planning~\cite{peng2025eye}, and text understanding~\cite{zhou2025egotextvqa}. In contrast, our benchmark follows a user-driven paradigm, focusing on whether MLLMs can interpret users’ internal states for seamless AI interactions. 



A detailed comparison between EgoIntrospect and prior works is in the appendix (Tab.~\ref{tab:dataset_comparison} and Tab.~\ref{tab:benchmark_comparison}).

\subsection{Affective Experience, Request Intent, Cognitive Memory}

\textbf{Affective Experience}.\ A key challenge for revisiting daily experiences with smart glass~\cite{hodges2006sensecam, yao2016highlight} is how to align recorded moments with what is meaningful to the user~\cite{le2016impact}. Existing work suggests that such moments can be indicated by both users' emotional states~\cite{staahl2009experiencing, sas2013affectcam} and attentional behavior~\cite{chang2021memx, li2021eye}.  Related HCI research has also explored affect sensing using diverse signals, including facial expressions~\cite{masai2017evaluation, li2024eyeecho, kwon2021emotion}, eye images~\cite{wu2020emo}, and physiological signals~\cite{jammot2025egoemotion}. Recent works have begun to study emotion understanding in egocentric settings, mb by jointly considering first-person visual context and wearable signals~\cite{zhao2022smart, cheng2024emotion, hu2025emobench, jammot2025egoemotion}. Leveraging this line of work, our study specifically probes whether MLLMs can address tasks related to affective experiences.


\textbf{Request Intent}.\ Several HCI studies aim to resolve under-specified requests using natural multimodal context, including deictic language, gaze, pointing, and dialogue history~\cite{wang2024g, lee2024gazepointar, yan2024voila, peng2025eye, sun2025visual}. More proactive settings further reduce explicit input~\cite{lee2025sensible}: the system infers user's unspoken request at a cued moment, a problem related to visual question generation~\cite{mostafazadeh2016generating} and egocentric intent modeling~\cite{pan2026egointent, veerabadran2025benchmarking} but centered on user-specific needs. One step further, proactive recommendation requires systems to infer not only what assistance may be useful, but also when it can be delivered without disruption.~\cite{lee2025sensible, cai2025aiget, zhang2025proactive}. \textit{\name} supports under-specified request understanding, proactive request prediction, and request recommendation by collecting user-centric annotations.

%




\textbf{Memory}.\ Continuous recording of egocentric videos have long been used for lifelogging, offering an external memory pool to help users recall past experiences~\cite{hodges2006sensecam, gelonch2019acceptability, sellen2007life}. A central challenge lies in how to identify memorable moments~\cite{niforatos2017can, sellen2010beyond}. These efforts can facilitate AI assistants for specific applications, such as support for cognitive impairment~\cite{ait2019smart, li2019fmt}, way-finding~\cite{qiu2024navmarkar}, and other situated memory tasks. Recent wearable memory agents further integrate egocentric sensing with LLMs or AI models to retrieve, summarize, and proactively present memory cues in real time~\cite{zulfikar2024memoro, haddad2025ar, Datta_2022_CVPR, yuan2025eoc}. Instead of addressing researcher-designed memory tasks as in prior work, our study focuses on both what users are likely to remember themselves and the lifespan of memory retrieval.


\section{Data Collection and Annotation Protocol}
\begin{figure}[t] 
    \centering 
    \includegraphics[
  width=\textwidth,
  height=0.4\textheight,
  keepaspectratio,
  trim=0.5cm 5.1cm 1.5cm 4.75cm,
  clip
]{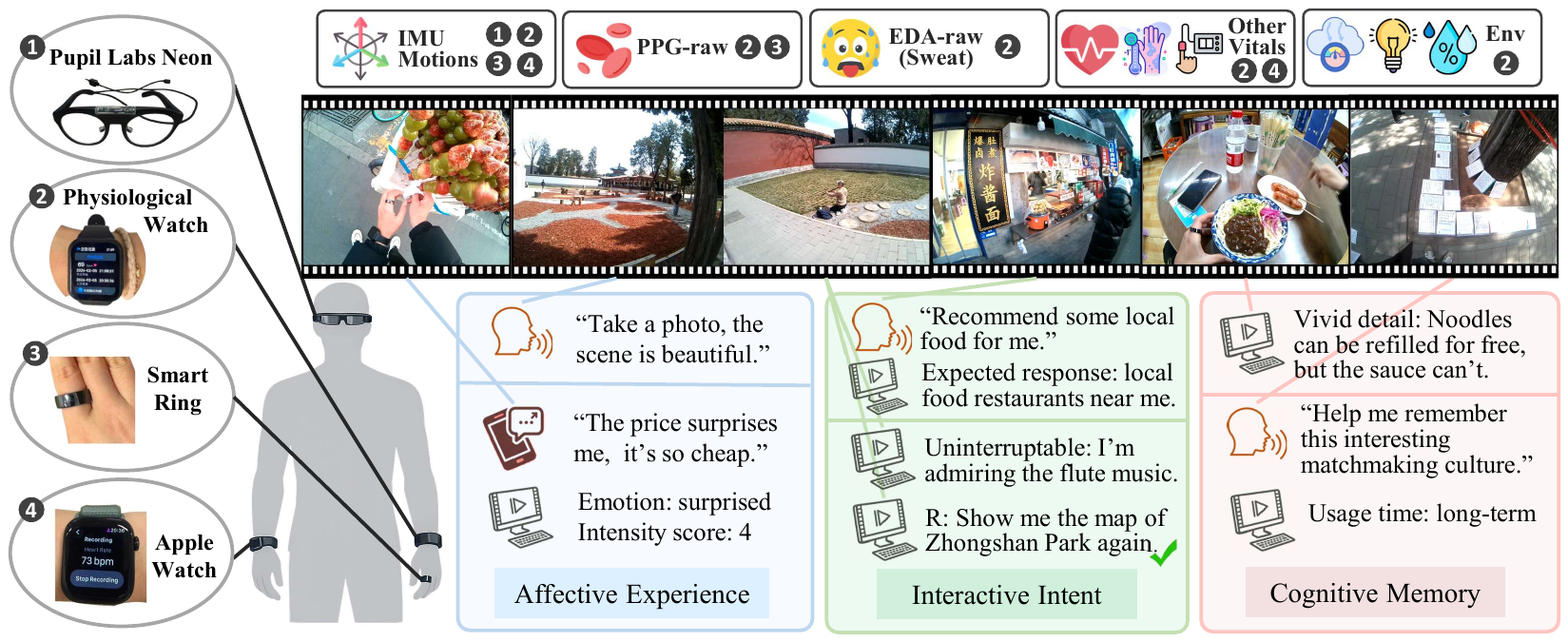}
\vspace{-1.5em}
   \caption{\emph{Capture and Annotation Overview of the \name}. Our dataset record users' natural daily routines with a multimodal wearable sensor suite. We synchronize exteroceptive (video, audio) and interoceptive (physiological, gaze, motion) signals for a rich multimodal capture. Annotations are obtained through two stages: an In-situ Labeling stage, where participants verbally mark key moments or send instant messages during recording, and a Retrospective Annotation stage, where participants subsequently review and refine annotations on our platform.}.
    \label{fig:settings} 
    \vspace{-2.5em}
\end{figure}

\subsection{Recording Setup, Participants and Ethics}
\textbf{Devices and data modalities}.\ As shown in Fig.~\ref{fig:settings}, our device setup combines egocentric smart glasses, a wristband, a smart watch, and a smart ring to capture synchronized multimodal signals across different body parts. The Pupil Labs Neon Glasses~\cite{baumann2023neon} record first-person video, eye tracking, audio, and IMU signals of head motion. The wristband records rich physiological and environmental signals. The Apple Watch captures daily activity and mobility statistics, while the smart ring provides fine-grained finger-centric physiological and motion sensing.

\textbf{Cross-device synchronization}.\ All other data streams were synchronized to the reference time of the smart glasses' video recording. At the start of each recording, participants performed three hand claps within the glasses' field of view. This procedure was repeated at the end of the recording. By visualizing the IMU data into waveforms, annotators identified the corresponding clap timestamps for all devices (Apple Watch, wristband, and ring). With the alignment process, we could retrieve synchronized sensor data through linear interpolation for given video segments. 

\textbf{Participants and Ethics}.\ We recruited N=60 participants through local campus and social-media. Participants received \$100 compensation for completing the study. The study was approved by the university’s Institutional Review Board (IRB). During consent, participants were informed about the types of data collected by the wearable sensors, the purpose of the study, the planned use of the collected data, and their right to withdraw.

\subsection{Experiment Pipeline}
\textbf{Pre-Recording}. We introduce the experiment procedure and confirms the activity schedule during recording with the participant. The introduction includes a briefing on the tasks requiring in-situ labeling but not the tasks that only require retrospective annotation.

\textbf{Data Collection}. At the beginning and the end of the recording, participants are required to clap hands for device synchronization. During recording, participants naturally follows their daily schedule, wearing our devices. Meanwhile, they need to intermittently complete in-situ labeling for our tasks. 

\textbf{Post-Processing and Annotation}. After the collection session, we process the data for annotation. We first export the video recording, compress and overlay gaze trace on the video. We then use ElevenLabs~\cite{elevenlabs_speech_to_text} or XunFei~\cite{iflytek_asr_llm_2026} to generate audio transcripts with word-level timestamp, and identify all verbal in-situ commands with GPT-4o~\cite{microsoft_foundry_gpt4o_2026}. Within the following week, participants return to the lab and finish retrospective annotations. For detailed procedure, please refer to Appendix ~\ref{app:experiment_pipeline}. 

\subsection{Annotation Protocol}

\noindent\textbf{Multi-stage Labeling Process}. Our pipeline (see Fig.~\ref{fig:settings}) combines \emph{in-situ labeling} with structured \emph{retrospective annotation} to capture both spontaneous intent and fine-grained semantic information. 

\begin{itemize}[nosep,leftmargin=10pt]
\item \emph{In-situ Labeling}. Participants provide real-time markers via two approaches to facilitate retrospective annotation:
(1) \textit{Voice-based commands}, which are spoken commands starting with \textit{``Hey Assistant''}, used for Task1.1, 2.1, and 3.2;
(2) \textit{Instant messages}, where participants sent a brief message to the experimenter whenever they noticed a salient emotional response (for Task1.2). 
\item \emph{Retrospective Annotation}. For tasks with in-situ labels, participants review the extracted message logs, then follow a standardized schema to fill-in pre-defined attributes. For tasks without \textit{in-situ} labels, participants review the whole recording to identify and annotate segments for each task. This multi-stage approach ensures that all subjective experiences are saved into structured data.
\end{itemize}

\textbf{Data Annotation Platform}.
Different from previous works, \name's annotation for each user-centered tasks are the participants themselves. We built a video annotation platform for retrospective annotation by modifying Vidat~\cite{zhang2020vidat}. The platform contains video playback and manipulation functions, as well as task-specific annotation UI. Additional details on the annotation interface and guidelines can be found in the Appendix ~\ref{App: annotation_platform}.

\section{Task Definition and Benchmark Design}
\subsection{Preliminaries}\label{sec:preliminary}

\textbf{Egocentric Context}.\ Let $v$ denotes a task-specific video clip, $s$ denote the synchronized physiological signal, $g$ denote user's attention signal, $a$ denote the audio track or transcription, depending on the target model's ability to understand dialogue. Our goal is not to decompose these tasks into task-specific agents, but to evaluate whether a general-purpose MLLM $f_{\theta}$ can infer user-centered states and intentions from egocentric context. 
Each benchmark question is represented as an instance $\mathcal{I} = \langle \mathcal{X}, q, \mathcal{C} \rangle$, where $\mathcal{X}$ denotes the available multi-modal context, $q$ is the natural-language question, and $\mathcal{C}$ is the candidate answer set. 
An MLLM receives a serialized multimodal prompt $\Phi(\mathcal{I})$ and predicts an answer: $\hat{y} = f_{\theta}(\Phi(\mathcal{I})).$ We also experiment with In-Context Learning (ICL) to help the model adapt to individual differences across participants.

\textbf{Target Space}.\ All tasks are formulated as selecting $y \subseteq \mathcal{C}$ from a candidate set $\mathcal{C} = \{c_i\}_{i=1}^k (k \geq 2)$, where $c_i$ represents text, label, object, or multimodal bundle. 
The ground-truth answer is denoted as $y \subseteq \mathcal{C}$, where $y$ can be single-answer or multi-answer depending on the task. The target space is carefully designed to ensure predictability for subjective tasks, as evidenced by the near-human performance reported in Appendices F--K.





\subsection{Affective Experience}


\begin{figure}[t] 
    \centering 
    \includegraphics[
  width=\textwidth,
  height=0.4\textheight,
  keepaspectratio,
  trim=0cm 3.7cm 1.8cm 4.3cm,
  clip
]{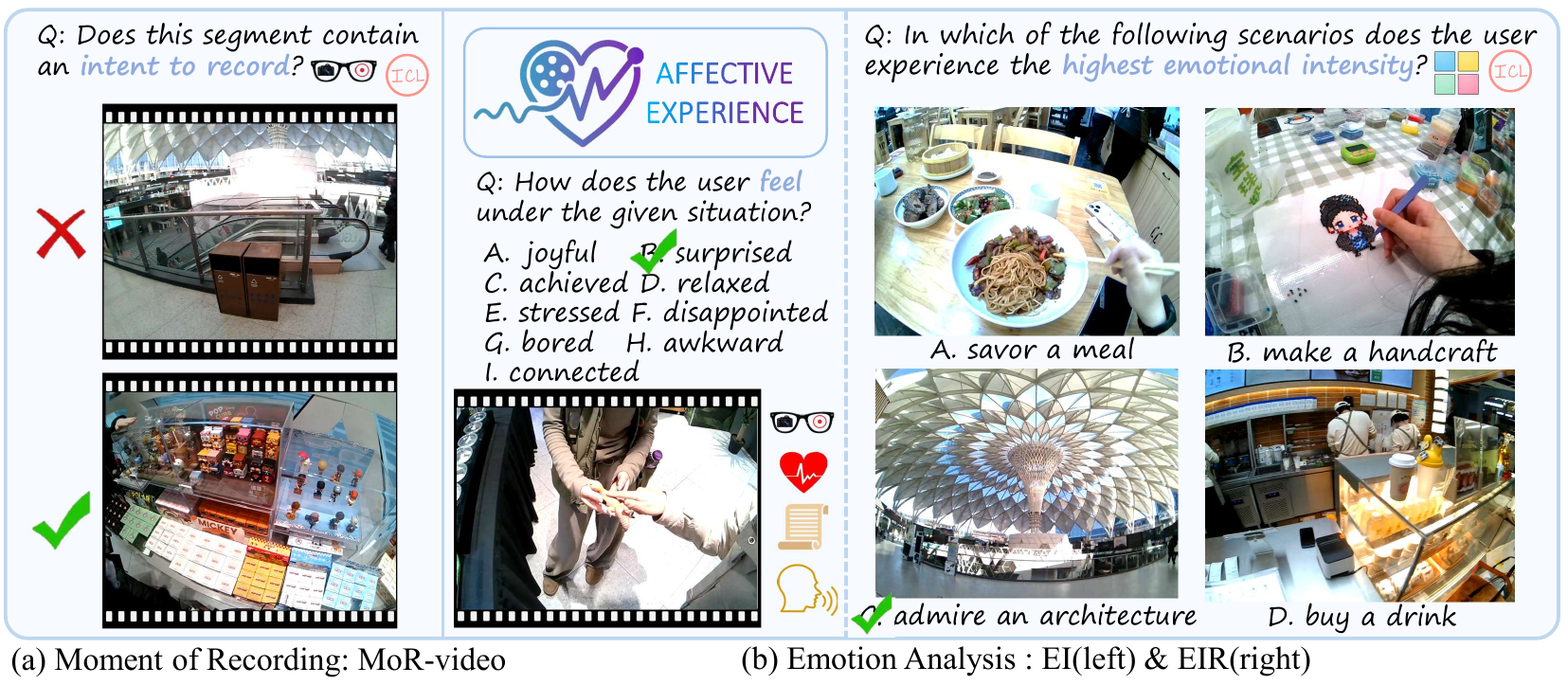}
\vspace{-2em}
    \caption{\emph{Illustrative examples of Affective Experience tasks}. (a) MoR-Video/-Photo determine whether a video segment or photo reflects the user’s capture intent. (b) Emotion Identification (EI) classifies the user’s affective state, and Emotion Intensity Recognition (EIR) selects the scenario with the highest emotional intensity. Icons indicate the specific experimental settings: (ICL) refers to the use of In-Context Learning examples, and the quad-pane icon ($\boxplus$) denotes multi-modal candidates.}
    \label{fig:task1} 
\vspace{-2em}
\end{figure}
\subsubsection{Moment of Recording}\label{sec:MoR}

\noindent\textbf{Motivation}.\ Capturing meaningful moments is one of the most prominent applications of AI glasses. However, existing systems require users to consciously identify interesting events and manually trigger recording, which inevitably interrupts ongoing interactions and implicitly places cognitive burden on the user. Moreover, what is considered meaningful varies significantly across individuals. Therefore, a future egocentric AI system should learn personalized notions of “interestingness” from historical examples and proactively identify such moments from continuous egocentric streams. 

\textbf{Data Annotation}. During data collection, participants marked desired capture moments by in-situ commands \textit{``Hey Assistant, take a photo/record a video''} and could optionally state the reason immediately. During retrospective annotation, they reviewed and annotated each command: for photo commands, they drew a bounding box $b_i^*$ around the intended capture target; for video commands, they marked the start and end time $[t_{s,i}^*, t_{e,i}^*]$ of the intended event segment $v_i^*$. 

\textbf{Benchmark Design}. We design two question types corresponding to the two command categories. A sample of this task is shown in Fig.~\ref{fig:task1}(a).

\begin{itemize}[nosep,leftmargin=10pt]
    \item \textit{MoR - photo.} We crop a video clip $v_i$ around the command timestamp and craft the candidate answer set to be $\mathcal{C}_i = \{c_i^+, c_{i,1}^-, c_{i,2}^-, c_{i,3}^-\}$, where each option is in form of a frame overlaid with bounding box, $c_i^+$ is the frame containing $b_i^*$ and the distractors $c_{i}^-$ are manually selected. Let the input be $\mathcal{X}_i=(v_i, a_i, g_i)$, the model output a single choice $\hat{y}_i = f_{\theta}(\Phi((v_i, a_i, g_i), q, \mathcal{C}_i)).$ 
    \item \textit{MoR - video.} We use the participant-annotated event segment $v_i^+$ as the positive segment, and randomly sample unlabeled segments from the same full recording as negative segments $v_i^-$.Let the input be $\mathcal{X}_i=(v_i, a_i, g_i)$, the model output a binary result $\hat{y}_i = f_{\theta}(\Phi((v_i, a_i, g_i), q)) \in \{0, 1\}.$ 
\end{itemize}

\textbf{Related to Existing Tasks}. Video highlight detection~\cite{Xiong_2019_CVPR,Badamdorj_2022_CVPR,Badamdorj_2021_ICCV,Liu_2022_CVPR}, which aims to identify salient moments from video sequences, has been extensively studied in computer vision. Compared to this task, our benchmark incorporates additional sensor modalities to capture personalized interesting moments, moving beyond generic saliency toward context-aware, user-specific understanding.

\subsubsection{Emotion Analysis}\label{sec:EA}
\noindent\textbf{Motivation}.\ Understanding a user’s emotional state is essential for delivering personalized and context-aware AI experiences. Our formulation requires models to perform situated reasoning over the environment, user context, and interaction history from first-person audiovisual signals augmented with rich physiological signals in naturalistic settings.

\textbf{Data Annotation}. During retrospective annotation, participants reviewed the full recording and marked segments associated with noticeable emotions. For each segment $v_i$, they provided an emotion label $e_i^*$, described the reason $e_i^{r,*}$ for the emotion, and rated its self-perceived intensity $e_i^{I,*}$ on a 1--5 scale. The emotion label can be selected from a predefined set or provided as a custom entry.

\textbf{Benchmark Design}.
We design two question types for this task. See a sample in Fig.~\ref{fig:task1}(b).

\begin{itemize}[nosep,leftmargin=10pt]
    \item \textit{Emotion identification (EI).} For each annotated emotion segment $v_i$, we construct a candidate answer set $\mathcal{C}_i=\{c_i^+, c_{i,1}^-, ... , c_{i,k}^-\}$, where each option $c$ consists of an emotion label and an associated reason. The correct option is $c_i^+=(e_i^*, e_i^{r,*})$, and the distractor options are generated by pairing other emotion labels with Gemini-generated reasons~\cite{geminiteam2026gemini31pro}. Let the input be $\mathcal{X}_i=(v_i, a_i, g_i, s_i)$, the model outputs a single choice $\hat{y}_i=f_{\theta}(\Phi(\mathcal{X}_i, q, \mathcal{C}_i))$.

    \item \textit{Emotion Intensity Recognition (EIR).} Instead of predicting the absolute intensity score, this task is formulated as within-user relative comparison. For the same participant and the same emotion label, we sample a set of annotated emotion segments $\{v_{i_1}, v_{i_2}, \dots, v_{i_k}\}$ with corresponding intensity scores $\{e_{i_1}^{I,*}, e_{i_2}^{I,*}, \dots, e_{i_k}^{I,*}\}$. The correct answer is the candidate with the highest or lowest participant-rated intensity. Each candidate in $\mathcal{C}_i$ is defined as $c_{i_j} = (v_{i_j}, a_{i_j}, g_{i_j}, s_{i_j})$. The model outputs a single choice $\hat{y}_i=f_{\theta}(\Phi(q, \mathcal{C}_i))$.
\end{itemize}

\noindent\textbf{Related to Existing Tasks}.\ Several recent efforts~\cite{cheng2024emotion,hu2025emobench,lian2025affectgptnewdatasetmodel, lin20243} have explored emotion analysis with MLLMs. They mainly evaluate whether models can reason emotions from curated multimodal samples, where emotional states are often inferred from visible expressions or speech. In contrast, our task focuses on the camera wearer's self-reported experience during continuous egocentric recording, where the user's emotion may not be directly observable from facial expressions.

\subsection{Interactive Intent}


\begin{figure}[t] 
    \centering 
    \includegraphics[
  width=\textwidth,
  height=0.4\textheight,
  keepaspectratio,
  trim=0 3.75cm 0 4.5cm,
  clip]{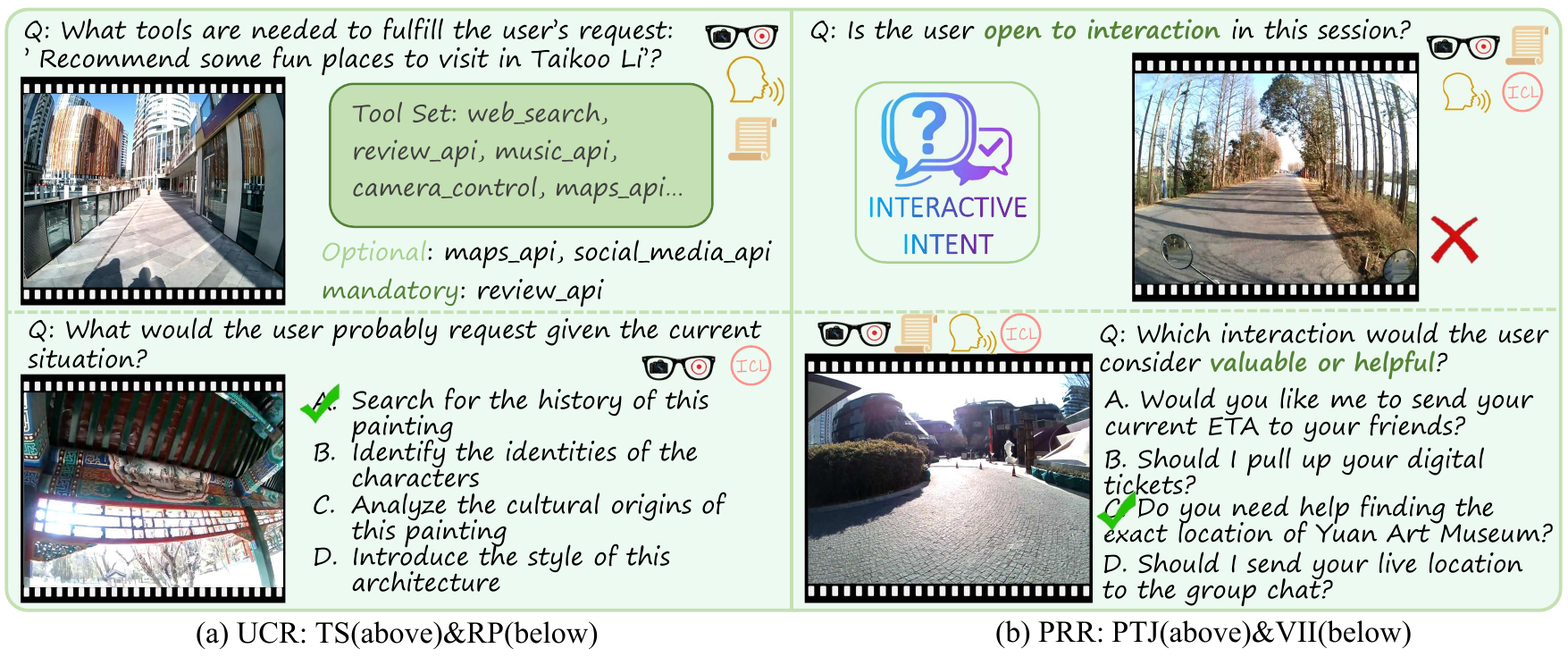}
\vspace{-2.5em}
    \caption{\emph{Illustrative examples of Interactive Intent tasks}. (a) Tool Selection (TS) identifies the tools needed for addressing the request, and Request Prediction (RP) predicts the user's possible request given the context. (b) Proactive Timing Judgment (PTJ) evaluates the user's openness to interaction, and Valuable Interaction Identification (VII) selects helpful proactive assistance from candidates.}
    \label{fig:task2} 
\vspace{-2em}
\end{figure}

\subsubsection{User-Initiated Contextual Request}\label{sec:tool-augmented planning}

\noindent\textbf{Motivation}.\ Despite the broad knowledge learned during pretraining, many real-world user requests still require external tools to be fully addressed. To act as an effective assistant, an MLLM must determine what external assistance is needed in a given situation and plan coherent multi-step tool usage to effectively resolve user requests.

\textbf{Data Annotation}. During data collection, participants marked user-initiated requests through in-situ voice commands, using an distinguishable identifier followed by a question or instruction. During retrospective annotation, participants reviewed each request segment $v_i$ and annotated the original request $u_i^*$ and the expected response requirement $\rho_i^*$. 

\textbf{Benchmark Design}. We design two question types for this task, with a sample shown in Fig.~\ref{fig:task2}(a).

\begin{itemize}[nosep,leftmargin=10pt]
    \item \textit{Tool Selection (TS)}. We evaluate whether the model can identify what external capability is needed to fulfill the user’s request. Given a set of available tools $\mathcal{T}=\{\tau_1,\tau_2,\dots,\tau_k\}$, we construct the candidate answer set as $\mathcal{C}_i=\mathcal{T}\cup\{\varnothing\}$, where $\varnothing$ denotes that no external tool is required. The ground-truth tool choice is derived by jointly considering $u_i^*$ and $\rho_i^*$. Let the input be $\mathcal{X}_i=(v_i,a_i,g_i)$, the model outputs one or more choices $\hat{y}_i=f_{\theta}(\Phi(\mathcal{X}_i,q,\mathcal{C}_i))$.
    
    \item \textit{Request Prediction (RP)}. For each request instance, we reformulate request $u_i^*$ and expected response $\rho_i^*$ into a more complete request $\tilde{u}_i^*$, which serves as the correct answer. We construct a four-option candidate set $\mathcal{C}_i=\{\tilde{u}_i^*, u_{i,1}^-, u_{i,2}^-, u_{i,3}^-\}$, where the distractor requests are generated by Gemini. Let the input be $\mathcal{X}_i=(v_i,\tilde{a}_i,g_i)$, where $\tilde{a}_i$ denotes the audio track or transcription with the spoken request removed, the model outputs a single choice $\hat{y}_i=f_{\theta}(\Phi(\mathcal{X}_i,q,\mathcal{C}_i))$.
\end{itemize}


\textbf{Related to Existing Tasks}. Existing agent tool-use benchmarks~\cite{qin2023toolllmfacilitatinglargelanguage,NEURIPS2024_e4c61f57} mostly evaluate tool execution or final outcome correctness, whereas our \textit{Tool Selection} question focuses on whether the model can identify the external capabilities needed for a real user request. \textit{Request Prediction} is related to visual question generation~\cite{mostafazadeh2016generating}, but instead of generating any plausible question from visual content, the model must recover the participant's actual request from captured signals.




\subsubsection{Proactive Request Recommendations}\label{sec:Proactive request rec}

\noindent\textbf{Motivation}.\ Unlike reactive assistants that require explicit user requests, proactive assistance enables the system to anticipate user needs based on ongoing context and past behavior, providing seamless and autonomous support. This is particularly important in egocentric streaming settings, where continuous observation allows the model to infer intent and intervene in a timely manner.

\textbf{Data Annotation}.\ Participants marked forbidden segments where they would not want to receive proactive recommendations, leaving remaining parts $\{v_i^{int}\}$ of as interruptible regions. For these segments, we generated recommendations at a five-minute granularity (detailed in Appendix~\ref{App:Proactive Recommendation Generation Pipeline for Task2.2}). Participants reviewed the generated recommendations and selected those they would be willing to accept. For each accepted recommendation $r_{i,j}^+$, they further annotated a presentation segment $v_{i,j}^{rec}$, indicating the time period in which they considered it appropriate for this recommendation to appear.

\textbf{Benchmark Design.}
We design two question types for this task. Samples are presented in Fig.~\ref{fig:task2}(b).

\begin{itemize}[nosep,leftmargin=10pt]
    \item \textit{Proactive Timing Judgment (PTJ)}. We sample positive segments from the interruptible parts and negative segments from the forbidden parts of the same recording. For each sampled segment $v_i$, let the input be $\mathcal{X}_i=(v_i,a_i,g_i)$ and binary candidate answer set be $\mathcal{C}=\{\textit{True},\textit{False}\}$, the model outputs a binary choice $\hat{y}_i=f_{\theta}(\Phi(\mathcal{X}_i,q,\mathcal{C}))$.
    \item \textit{Valuable Interaction Identification (VII)}. For each accepted recommendation $r_{i,j}^+$, we construct a four-option candidate set $\mathcal{C}_{i,j}=\{r_{i,j}^+, r_{i,j,1}^-, r_{i,j,2}^-, r_{i,j,3}^-\}$, where the distractors are randomly sampled from recommendations rejected by the participants. Let the input be $\mathcal{X}_{i,j}=(v_{i,j}^{rec},a_{i,j},g_{i,j})$, the model outputs a single choice $\hat{y}_{i,j}=f_{\theta}(\Phi(\mathcal{X}_{i,j},q,\mathcal{C}_{i,j}))$.
\end{itemize}


\textbf{Related to Existing Tasks}.\ Several prior works study proactive recommendation with LLMs \cite{zhao2024recommender,gao2023chat}, but primarily focus on modeling user preferred topics rather than interaction intent. Proactive interaction understanding with MLLMs remains largely underexplored. Existing work on proactive timing typically aims to identify key visual evidence for VQA~\cite{wang2025streambridge,zhang2025eyes,qian2025dispider,zhang2025proactive}, whereas our task focuses on modeling internal user states to determine whether an intervention would be disruptive. Moreover, these prior efforts emphasize responding to given queries at the appropriate time, rather than proactively recommending potential interaction options as required in our benchmark.


\subsection{Cognitive Memory}


\subsubsection{Memory Recall Prediction}\label{sec:Memory Reachability}

\textbf{Motivation}.\ Human memory is inherently selective and prone to decay, creating a functional gap between exhaustive digital logs and subjective recall. While existing systems focus on objective visual search, they fail to account for why specific vivid details persist while entire events become obscured. An AI model that understands human memory mechanisms could construct a hierarchical personal timeline that not only supports cognitively aligned retrieval but also enhances memory recall.

\textbf{Data Annotation}.\ Post-recording, participants provide retrospective accounts of vivid details they can recall from the session. Based on these, experimenters annotate timestamps for positive instances, while negative distractors are generated by MLLMs from the same video segments.

\begin{figure}[t] 
    \centering 
    \includegraphics[
  width=\textwidth,
  height=0.4\textheight,
  keepaspectratio,
  trim=0 4.0cm 0 3.8cm,
  clip]{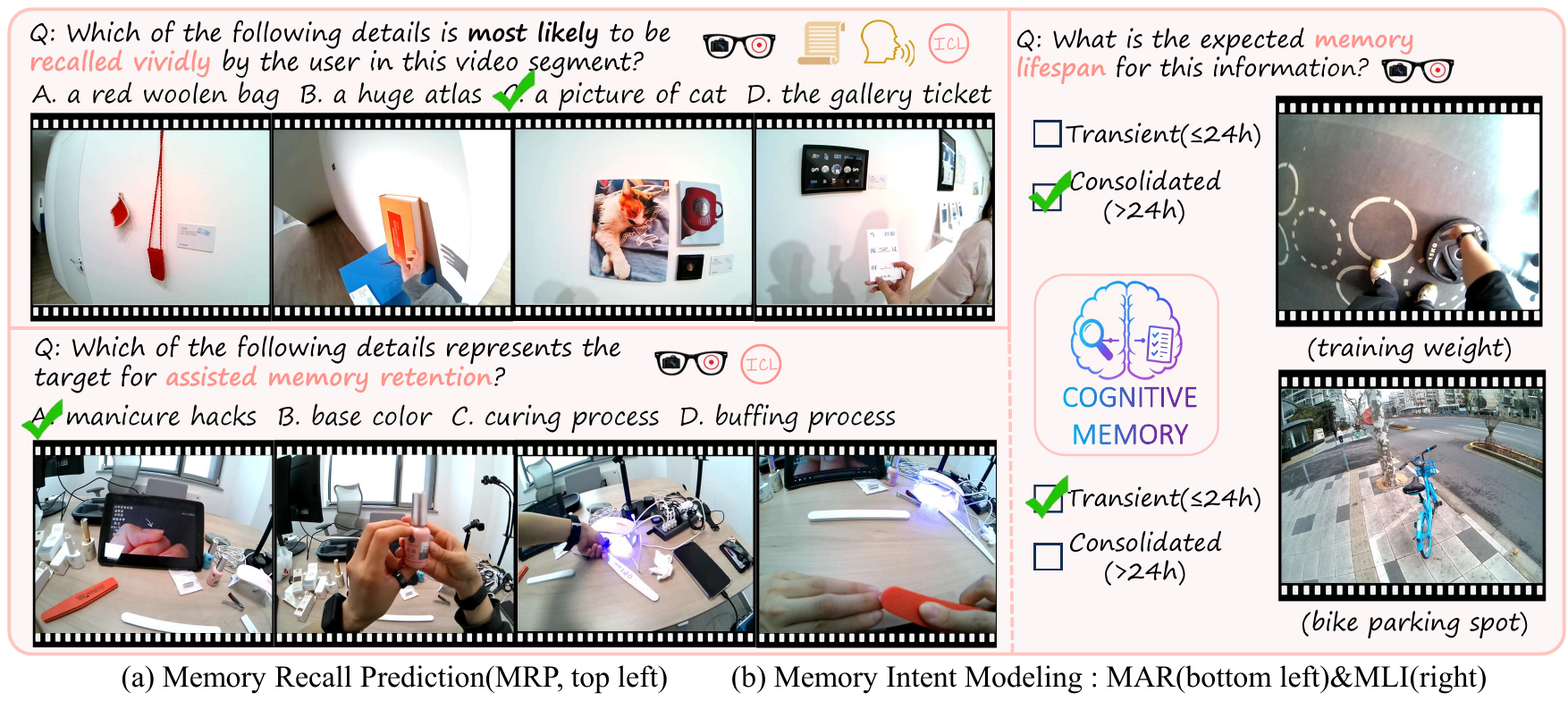}
\vspace{-2em}
    \caption{\emph{Illustrative examples of Cognitive Memory tasks}. (a) Memory Recall Prediction evaluates the model's ability to identify specific, vivid details that are most reachable in a user's memory. (b) Memory Assistance Recognition (MAR) identifies specific items the user explicitly intended to preserve (MAR-Event is shown; see MAR-Object in text), and Memory Lifespan Identification predicts the temporal utility and retention requirements of a designated memory item.}
    \label{fig:task3} 
\vspace{-2em}
\end{figure}


\noindent \textbf{Benchmark design}. This task assesses individual memory reachability by requiring the model to identify a specific, vivid detail that most likely to be recalled by the user of a past event from a set of plausible distractors. Given the multimodal input $\mathcal{X}_i = (v_i,a_i g_i)$, a query $q$, and a candidate set of details $\mathcal{C}_i$, the model must select the correct ground-truth option. The prediction is formulated as: $\hat{y}_i = f_{\theta}(\Phi(\mathcal{X}_i, q, \mathcal{C}_i)) \in \mathcal{C}_i$ . Examples of this task are illustrated in Fig.~\ref{fig:task3}(a).


\noindent\textbf{Related to Existing Tasks}.\ Early work in egocentric vision has explored the discovery of important objects and people for egocentric video summarization~\cite{lee2012discovering,lu2013story}, yet they do not directly address memorability. While estimating the memorability of images~\cite{isola2011makes} and videos~\cite{dumont2023modular} has been previously studied, prior work focus on third person visual content rather than first-person video. In addition, how MLLMs interpret human memory mechanisms has not been explored in existing benchmarks.


\subsubsection{Memory Intent Modeling}\label{sec:Memory Intent Modeling}

\noindent\textbf{Motivation}.\ While the aforementioned memory recall task addresses what users can vividly remember, an AI model must also consider what users want it to memorize. Notably, in continuous egocentric streaming, captured data comprises both passively observed perceptual inputs and salient content that warrants intentional preservation. Developing models that understand user memory needs is therefore critical for reducing cognitive load and transforming passive recording into accurate and effective memory assistance in complex real-world scenarios.

\textbf{Data Annotation}.\ During recording, participants identify specific recall targets—including events and objects—and assign binary lifespan labels during retrospective sessions. For MAR, $v_i$ contains user-annotated memory intent, with negative distractors generated by MLLMs to ensure competitive difficulty. For MLI, $v_i$ consists of the exact information segment to be memorized, with ground truth labels derived directly from participant annotations.

\noindent\textbf{Benchmark Design}. 
We characterize memory intent through two dimensions: the identification of items to be retained and the temporal relevance. Samples of these tasks are illustrated in Fig.~\ref{fig:task3}(b).

\noindent(1) \textbf{Memory Assistance Recognition (MAR)}: This task identifies items that a user explicitly intends to preserve for future use. Based on the memory item type, we define the task with two input formats:
\begin{itemize}[nosep,leftmargin=10pt]
\item \textit{MAR - Object}.\ For memory items with bounding box annotation, the candidate answer set $\mathcal{C}_i = \{c_i^+, c_{i,1}^-, c_{i,2}^-, c_{i,3}^-\}$ consists of frames overlaid with bounding boxes, where $c_i^+$ contains $b_i^*$ and the distractors $c_{i}^-$ are manually selected. Given the input $\mathcal{X}_i=(v_i, g_i)$, the model predicts a single choice: $\hat{y}_i = f_{\theta}(\Phi((v_i, g_i), q, \mathcal{C}_i)).$
\item \textit{MAR - Event}.\ For memory items annotated as events, we crop a video clip $v_i$ around the specific timestamps. We then craft textual distractors based on other events occurring within the same segment. With input $\mathcal{X}_i=(v_i, g_i)$, the model outputs a single choice: $\hat{y}_i = f_{\theta}(\Phi((v_i, g_i), q, \mathcal{C}_i)).$
\end{itemize}

\noindent (2) \textbf{Memory Lifespan Identification (MLI)}: This task assesses the model's ability to predict the temporal utility of specific information the user intends to remember. Inspired by the cognitive process of memory consolidation, the model must discern whether a designated memory item represents a transient need or warrants long-term retention. Given the multimodal input $\mathcal{X}_i = (v_i, g_i)$ and a query $q$ regarding a pre-identified memory item, the model predicts a binary classification $\hat{y}_i = f_{\theta}(\Phi(\mathcal{X}_i, q)) \in \{\mathcal{S}, \mathcal{L}\}$, where:
\begin{itemize}[nosep,leftmargin=10pt]
\item $\mathcal{S}$ (\textbf{within\_24h}): the memory item is relevant only within a short-term, 24-hour window;
\item $\mathcal{L}$ (\textbf{more\_than\_24h}): the memory item requires consolidation for retrieval beyond a single day.
\end{itemize}


\noindent\textbf{Related to Existing Tasks}.\ Episodic memory \cite{barmann2022did,grauman2022ego4d,Datta_2022_CVPR,wang2023lifelongmemory,Manigrasso_2026_WACV} has been widely studied in egocentric vision. However, prior work typically focuses on evaluating grounding ability by asking models when and where events occur using annotator-defined queries. In contrast, our benchmark adopts a user-centric formulation centered on what users would want to remember for future retrieval.


\subsection{Summary of Benchmark Experiments Setup}
We sampled 15 participants to form the test set and tested 21 SOTA VLMs on our benchmark. We conducted ablation study regarding data modalities, validate in-context learning effects and present human-level results on sampled test set. Detailed results are provided in Appendices F--K. 


\section{Conclusion}
In this paper, we introduce EgoIntrospect, the first dataset and benchmark aimed at advancing personalized egocentric AI assistants. Compared to prior datasets, EgoIntrospect adopts unconstrained capture settings and self-annotation to more reliably capture internal user states. Therefore, our proposed benchmark suite provides a snapshot of how users intend to use wearable AI assistants, highlighting the gaps in current MLLMs. Our work takes an important step toward bridging prevailing MLLMs with AI assistant for wearable devices, and points to exciting future research directions for personalized and proactive AI user experiences.



\bibliographystyle{unsrt} 
\bibliography{references}  

\newpage
\appendix 

\section*{\centering \Large Appendix} 
\addcontentsline{toc}{section}{Appendix}

\startcontents[appendices]
\printcontents[appendices]{}{1}{%
    \section*{Table of Contents} 
    \hrule 
}
\vspace{0.5cm}
\hrule 
\newpage

\section{Annotation Platform}
\label{App: annotation_platform}
\subsection{Task1.1 - Moment of Recording}
\begin{figure}[t]
    \centering
    \includegraphics[width=0.6\textwidth]{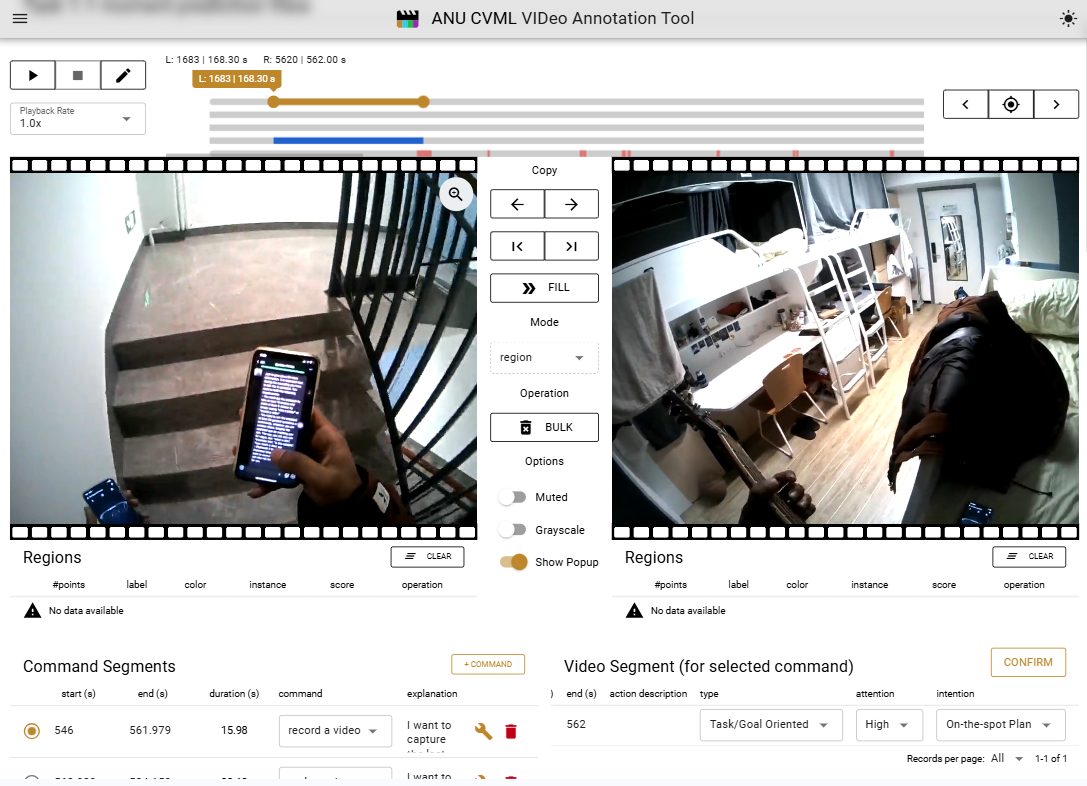}
    \caption{Annotation interface for Task1.1 (Moment of Recording).}
    \label{fig:task1.1_annotation_platform}
\end{figure}
Each participant annotated their egocentric video recordings using a custom web-based tool, Vidat, following a structured retrospective labeling protocol. The annotation interface displayed the full video alongside a timeline, with candidate time intervals pre-populated from automatic speech transcription as an initial reference. The transcription was quite accurate in locating the initiation of the recording commands, however, participants had to review and adjust these intervals to match their exact intentions. This process involved tweaking the boundaries to ensure that the selected video or photo segment corresponded precisely to the moment they intended to record. Participants were able to exclude content they didn’t want to capture, ensuring that the annotations reflected their true intent at the time of recording.

For each identified moment, participants created a command entry by selecting the time interval and specifying the intended capture modality. They were also required to provide a free-text \textit{explanation} field, describing the motivation or context behind their decision to record that particular moment (e.g., "interesting mural on the wall" or "wanted to remember this recipe").

Two distinct annotation workflows were followed, depending on the selected capture modality:

\textbf{Photo Annotation.} When participants selected \textit{make a photo} as the command type, they were required to identify the specific object or scene they intended to photograph by drawing bounding boxes on the relevant video frames within the annotated interval. Each bounding box was labeled with the object and included an \textit{expectation} field, where participants described either what they hoped to capture or communicate through the photograph or how they envisioned the photo's aesthetic qualities. The bounding boxes could be drawn across multiple frames within the same command interval, allowing participants to mark objects that appeared briefly or at varying positions. These annotated crops served as the ground truth for the photo object prediction benchmark, where models were tasked with identifying the correct object from a set of four candidate image crops extracted from the video scene.

\textbf{Video Annotation.} When participants selected \textit{record a video} as the command type, they were asked to complete a structured video segment descriptor in addition to the free-text explanation. This descriptor included the start and end times of the intended video recording, a summary of the ongoing activity in free text, and three categorical attributes characterizing the nature of the moment: the moment type, selected from the categories {\textit{Information Moment}, \textit{Visual Aesthetic Curiosity}, \textit{Emotional Reaction}, \textit{Social Interaction}, \textit{Task/Goal Oriented}}; the attention level, chosen from {\textit{High}, \textit{Normal}, \textit{Low}}; and the degree of planning, selected from {\textit{Pre Planned}, \textit{On-the-spot Plan}, \textit{Spontaneous}}. This structured data was required for all video annotation entries, ensuring that each video capture was accompanied by a detailed semantic description. The focus levels (High, Normal, Low) were used to help participants better characterize their attention during the recording, with task-oriented moments typically showing higher attention and curiosity-driven moments showing lower attention.

The quality of annotations was ensured through the presence of an experimenter from the research group, who guided participants through the UI, provided clarifications when needed, and ensured that all tasks were completed properly and in the correct format.

The result of this annotation process is a fully labeled timeline for each video session. This timeline includes both the intended capture moments, with their associated semantic attributes, object localizations, and personal motivations, providing a rich, structured dataset for training and evaluating models on egocentric recording intention prediction.

\subsection{Task1.2 - Emotion Analysis}
\begin{figure}[t]
    \centering
    \includegraphics[width=0.6\textwidth]{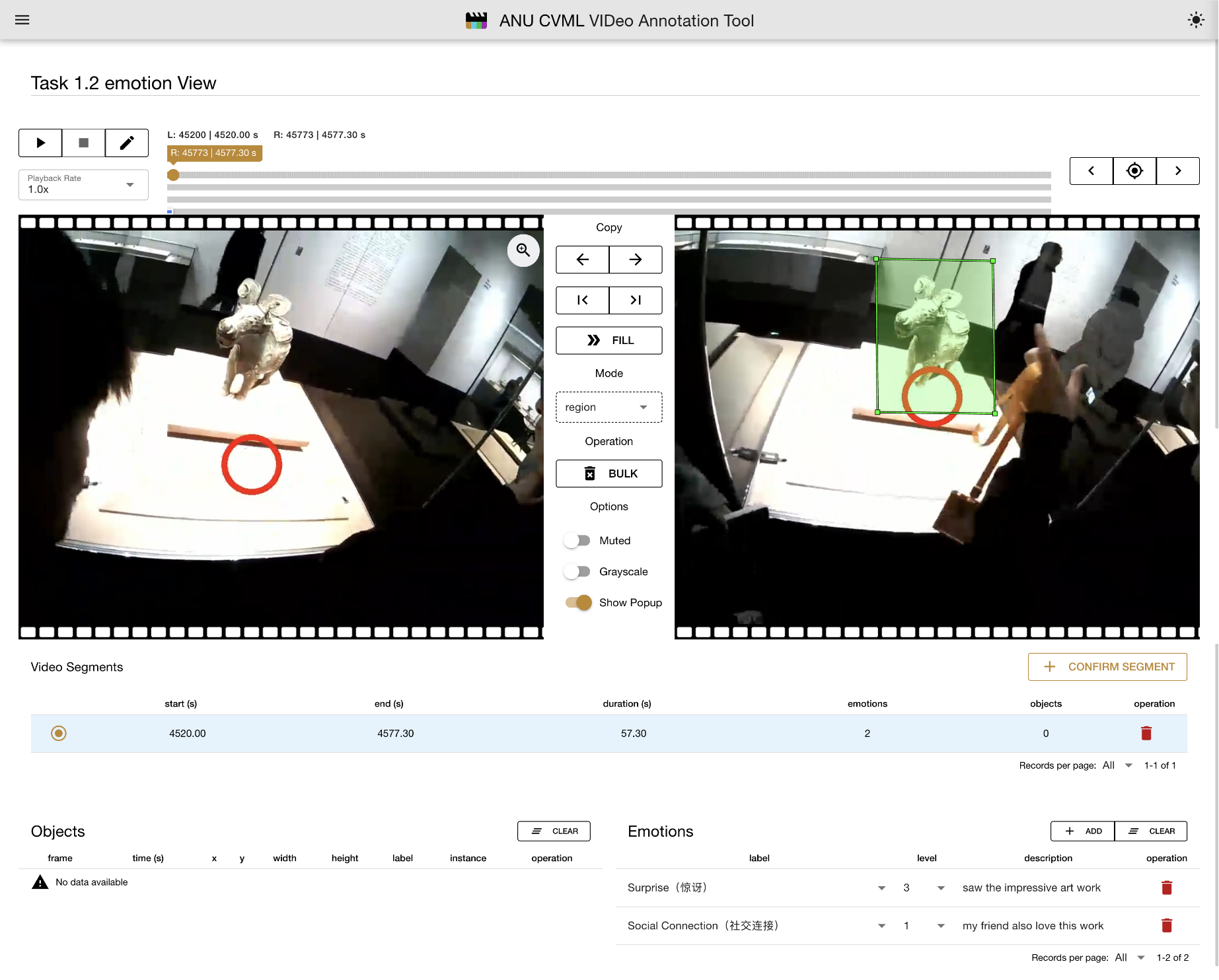}
    \caption{Annotation interface for Task1.2 (Emotion Analysis).}
    \label{fig:task12_annotation_platform}
\end{figure}

Participants annotated emotion-related moments on the Vidat platform through a retrospective review of their full egocentric recordings. To facilitate recall of fleeting affective experiences, the interface displayed the full video together with audio and gaze-overlay playback, allowing participants to inspect the surrounding context of each moment. During data collection, whenever participants noticed a salient emotional response, they sent a free-text instant message to the experimenter. These messages were not treated as labels themselves, but served as temporal anchors during retrospective annotation.

\textbf{Emotion Segment Annotation.} Based on the message timestamps, participants located the corresponding moments on the timeline and reviewed the surrounding footage to determine whether a noticeable emotion was present. For each identified emotional moment, they created or refined an annotation segment by selecting a time interval around the anchor timestamp. They were instructed to adjust the segment boundaries so that the annotated interval covered not only the emotional reaction itself but also the triggering event that gave rise to the emotion. In addition to the anchor-guided review, participants were also allowed to add new emotional segments while watching the full recording if they recalled other noticeable affective experiences that had not been marked during data collection.

\textbf{Emotion Description and Intensity.} For each segment, participants could assign one or multiple emotion labels selected from a predefined set consisting of {\textit{Joy}, \textit{Surprise}, \textit{Stress}, \textit{Social Connection}, \textit{Achievement}, \textit{Disappointment}, \textit{Awkward}, \textit{Boring}, \textit{Relaxation}}, with an additional \textit{Other} option for custom entries. They also provided a free-text \textit{reason} field to explain what caused the emotion, describe how they felt in that moment, and optionally clarify the relative importance of multiple selected labels. In addition, participants rated the self-perceived emotion intensity on a 1--5 scale, where 1 indicated a very weak feeling and 5 indicated a very strong feeling.

\textbf{Trigger Localization.} When the emotional response was clearly triggered by a visible object or person in the scene, participants were asked to draw a bounding box on the frame where that trigger could be clearly identified (e.g., if a participant felt joy upon seeing a friend, they would draw the bounding box on the frame where the friend became visible). Throughout the annotation session, an experimenter was present to explain the annotation rules, help participants refine ambiguous temporal boundaries, remind them of potentially missed emotional moments, and verify that all required fields were completed in a consistent format.

The result of this annotation process is a structured collection of emotion segments for each recording session. Each annotated entry includes the start and end timestamps of the emotional moment, one or multiple emotion labels with their corresponding intensity levels, the participant's free-text reason, and, when applicable, the bounding-box coordinates and associated frame for the visible object or person that triggered the emotion.

\subsection{Task2.1 - User-Initiated Contextual Request}

\begin{figure}[t]
    \centering
    \begin{subfigure}[b]{0.48\linewidth}
        \centering
        \includegraphics[width=\linewidth]{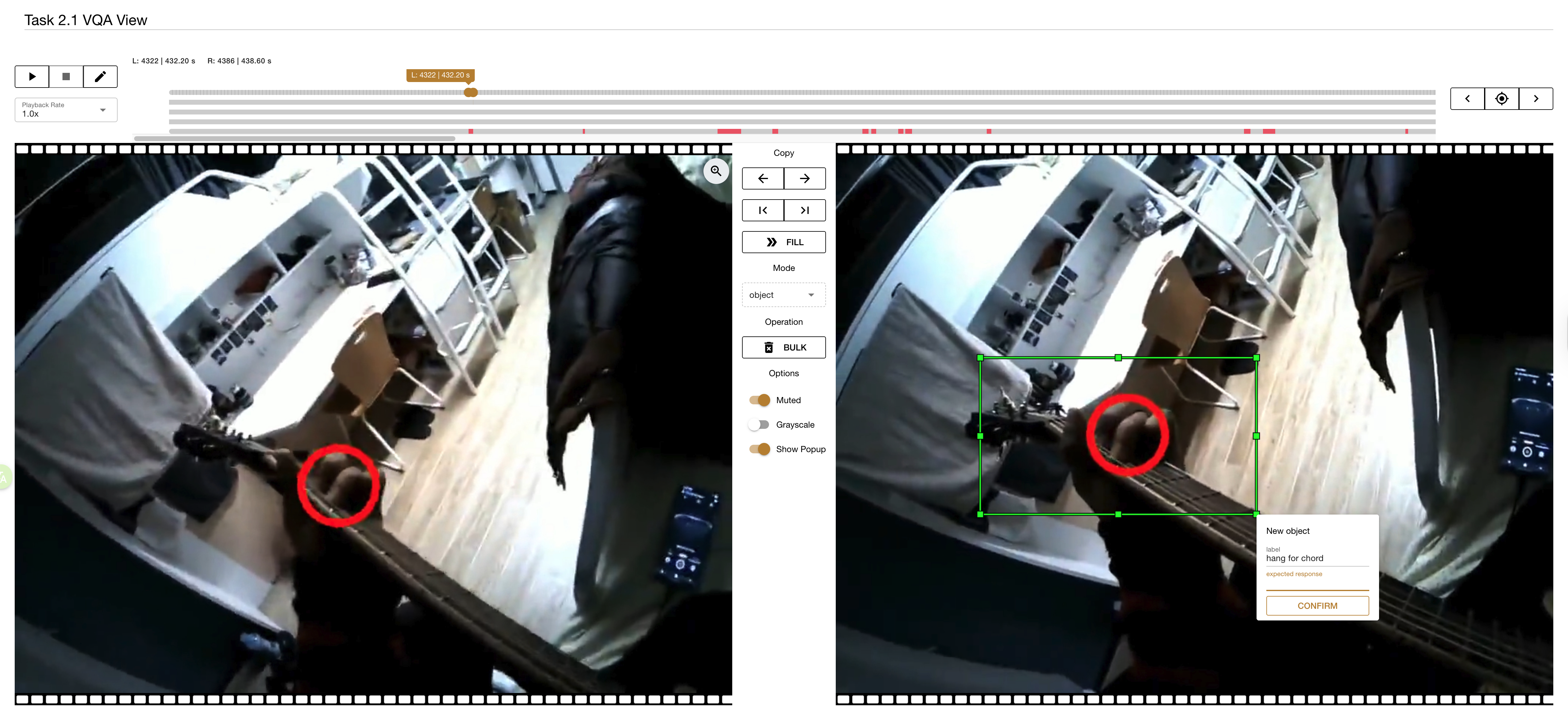}
        \label{fig:task2.1_vidat_screenshot1}
    \end{subfigure}
    \hfill
    \begin{subfigure}[b]{0.48\linewidth}
        \centering
        \includegraphics[width=\linewidth]{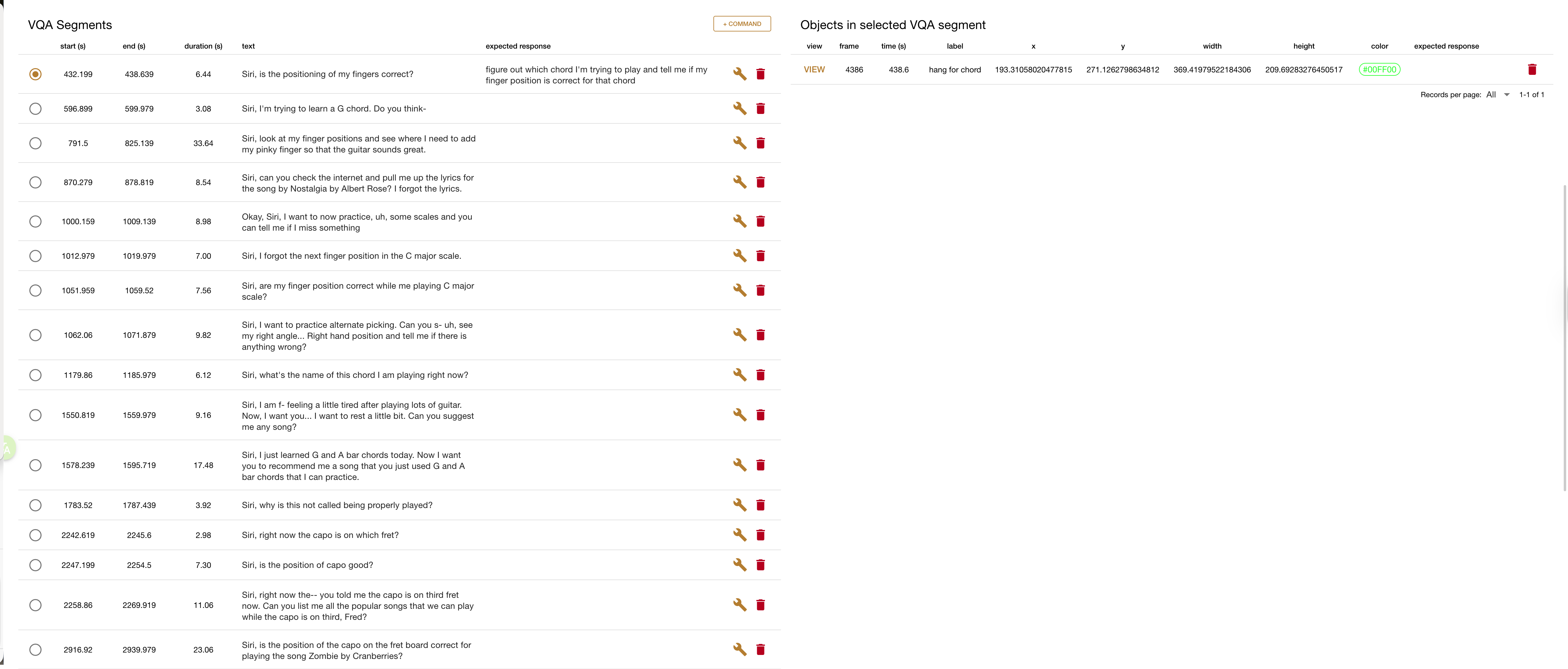}
        \label{fig:task2.1_vidat_screenshot2}
    \end{subfigure}

    \caption{Screenshot for task2.1's annotation UI.}
    \label{fig:task2.1_vidat_screenshot}
\end{figure}
\textbf{Annotation Platform.} As shown in Figure~\ref{fig:task2.1_vidat_screenshot}, the interface for Task2.1 is organized around three main components. First, a timeline control panel allows annotators to navigate the full recording and define the temporal span of each query event using two frame-accurate handles. The panel supports playback at multiple speeds, keyboard-based frame navigation, and visual indicators for previously annotated VQA segments, enabling annotators to quickly review and revise existing annotations. Second, a dual-view canvas displays the video frames corresponding to the start and end handles. For visually grounded queries, annotators can draw bounding boxes around the referred objects, resize or reposition them, and assign open-vocabulary labels. Each object annotation is linked to a selected VQA segment, ensuring that visual referents are structurally associated with the corresponding spoken query. Third, two tables summarize the current annotations: the VQA segment table stores the temporal boundaries, query text, and expected response for each query event, while the object table lists the bounding boxes and object-level expected responses associated with the selected segment.

\textbf{Annotation Procedure and Output.}
For each recording session, the experimenter helps upload the participant’s egocentric video together with the automatically extracted request events generated by the preprocessing pipeline, including their interval timestamps and transcribed text. Then, the participant reviews and refines these automatically generated entries. After confirming the request content, the participant writes the requirements describing what an ideal AI assistant should provide, such as the required information type, level of detail, or preference for where to find external knowledge. If the request refers to a visible object or scene entity, the participant further annotates its visual grounding by drawing bounding box on it's corresponding frame. The label describes what the object is and how it is related to the request. 

The final annotation is exported as a JSON file. Each exported file contains a list of request segments. Each segment includes the corrected start time, end time, duration, verified request transcript, segment-level expected response, and an optional list of visual referent annotations. 

\subsection{Task2.2 - Proactive Request Recommendations}
Task2.2 annotations were collected with two complementary modules. The first module supports Proactive Timing Judgment (PTJ) by asking participants to mark intervals in which a proactive assistant should not interrupt them. The second module supports Valuable Interaction Identification (VII) by asking participants to review, select, edit, order, and temporally localize generated proactive recommendations.

\textbf{Video availability selection module.}
Participants reviewed the egocentric recording with timeline-based playback and marked any interval during which they would not want to receive a proactive request and provide the reason. Each marked non-interruptible interval was visualized as a red bar on the timeline. The interface also displayed the corresponding start and end timestamps together with the participant-provided reason, allowing participants to verify and revise the annotations before export. These annotations provide the ground truth for PTJ, where the benchmark evaluates whether a model can identify contexts in which interruption is inappropriate.

\textbf{Recommendation selection and reordering module.}
For each candidate video segment, the interface displayed the fixation-overlaid video together with a list of proactive recommendations generated by the recommendation pipeline. Participants indicated which recommendations they would accept in the observed context, optionally revised the wording to better match their intended request, reordered accepted recommendations by preference or priority, and annotated the time window in which each accepted recommendation should be presented. These annotations provide the ground truth for VII, where the benchmark evaluates whether a model can select the participant-preferred proactive request from a set of candidates. The module includes four main interface components:

\begin{enumerate}
    \item \textbf{Video playback window.} The video player shows the egocentric clip, enabling participants to evaluate recommendations against the visual context.
    \item \textbf{Recommendation list area.} The interface lists all candidate recommendations for initial screening. Participants select recommendations they would be willing to receive; selected items are added to an accepted list that can be reordered by drag-and-drop. For each accepted item, participants annotate the appropriate presentation time window, which is shown as a blue bar on the timeline to support timing verification.
    \item \textbf{Recommendation editing area.} Participants can refine the wording of an accepted recommendation or rewrite it entirely when the generated text captures the right intent but not the preferred expression.
    \item \textbf{Object grounding tool.} When a recommendation refers to an ambiguous object or region, participants can draw a bounding box on the video frame to identify the intended referent. The interface records the bounding-box coordinates and frame timestamp alongside the recommendation annotation.
\end{enumerate}

\begin{figure*}
  \centering
  \includegraphics[width=0.98\textwidth]{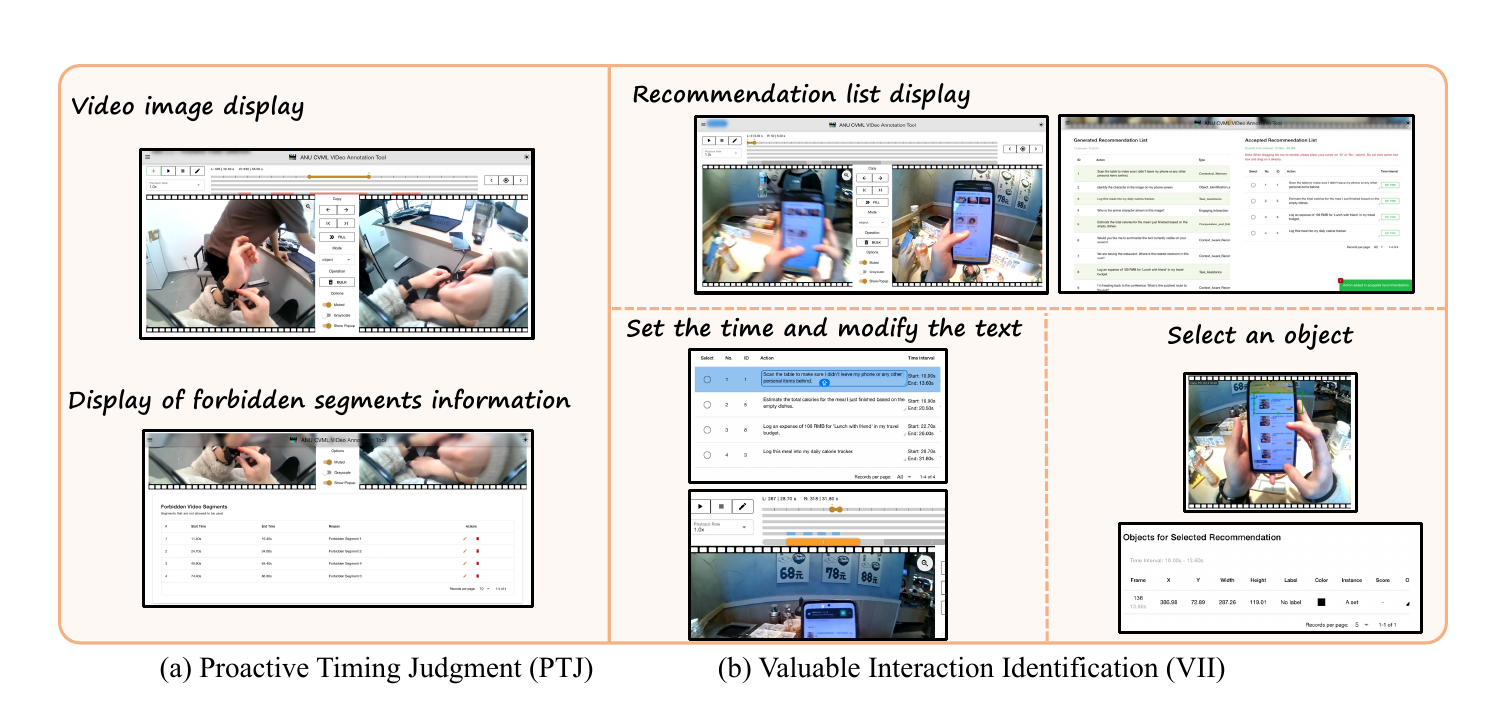}
  \vspace{-0.8em}
  \caption{Task2.2 annotation platform}
  \label{fig:task2_2_annotation_platform}
\end{figure*}

\subsection{Task3.1 - Memory Recall Prediction}
To ensure high-quality and consistent benchmarks, we developed a systematic pipeline to transform participants' raw episodic memories into structured, multi-level annotations.

\textbf{Data Preprocessing}. Following each recording session, participants provide a detailed account of the events they vividly recall, guided by a structured template that includes high-level event settings, locations, participants, primary activities, vivid details, and overall impressions. We accept accounts in both audio and text formats. These original files are processed through a standardized pipeline: audio files (e.g., .m4a) are first transcribed into text using the Xunfei or ElevenLabs API. Subsequently, a Large Language Model (LLM) is employed to decompose these vivid details into more granular, atomic segments.

\begin{figure*}
  \centering
  \includegraphics[width=0.98\textwidth]{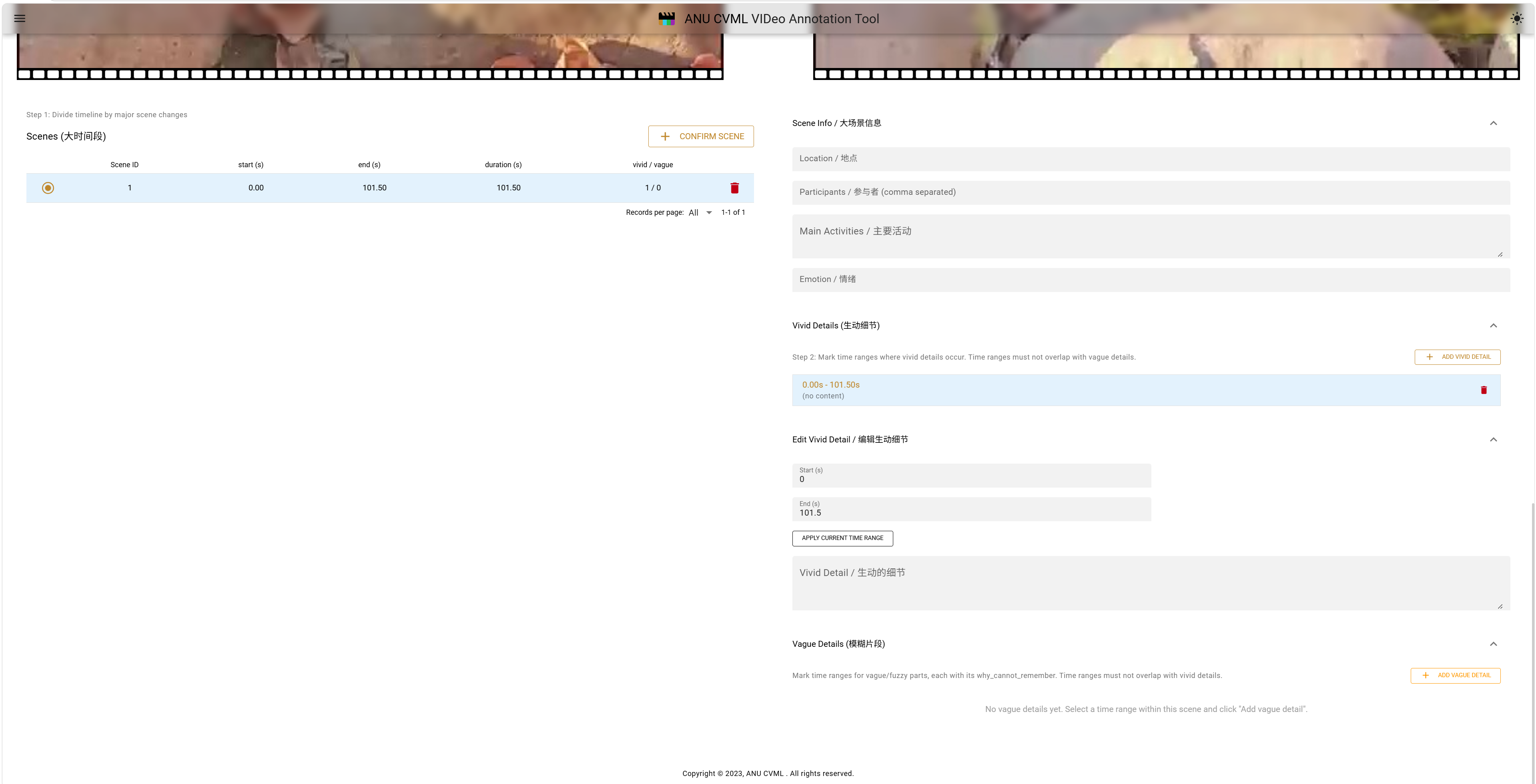}
  \vspace{-0.8em}
  \caption{Task3.1 annotation platform}
  \label{fig:task3_1_annotation_platform}
\end{figure*}

\textbf{Annotation}. The annotation is performed on the Vidat platform. After uploading the video, experimenters execute the following systematic steps: first, they establish timestamps for each high-level event setting; second, within each event setting, they document the specific location, participants, and primary activities. Finally, experimenters determine the timestamps for each atomic vivid detail. In cases where a detail occurs multiple times or is difficult to locate within the video, we consult with the participants for clarification. The process concludes by recording the exact content for each identified vivid detail. All annotations are completed by experimenters who strictly adhere to the participants' original accounts to ensure data integrity. See Fig.~\ref{fig:task3_1_annotation_platform} for details of annotation platform for MRP.
\subsection{Task3.2 - Memory Intent Modeling}

\textbf{In-situ Labeling.} During the data collection phase, participants are instructed to provide real-time verbal cues whenever they encounter an event or information they wish a "memory agent" to record. Specifically, participants trigger the labeling by saying, "Hey assistant, help me remember...", optionally followed by a brief description of the target information.

\begin{figure*}
  \centering
  \includegraphics[width=0.98\textwidth]{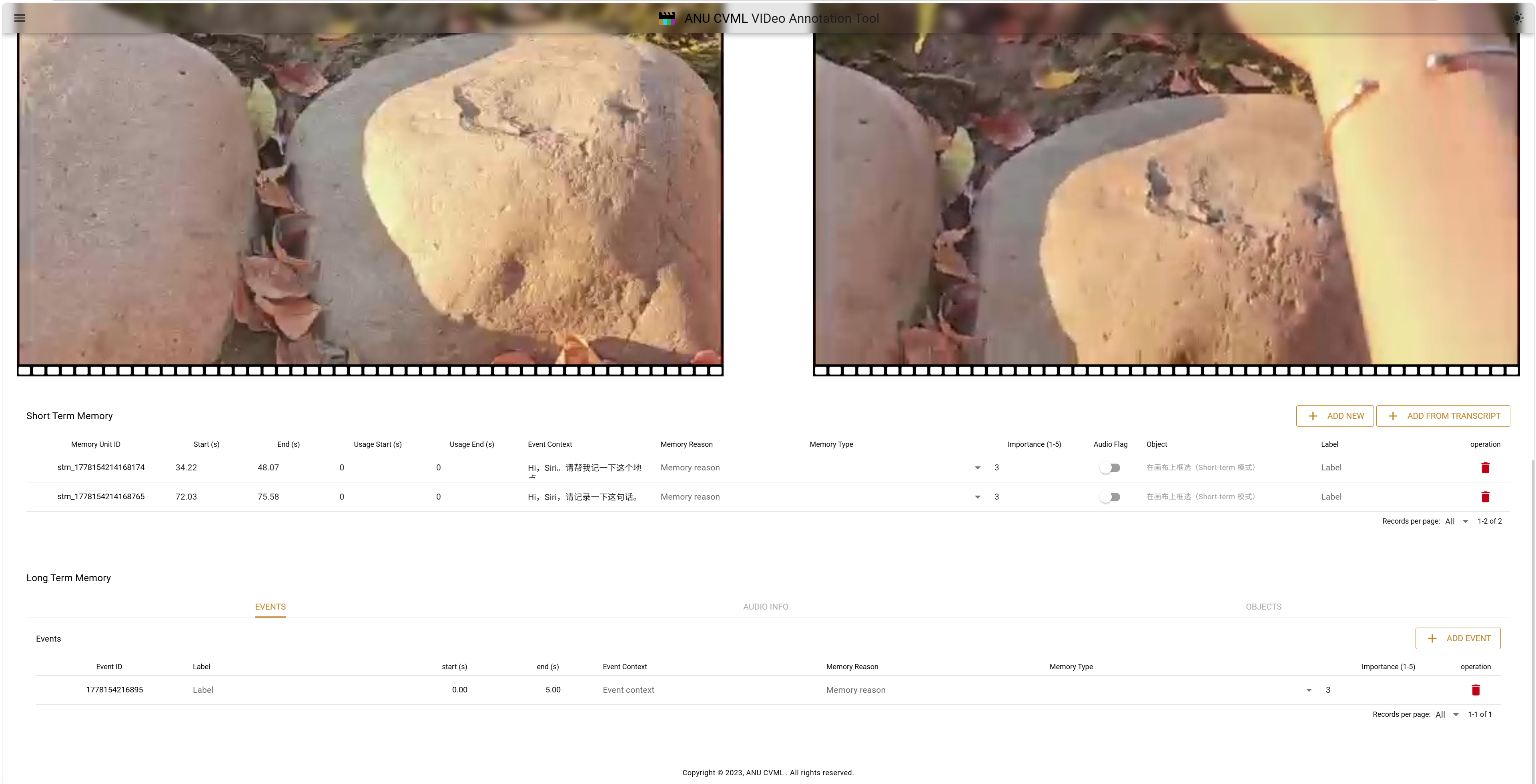}
  \vspace{-0.8em}
  \caption{Task3.2 annotation platform}
  \label{fig:task3_2_annotation_platform}
\end{figure*}

\textbf{Retrospective Annotation.} During the subsequent in-lab session, we employ an automated pipeline to transcribe the audio and extract the in-situ identifiers along with their corresponding timestamps into a structured JSON file. These data, along with the egocentric videos, are then integrated into the Vidat platform for refined annotation. Participants first calibrate the precise timestamps for each memory trigger. They then provide multi-dimensional metadata for each item, including the \textit{memory content}, \textit{motivation (reason)}, and \textit{memory type} (e.g., Possibly Useful Information, Task Reminder, Permanent Precious Memory, Knowledge/Insights, or Custom). For object-related memories, participants are required to draw a bounding box around the target object in a representative frame. Finally, each item is categorized as either \textbf{short-term} or \textbf{long-term} based on its projected utility: information intended for use within 24 hours is classified as short-term, while all other instances are labeled as long-term. See details in Fig. ~\ref{fig:task3_2_annotation_platform}

\subsection{Task3.2 - Memory Intent Modeling}
\section{Data Processing Algorithms for Annotation.}

\subsection{Transcript and Command Extraction}

The transcription pipeline processes continuous audio, extracts word-level transcriptions, identifies user commands directed at an AI assistant, and classifies them into categories.

\textbf{Stage 1: Audio Loading \& Chunking}

\textbf{Input:} Audio file (WAV format)

The audio is split into chunks for processing. Each chunk represents a segment of the original audio.

\[
\{A_0, A_1, \dots, A_{N-1}\}, \quad N = \left\lceil \frac{\text{len}(A)}{C_{\text{ms}}} \right\rceil
\]

Each chunk is defined as $A_i = \text{audio}[i \cdot C_{\text{ms}} : (i+1) \cdot C_{\text{ms}}]$.

\textbf{Stage 2: Transcription Engine Selection}

Three engines are supported:

\begin{itemize}
    \item \textbf{Xunfei:} Word-level transcription with timestamps.
    \item \textbf{ElevenLabs:} Word-level transcription with timestamps and speaker ID.
    \item \textbf{GPT-4o:} Plain text transcription without timestamps.
\end{itemize}

\[
T_{\text{xunfei}}(A_i) \rightarrow \{w_j = (\text{text}_j, \text{start}_j, \text{end}_j)\}_{j=1}^{M_i}
\]

\[
T_{\text{eleven}}(A_i) \rightarrow \{w_j = (\text{text}_j, \text{start}_j, \text{end}_j, \text{speaker}_j)\}_{j=1}^{M_i}
\]

\[
T_{\text{gpt}}(A_i) \rightarrow \text{text}_i
\]

\textbf{Stage 3: Temporal Alignment}

Timestamps for each chunk are adjusted using a time offset:

\[
t_{\text{offset}, i} = i \cdot C_{\text{ms}}, \quad
w_j^{\text{adj}} = (\text{text}_j, \text{start}_j + t_{\text{offset}, i}, \text{end}_j + t_{\text{offset}, i})
\]

\textbf{Stage 4: Full Transcription Aggregation}

All chunk transcriptions are combined into a single transcript:

\[
\mathcal{T} = \bigcup_{i=0}^{N-1} T(A_i) = \{w_1^{\text{total}}, \dots, w_K^{\text{total}}\}, \quad K = \sum_{i=0}^{N-1} M_i
\]

\textbf{Stage 5: Command Parsing}

Commands directed at the AI assistant are identified using GPT-4.1. A context window around the trigger phrase is extracted and analyzed. Retry logic is applied up to three times:

\[
P: \mathcal{T} \rightarrow \{c_k\}_{k=1}^{L}
\]

\textbf{Stage 6: Command Classification}

Parsed commands are classified as:

\[
c_k =
\begin{cases}
\text{recording} & \text{if command matches \{``make photo'', ``record video''\}} \\
\text{memory} & \text{if command matches \{``remember''\}} \\
\text{general} & \text{otherwise}
\end{cases}
\]

\textbf{Stage 7: Output Structuring}

Commands are organized by category:

\[
C_{\text{recording}} = \{c : \text{type}(c) = \text{recording}\}, \quad
C_{\text{memory}} = \{c : \text{type}(c) = \text{memory}\}, \quad
C_{\text{general}} = \{c : \text{type}(c) = \text{general}\}
\]

Each command includes:
\begin{itemize}
    \item type: Command category
    \item original\_text: Exact transcript text
    \item command: Normalized command
    \item explanation: User reasoning (generated by GPT, later reviewed)
\end{itemize}

\subsection{Proactive Recommendation Generation Pipeline for Task2.2}
\label{App:Proactive Recommendation Generation Pipeline for Task2.2}

\begin{figure}[t]
    \centering
    \includegraphics[width=\textwidth]{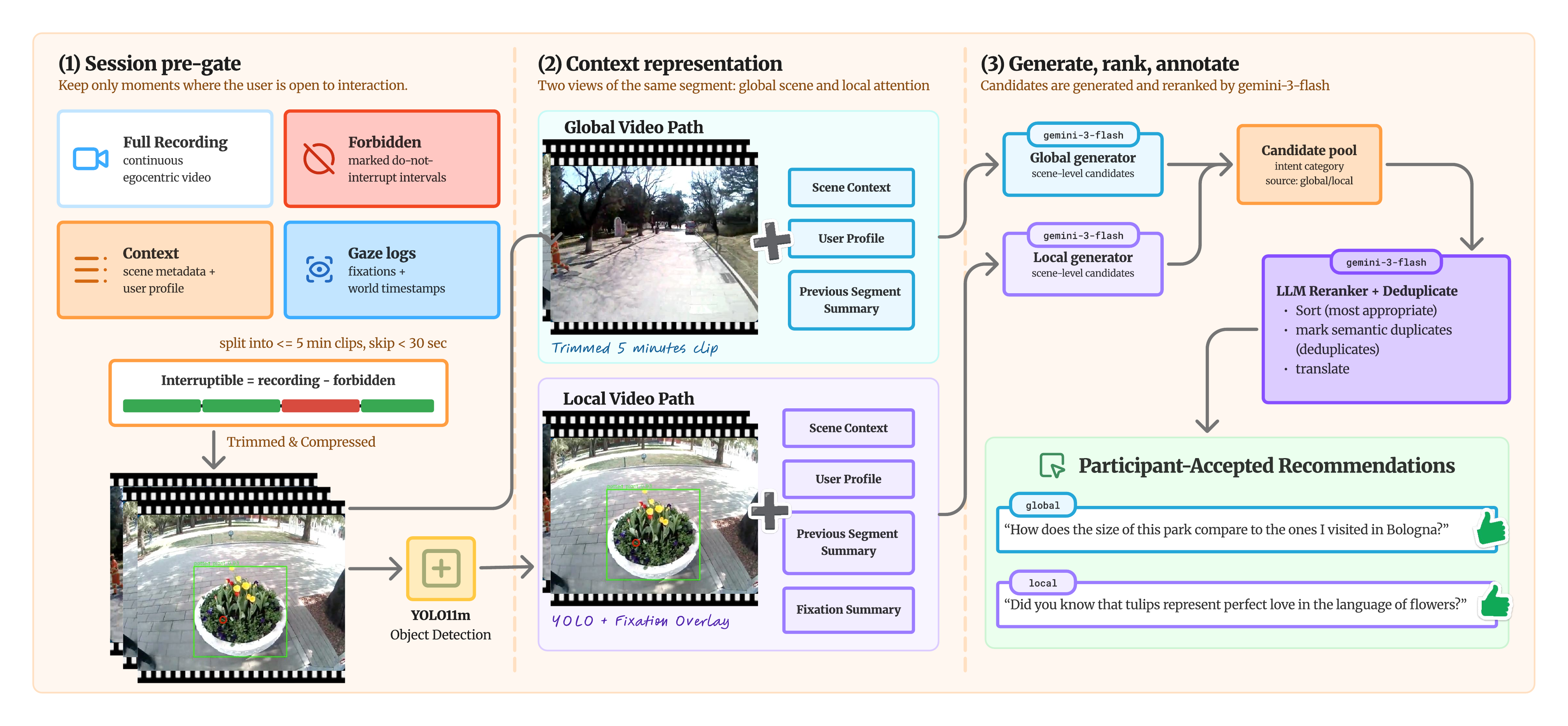}
    \caption{Overview of the Task2.2 proactive request recommendation generation pipeline. The system first removes participant-marked non-interruptible intervals, constructs five-minute interruptible segments, extracts global video context and local fixation-aware context, generates candidate recommendations with \texttt{gemini-3-flash-preview}, reranks and deduplicates them, and exports the full per-segment output for participant review.}
    \label{fig:task22_pipeline}
\end{figure}

The proactive recommendation pipeline converts continuous egocentric recordings into ranked interaction requests for Task2.2. Its goal is to propose interaction requests that a smart-glasses assistant could proactively surface at moments when the participant is willing to be interrupted. For each recording, it uses the egocentric video, participant-marked non-interruptible intervals, participant profile, scene-context annotations, fixation logs, and world-camera timestamps. It outputs a record per processed segment, including segment metadata, global and local summaries, fixation proportions, intermediate candidate pools, usage statistics, and final ranked proactive recommendations. The pipeline combines three sources of context: (i) the full egocentric video segment, (ii) object-level gaze fixation, and (iii) participant-specific profile and scene metadata. As summarized in Figure~\ref{fig:task22_pipeline}, it first gates out non-interruptible intervals, extracts fixation-aware visual context, generates global and local candidates in parallel, and reranks and deduplicates the resulting pool.

\paragraph{Stage 1: Interruptibility Pre-gating and Segment Construction.}
The inputs are full egocentric video $\mathcal{V}$, participant-marked forbidden intervals, scene-context metadata, participant profile, fixation CSV, and world-timestamp CSV.

Participants retrospectively mark intervals in which proactive recommendations should not be shown, such as privacy-sensitive, socially inappropriate, or otherwise undesirable moments. Let the raw forbidden intervals be
\[
    \mathcal{B} = \{[b_i^s, b_i^e]\}_{i=1}^{M}.
\]
Only the portion of each interval within the valid recording window contributes to the complement, and overlapping spans are merged into a single non-interruptible set:
\[
    \widetilde{\mathcal{B}}
    =
    \operatorname{Merge}\left(
    \{[\max(0,b_i^s), \min(T,b_i^e)] \mid b_i^e > b_i^s\}_{i=1}^{M}
    \right).
\]
The complementary interruptible intervals over the full recording duration $T$ are
\[
    \mathcal{I} = [0,T] \setminus \bigcup_{[s,e]\in\widetilde{\mathcal{B}}} [s,e].
\]
Each interruptible interval is split into segments of at most 300 seconds, with segments shorter than 30 seconds discarded. For each retained segment $v_k=[t_k^s,t_k^e]$, the original 1600$\times$1200 recording is trimmed and compressed to height 480 pixels, 500k video bitrate, and 2 FPS. The segment inherits the scene-context annotation with the largest temporal overlap, and the previous segment summary is carried forward as temporal context.

\paragraph{Stage 2: Visual Object Detection.}

For each segment, an Ultralytics YOLO11m obeject detector~\cite{Jocher_Ultralytics_YOLO_2023} returns, for each processed frame $f_j$, a set of detections
\[
    D_j = \{(o_{j,l}, b_{j,l}, p_{j,l})\}_{l=1}^{n_j},
\]
where $o_{j,l}$ is the predicted object class, $b_{j,l}=(x_1,y_1,x_2,y_2)$ is its bounding box, and $p_{j,l}$ is the confidence score. These detections define the object vocabulary used by the local recommendation path and separate attended objects from broader background context.

\paragraph{Stage 3: Gaze--Video Synchronization.}

Fixation logs and processed video frames differ in timestamp space and spatial resolution. Gaze coordinates are calibrated in the original 1600$\times$1200 world-camera coordinate system, while detection and generation operate on the compressed clip. After removing non-finite rows, the synchronizer filters both streams to the current segment:
\[
    \tau_k^s = \tau_0 + 10^9 t_k^s,\quad
    \tau_k^e = \tau_0 + 10^9 t_k^e,
\]
where $\tau_0$ is the first world-camera timestamp in nanoseconds, $t_k^s$ and $t_k^e$ are segment-relative times in seconds, and $\tau_k^s,\tau_k^e$ are the corresponding gaze-log bounds. World-camera timestamps are indexed to match the 2 FPS processed video. Fixation coordinates are scaled to the processed resolution by
\[
    x' = x \cdot \frac{W_{\text{video}}}{W_{\text{orig}}}, \qquad
    y' = y \cdot \frac{H_{\text{video}}}{H_{\text{orig}}}.
\]
Fixations that cross segment boundaries are clipped, and their durations are recomputed, yielding fixation points that can be compared directly with YOLO boxes.

\paragraph{Stage 4: Fixation Proportion Computation.}

The fixation module converts gaze behavior into an object-level attention distribution. For each clipped fixation $g_r=(x_r,y_r,\tau_r^s,\tau_r^e,d_r)$, where $(x_r,y_r)$ is the scaled fixation point and $d_r=(\tau_r^e-\tau_r^s)/10^6$ is its duration in milliseconds, the corresponding video frame is found by timestamp lookup. If the fixation point falls inside one detected object box, the duration is credited to that object class. If multiple boxes contain the point, non-person objects are prioritized over person boxes, and the smallest box in the selected group is used. Fixations outside all detected boxes remain visible in the overlay video but are excluded from the denominator of $\rho(o)$.

The accumulated fixation duration for each object class is normalized into a percentage:
\[
    \rho(o) =
    \frac{\sum_{r: \operatorname{hit}(g_r)=o} d_r}
    {\sum_{o'} \sum_{r: \operatorname{hit}(g_r)=o'} d_r}
    \times 100.
\]
Objects below a 5\% fixation threshold are removed, and at most the top five object classes are retained. The remaining proportions are renormalized to integer percentages that sum to 100:
\[
    F_k = \{(o_1,\hat{\rho}_1), \dots, (o_m,\hat{\rho}_m)\}, \quad m \leq 5.
\]
The resulting fixation summary $F_k$ is serialized as lines such as ``menu: 52\%'' or ``laptop: 31\%''. The pipeline also renders a fixation-overlaid clip with gaze points and fixation-hit boxes. This rendered video is used by local generators and rerankers that benefit from an explicit visualization of what the user attended to.

\paragraph{Stage 5: Parallel Global and Local Recommendation Generation.}

The core recommendation stage has two complementary generation paths.

\textbf{Global path.} The global generator receives the processed clip, scene context, participant profile, and previous-segment summary. It produces a short factual video summary and proactive candidates grounded in the overall scene, such as navigation help, task support, or contextual reminders.

\textbf{Local path.} The local generator receives the fixation-overlaid clip, fixation summary $F_k$, participant profile, scene context, and previous-segment summary. Its outputs are restricted to objects in $F_k$, capturing opportunities such as object identification, text translation, product comparison, artifact logging, or memory-related questions about attended objects.

Both generators use a structured response schema:
\[
    G(v_k) \rightarrow (\text{summary}, \mathcal{R}^{\text{global}}_k), \qquad
    L(\tilde{v}_k,F_k) \rightarrow (\text{summary}, \mathcal{R}^{\text{local}}_k),
\]
where $v_k$ is the compressed clip and $\tilde{v}_k$ is its fixation-overlaid counterpart.
The final generation pipeline fixes both global and local generators to \texttt{gemini-3-flash-preview}, selected through qualitative analysis and trial runs for its price-performance trade-off on grounded, diverse, and well-structured recommendations. Structured outputs are enforced with response models and retry logic. The nine output categories are: (1) object identification and recognition; (2) translation, text, and audio recognition; (3) procedural guidance; (4) decision support; (5) contextual memory; (6) computation and estimation; (7) task assistance; (8) context-aware recommendation; and (9) engaging interaction. The prompts require concise first-person phrasing, realistic assistant-executable actions, English outputs, and strict scene grounding. The global and local paths run in parallel when fixation data are available; otherwise, the pipeline uses the global path only. Table~\ref{tab:task22_generation_prompts} summarizes the generation and reranking prompt templates; the full executable prompts are released with the pipeline repository.

\begin{table}[htbp]
    \centering
    \caption{Prompt templates used by the Task2.2 recommendation generation pipeline.}
    \scriptsize
    \setlength{\tabcolsep}{2pt}
    \begin{tabularx}{\textwidth}{@{} l >{\raggedright\arraybackslash}X >{\raggedright\arraybackslash}X @{}}
        \toprule
        Prompt           & Inputs                                                                                          & Core Instructions / Output                                                                                                                                                                                                                                                                                                        \\
        \midrule
        Global generator & Video clip; scene context; participant profile; previous-segment summary                               & Produce a factual video segment summary and proactive questions, commands, or interactions for each intent category. Recommendations must be grounded in the visible scene, personalized to the participant profile, actionable by a smart-glasses assistant, concise, first-person, English, and non-hallucinatory.                           \\
        Local generator  & Fixation-overlaid clip; fixation summary; scene context; participant profile; previous-segment summary & Produce a factual video segment summary and fixation-grounded recommendations. All recommendations must directly relate to objects in the fixation summary and avoid broad scene-level suggestions handled by the global path.                                                                                                                  \\
        Reranker         & Candidate pool; scene context; participant profile; previous-segment summary; fixation summary         & Sort all candidates from most to least appropriate, mark semantic duplicates without dropping inputs, move ungrounded or hallucinated candidates to the bottom, translate recommendation text to Simplified Chinese, and provide a short ranking reason while preserving the original English recommendation and category fields. \\
        \bottomrule
    \end{tabularx}
    \label{tab:task22_generation_prompts}
\end{table}

\paragraph{Stage 6: Candidate Pool Construction.}

The structured outputs from both generators are flattened into a single candidate pool:
\[
    \mathcal{C}_k =
    \{(r_i, z_i, s_i)\}_{i=1}^{N},
\]
where $r_i$ is the recommendation text, $z_i$ is the intent category, and $s_i \in \{\text{global},\text{local}\}$ records the source path. This tag is preserved so downstream analysis can compare broad scene-level recommendations with fixation-specific recommendations.

\paragraph{Stage 7: LLM Reranking and Semantic Deduplication.}

\texttt{gemini-3-flash-preview} is also used as the LLM reranker. Given the candidate pool, scene context, participant profile, previous summary, and fixation summary, the reranker sorts candidates from most to least appropriate, checks grounding, marks semantic duplicates, translates recommendations into Simplified Chinese for Chinese-speaker participants while preserving the English source text, and provides a short ranking reason. The structured call validates that the multiset of input candidates matches the multiset of output candidates, preventing dropped, added, or rewritten candidates. Candidates marked as semantic duplicates are then removed, and the remaining recommendations receive monotonically increasing IDs.

\paragraph{Stage 8: Output Structuring for Annotation.}

For each processed segment, the full output contains segment metadata, original and fixation-overlaid clip paths or URLs, participant profile, scene context, previous summary, global and local video segment summaries, fixation proportions, intermediate candidate pools, token usage, and final ranked recommendations. These records are imported into the retrospective annotation workflow, where participants review the ranked recommendations, select those they would accept, and annotate when they should be presented. The generation pipeline therefore supplies diverse, context-grounded candidates, while participant feedback provides the ground truth for Task2.2.


\section{Detailed Experiment Pipeline}
\label{app:experiment_pipeline}

Each experimental session comprised two consecutive parts: (1) a 2--4 hour in-the-wild recording session with continuous wearable sensing, and (2) a 3--4 hour annotation session conducted on the same day or after a short interval. All procedures were approved by the IRB; participants provided written informed consent and received compensation at no less than local minimum wage. Participants were required to wear smart glasses without prescription eyeglasses (contact lenses permitted). Prior to recording, participants completed the \texttt{Scene context} field in the Participant Metadata spreadsheet and were instructed to imagine a smart-glasses assistant supporting photo/video capture, memory assistance, and open-domain QA, issuing all commands prefixed by in-situ identifier. 

\subsection{Wearable Hardware Stack}

Table~\ref{tab:pipeline_hardware} summarizes the sensor suite. The Pupil Labs Neon Glasses were tethered to a Samsung phone running the Neon Companion app. The $\tau$-Ring was worn on either hand; the BioVital S2 Wristband on the non-dominant wrist; and an Apple Watch on the dominant wrist. All devices were time-synchronized to the egocentric video via visual calibration procedures described below.

\begin{table}[t]
\centering
\scriptsize
\setlength{\tabcolsep}{2pt}
\begin{tabularx}{\textwidth}{@{}>{\centering\arraybackslash}X >{\centering\arraybackslash}X >{\centering\arraybackslash}X >{\centering\arraybackslash}X@{}}
\toprule
Device & Placement & Role & Data Stream \\
\midrule
Pupil Labs Neon Glasses & Head & Egocentric video + eye gaze & RGB @ 30 fps, 2D gaze \\
$\tau$-Ring & Index Finger & Physiological + behavioral sensing & HRV, PPG, IMU, temp. \\
Apple Watch (S8/9/11) & Dominant wrist & Physiological sensing & Accel., heart rate, IBI \\
BioVital S2 Wristband & Non-dominant wrist & Physiological sensing + event marker & EDA, sEMG, motion, temp., 8-ch.\ PPG, button-press \\
Samsung Galaxy S25 / S25 Ultra & Bag or Pocket & Glasses host + recording proxy & --- \\
\bottomrule
\end{tabularx}
\caption{Wearable hardware stack.}
\label{tab:pipeline_hardware}
\end{table}

Four Samsung phones were used across sessions: Galaxy S25 (12\,GB RAM, 256\,GB storage; units S25-1, S25-2, S25-3) and Galaxy S25 Ultra (12\,GB RAM, 512\,GB storage; unit S25Ultra-1). Apple Watch Series 8, 9, and 11 were used interchangeably on the dominant wrist across the project. Data from all Apple Watch series are treated as equivalent for the core physiological signal stream (accelerometer, heart rate, IBI). The glasses draw power from the paired phone via data cable; the phone battery was verified sufficient for the full session before departure.

\paragraph{Evolution of the wrist-worn stack.} The wrist-worn sensing stack evolved during early piloting. The first sessions used a naive self-developed wristwatch produced by an early-stage vendor. This device had no explicit start/stop button and required post-hoc temporal alignment via a PPG-based black/white boundary method: the experimentor rapidly moved the wristband across a high-contrast edge while the glasses recorded the scene; the sharp PPG signal change at each boundary crossing was later matched to the corresponding video frame. Data quality from this device proved unreliable, and the device was discarded after initial piloting. It was replaced by the BioVital S2 Wristband, which provides dense multi-modal physiological sensing, an integrated event-marker button, and auto-start on wear. The BioVital S2 simultaneously acquires seven biological modalities: EDA, sEMG, accelerometer, gyroscope, skin temperature, dynamic pulse rate and blood oxygen via an 8-channel PPG sensor, and heart rate variability (HRV).

\subsection{Pre-recording Device Preparation and Calibration}

The experimenter executed the following steps in strict order, recording all timestamps to the nearest second.

\paragraph{Pupil Labs Neon Glasses.} Connected the glasses to the Samsung phone via data cable. Launched the Neon Companion app and granted camera sync permission when prompted. Verified the egocentric feed and eye-tracking overlay via the preview button; the red circle indicates the estimated gaze point. Created a participant profile and performed eye-tracking calibration by tapping the preview screen. The participant fixated a target while the experimenter dragged the gaze dot to the true fixation point until converged. Tapped record and immediately locked the phone screen to prevent accidental stoppage. Recorded the glasses start timestamp.

\paragraph{$\tau$-Ring.} Placed the ring in the charging case; a green light in the upper-left corner indicates active charging. The magnetic charging interface is position-sensitive and may require several attempts to seat correctly. Launched the OpenRing app, paired via \texttt{Dashboard} $\rightarrow$ \texttt{Scan Ring} $\rightarrow$ \texttt{Start Scanning}, selected the target ring from \texttt{Discovered Devices}, and tapped \texttt{Connect}. On the \texttt{Dashboard}, tapped \texttt{Update Time} then \texttt{Calibrate}. In the \texttt{Offline} tab, tapped \texttt{Get File List}; formatted storage (\texttt{Format File System}) if non-empty. Set both \texttt{Total Exercise Duration} and \texttt{Exercise Segment Duration} to 14,400 seconds (4 hours), then tapped \texttt{Start Exercise}. The ring records autonomously and stops automatically; it cannot be manually interrupted. Recorded the ring start timestamp.

\paragraph{BioVital S2 Wristband.} The participant wore the wristband on the non-dominant wrist; logging starts automatically when worn and stops on removal. Verified the display showed \texttt{recording}. Recorded the wristband start timestamp.

\paragraph{Apple Watch.} Worn on the dominant wrist. Launched WatchSensor-watchOS from the bottom application dock and started recording. Recorded the watch start timestamp.

\paragraph{Cross-device alignment.} After setting up all the devices, the participant needs to clap three times while looking at their hands, separating hands immediately after each clap, and holding still for approximately one second between claps. The IMU acceleration peak at each clap was matched to the video frame showing hand contact. This procedure was repeated at the end of the session to mark the end boundary. The experimenter recorded both alignment timestamps. Post-hoc, the detected IMU peak was temporally offset to its corresponding video frame, yielding a per-device time offset to align the ring and Wristband stream to the egocentric video timeline. 

\subsection{In-the-Wild Recording Tasks}

Participants performed their planned daily activities while wearing the full sensor stack and issued voice commands with the following minimum quotas:

\paragraph{Task 1: Photo / video capture (suggests $\geq$ 10 commands).} Format: ``Hey Assistant, take a photo'' or ``Hey Assistant, start recording,'' optionally with a description (e.g., ``this sculpture is hilarious''). Purpose: capture visual moments deemed worth recording.

\paragraph{Task 2: Open-domain questions / commands (suggests $\geq$ 20 commands).} Situationally grounded, non-trivial questions spanning definition (``Hey Assistant, What is this dish called?''), explanation (``Hey Assistant, Why is my commute longer today?''), comparison (``Hey Assistant, Which product fits my needs better?''), subjective retrieval (``Hey Assistant, Did I feed my cat this morning?''), arithmetic, assistance, translation, and objective information (``Hey Assistant, Is parking allowed here?'').

\paragraph{Task 3: Memory assistance} Format: ``Hey Assistant, help me remember this'' for prices, routes, phone numbers, or to-do items expected to be useful shortly after utterance.

\paragraph{Command marking.} Each command was marked by pressing and holding the lower (marker) button on the BioVital S2 Wristband for approximately 2 seconds; a confirmation page appeared, and the participant slid the close slider to dismiss it. The upper (red) button was power only; accidental long-press would trigger shutdown and was avoided during the session.

\paragraph{Emotion self-report.} On salient emotional events, participants sent themselves a timestamped message (e.g., ``1'', ``surprised'', ``frustrated'', ``bored'') via WeChat or any time-stamped messaging app. These timestamps guided later emotion annotation.

\paragraph{Scene selection guidelines.} Participants were advised to: choose well-lit hours for outdoor activities; include at least one walking/moving segment; select scenarios where questions naturally arise (avoiding prolonged silent activities); ensure hands and manipulated objects were visible to the egocentric camera; and engage actively in multi-person interactions when applicable.

\subsection{Post-recording Shutdown and Data Export}

Shutdown order: (1) IMU clap alignment (end boundary), (2) wristband removed (auto-stops), (3) Apple Watch stopped, (4) glasses stopped, (5) participant completed \texttt{User profile} in the Participant Metadata spreadsheet.

\paragraph{Glasses export.} Raw data was exported from the Samsung phone to the PC via USB. Cloud-processed data was uploaded via the Neon Companion app to Pupil Cloud with the network proxy active. Both raw and processed data were saved to the participant's raw data folder.

\paragraph{Video compression.} The exported video was compressed to produce a concatenated video with audio and a muted variant for annotation.

\paragraph{Ring export.} Via the OpenRing app, the current session was downloaded to the phone and transferred to the PC. The resulting binary log ($\sim$10 MB nominal; $\sim$300 B indicates low battery; 0 B indicates firmware failure) was saved to the participant's raw data folder.

\paragraph{Wristband export.} The BioVital S2 was connected to the PC via USB and restarted; it mounted as a USB drive. Files were copied to a local import folder and processed through the BiovitalLab app. To create a project for the participant, import the matching record, and export to CSV format. The exported data was uploaded to the participant's raw data blob container.

\paragraph{Watch export.} Apple Watch data were synced to the paired iPhone and uploaded to the participant's raw data folder.

\subsection{Annotation Phase}

The annotation phase consisted of one mandatory same-day task and several subsequent tasks scheduled after a short delay to balance memory fidelity with participant availability.

\paragraph{Same-day tasks.}

\begin{itemize}[nosep]
    \item \textbf{Task 2.2.1: Do-not-disturb segments.} Using Vidat, the participant stated personal interruption criteria then annotated all segments where they would not want the assistant to intervene. Exported as \texttt{forbidden\_segments\_Px\_Name.json}, serving as prerequisite input for the recommendation-generation algorithm.
    \item \textbf{Task 1.2: Emotion annotation.} The participant identified segments eliciting salient emotional reactions, drew bounding boxes around triggering objects when visible, selected emotion label and intensity, and optionally provided free-text explanation. Exported as \texttt{annotations\_task1.2\_Px\_Name.json}.
\end{itemize}

\paragraph{Between-session preprocessing.}

\begin{enumerate}[nosep]
    \item \textbf{Audio extraction.} Audio was extracted from the compressed video for downstream transcription.
    \item \textbf{Speech transcription.} A Whisper-based pipeline was applied to the extracted audio to produce word-level transcripts and parse verbal commands. This yielded five JSON files stored in the transcripts folder: \texttt{all\_transcripts}, \texttt{all\_commands}, \texttt{general} (open-domain questions), \texttt{memory} (memory commands), and \texttt{recording} (photo/video commands).
    \item \textbf{Recommendation generation} (for Task 2.2.2). The recommendation pipeline consumed the egocentric video, participant profile, scene context, fixation logs, and do-not-disturb segments to generate ranked proactive recommendations for each interruptible segment. The output was a combined JSON file imported into Vidat for participant review.
\end{enumerate}

\paragraph{Subsequent annotation tasks (with-in one week).}

\begin{itemize}[nosep]
    \item \textbf{Task 3.1.1: Oral recall.} Following each recording session, participants provided a detailed retrospective account of events they vividly recalled. They were audio-recorded while freely narrating everything they remembered from the recording period. A structured template guided the narration: high-level event settings (e.g., restaurant, commute, gym), locations, participants, primary activities, vivid details (salient objects, surprising moments, sensory impressions), forgotten items or fuzzy memories, overall emotional valence, and a self-rated memory completeness score (0--5 integer). Accounts were accepted in both audio (e.g., \texttt{.m4a}) and text formats; most participants chose to narrate orally. Audio files were transcribed into text using the Xunfei or ElevenLabs API. The resulting transcript, whether originally text or transcribed from audio, was then processed through an LLM pipeline that decomposed free-form recollections into atomic, timestamp-granular detail segments for downstream annotation.
    \item \textbf{Task 1.1: Moment-of-record prediction.} The audio-derived command JSON was loaded into Vidat. For each detected photo command: the participant drew a bounding box on the target object, labeled the category, and filled \texttt{action description}, \texttt{type}, \texttt{attention}, and \texttt{intention} (planned beforehand / on-site / spontaneous). For video commands: selected the target segment. Missed commands were added manually with \texttt{+COMMAND}.
    \item \textbf{Task 2.1: VQA.} The open-domain command JSON was loaded into Vidat. The participant verified each detected question: deleted erroneous or duplicate entries, corrected transcription errors, and wrote the Expected Response. Bounding boxes were drawn for deictic references (``this,'' ``that,'' ``it'').
    \item \textbf{Task 2.2.2: Recommendation selection.} The algorithm-generated recommendation JSON was uploaded into Vidat. The participant selected useful information, excluded unwanted suggestions, ranked by preference, and drew bounding boxes for ambiguous object references.
    \item \textbf{Task 3.1.2: Memory recall annotation.} The transcribed oral-recall output was loaded into Vidat. Experimenters established timestamps for each high-level event setting, documented location, participants, and primary activities within each setting, and localized each atomic vivid detail to its precise temporal boundaries. The exact content of each detail was recorded per the participant's original account.
    \item \textbf{Task 3.2: Memory annotation.} The memory command JSON was loaded into Vidat. Short-term: verified each ``help me remember'' utterance, edited text, and completed \texttt{reason}, \texttt{type}, anticipated \texttt{usage}, \texttt{importance}, and \texttt{usage\_time}; drew bounding boxes and \texttt{Role} field for visually grounded memories; marked audio segments for auditory memories. Long-term: the participant voluntarily added any additional information worth remembering with free-text justification.
\end{itemize}

\section{Dataset and Benchmark Statistic}

\begin{table}[t]
\centering
\scriptsize
\setlength{\tabcolsep}{2pt}
\begin{tabularx}{\linewidth}{l l l l l l l l}
\toprule
\textbf{Dataset} & \textbf{Domain} & \textbf{Modality} & \textbf{Hours} & \textbf{\#Clips} & \textbf{Dur./Clip} & \textbf{Capture Protocol} & \textbf{Annot. Source} \\
\midrule
EPIC-KITCHENS~\cite{damen2022rescaling} & Kitchen & Video & 100 & 700 & 8.5 min & Researcher-defined & External \\
Ego4D~\cite{grauman2022ego4d} & Daily Activities & Video + Gaze + 3D & 3,670 & 9,645 & 22.8 min & Researcher-defined & External \\
EgoExo4D~\cite{Grauman_2024_CVPR} & Skilled Activities & Video + Gaze + 3D & 1,286 & 5,035 & 1--42 min & Researcher-defined & External \\
EgoExoLearn~\cite{Huang_2024_CVPR} & Task Execution & Video + Gaze & 120 & 432 & 13.4 min & Researcher-defined & External \\
EgoLife~\cite{yang2025egolife} & Daily Life & Video + Gaze + 3D + IMU & 266 & 6 & 44.3 h & Researcher-defined & External \\
\midrule
\textbf{EgoIntrospect} & Daily Life & Video + Gaze + PS* & 180 & 60 & 119-240 min & \textbf{Subject-chosen} & \textbf{Self-annotation} \\
\bottomrule
\end{tabularx}
\vspace{2pt}
\caption{Comparison of egocentric datasets by capture protocol and annotation source. PS* denotes Physiological Signals. EgoIntrospect differs by adopting user-chosen capture with self-annotation.}
\label{tab:dataset_comparison}
\end{table}

\begin{table}[t]
\centering
\scriptsize
\setlength{\tabcolsep}{4pt}
\begin{tabular}{l l l l l l l}
\toprule
\textbf{Benchmark} & \textbf{Source} & \textbf{\#QAs}  & \textbf{GT Source} & \textbf{Question Type} & \textbf{Task Type} \\
\midrule
EgoSchema~\cite{mangalam2023egoschema} & Ego4D & 5,063  & LLM+Annotator  & MCQ & Causality   \\
EgoPlan-Bench~\cite{chen2026egoplan} & Ego4D \& EPIC-Kitchens & 4,939 & Researchers & LLM+Annotator & MCQ & Planning  \\
EgoThink~\cite{cheng2024egothink} & Ego4D & 700  & Annotator & OE & G\&F\&P   \\
EgoTaskQA~\cite{jia2022egotaskqa} & LEMMA & 40,000  & Annotator + Program & OE & G\&F\&C  \\
EgoMemoria~\cite{ye2024mm} & Ego4D & 7,026  & LLM  & MCQ & Grounding  \\ 
HourVideo~\cite{chandrasegaran2024hourvideo} & Ego4D & 12,976 &  LLM+Annotator & MCQ & G\&F\&C  \\
EgoLifeQA~\cite{yang2025egolife} & EgoLife &3,000  & LLM+Annotator & MCQ & G\&C \\
\midrule
\textbf{EgoIntrospectQA} & EgoIntrospect & \textbf{4,060}  & \textbf{Subjects} & \textbf{Multiple*} & \textbf{A\&I\&M} \\
\bottomrule
\end{tabular}
\vspace{3pt}
\caption{Comparison of EgoIntrospectQA with related egocentric VQA benchmarks. We summarize key properties including data source, dataset scale, question source, ground-truth annotation, question type, and task type. Existing benchmarks mainly focus on fundamental reasoning tasks such as grounding (G), forecasting (F), planning (P), and causality (C). In contrast, our benchmark emphasizes understanding user internal states, including Affective Experience (A), Interaction Intent (I), and Cognitive Memory (M). Regarding question formats, prior work focuses on multiple-choice questions and open-ended questions. *:In contrast, our benchmark covers multiple formats, including MCQ, BC (binary classification) and MC (multi-class classification) }
\label{tab:benchmark_comparison}
\end{table}

\subsection{Comparison with Previous Works}

To highlight the unique contributions of our work, we contrast \textbf{EgoIntrospect} and \textbf{EgoIntrospectQA} with existing egocentric datasets and VQA benchmarks across several dimensions, as summarized in Table~\ref{tab:dataset_comparison} and Table~\ref{tab:benchmark_comparison}.

\subsubsection{Egocentric Dataset Characteristics}
Existing large-scale datasets, such as Ego4D~\cite{grauman2022ego4d} and EgoExo4D~\cite{Grauman_2024_CVPR}, have significantly advanced egocentric vision. However, as shown in Table~\ref{tab:dataset_comparison}, these datasets typically follow a \textit{researcher-defined} capture protocol and rely on \textit{external} annotators, which often limits the data to observable physical actions. 

In contrast, \textbf{EgoIntrospect} introduces a \textbf{subject-chosen} capture protocol where participants record continuous sessions (up to 240 minutes) reflecting their natural daily lives. Distinct from prior works like EgoLife~\cite{yang2025egolife}, our dataset is the first to integrate \textbf{Physiological Signals} (e.g., PPG, GSR) alongside video and gaze. Crucially, we employ \textbf{Self-annotation}, addressing the intrinsic ambiguity of "mind-reading" in egocentric vision. By obtaining ground truth directly from the subjects, we provide authentic labels for internal states that remain inaccessible to external observers.

\subsubsection{Egocentric VQA Benchmarks}
As illustrated in Table~\ref{tab:benchmark_comparison}, existing benchmarks like EgoSchema~\cite{mangalam2023egoschema} and HourVideo~\cite{chandrasegaran2024hourvideo} primarily evaluate external reasoning tasks such as spatial-temporal grounding (G), forecasting (F), and planning (P). Their questions are largely formulated by researchers or LLMs.

\textbf{EgoIntrospectQA} expands the scope of egocentric understanding to a multi-dimensional task suite (\textbf{A\&I\&M}). Our benchmark stands out in three aspects:
\begin{enumerate}
    \item \textbf{Subject-Sourced Intelligence:} Both questions and ground-truth are generated by the \textbf{subjects}, ensuring that tasks reflect genuine human cognitive processes rather than external heuristic guesses.
    \item \textbf{Internal-External Fusion:} We challenge MLLMs to infer internal states, including \textbf{Affective Experience (A)}, \textbf{Interaction Intent (I)}, and \textbf{Cognitive Memory (M)}. This allows for a more holistic assessment of agents in understanding both "what is happening" and "why the user feels/acts so."
    \item \textbf{Format Diversity:} Moving beyond standard MCQ aDnd OE formats, we incorporate \textbf{Multiple Formats} (*), including Binary Classification (BC) and Multi-class (MC) Classification (e.g., 9-way classification), providing a more granular evaluation of model performance across diverse task requirements.
\end{enumerate}
\subsection{Capture Devices}
To capture a comprehensive representation of user states and environmental interactions, we employed a multi-device wearable platform capable of recording diverse data modalities. The hardware configuration is designed to synchronize first-person visual perspectives with fine-grained physiological and motion signals.

As summarized in Table~\ref{tab:devices_modalities}, our primary visual and gaze data are captured using the \textbf{Pupil Labs Neon Glasses}, which provide high-resolution egocentric video alongside precise eye-tracking metrics. To monitor internal physiological responses, participants wore a \textbf{BioVital S2 Wristband} and a \textbf{Smart Ring}, providing redundant yet complementary signals such as photoplethysmography (PPG), electrodermal activity (EDA), and skin temperature. Furthermore, an \textbf{Apple Watch} was utilized to collect high-frequency inertial measurements (IMU) and environmental context (e.g., GPS, altitude, and ambient volume). 

This multi-modal setup allows for the joint analysis of external stimuli and internal states. All devices were time-synchronized to a common clock to ensure temporal alignment across video, motion, and physiological streams, providing a rich basis for understanding human intent and affective status in-the-wild.
\begin{table}[htbp]
\centering
\scriptsize
\begin{tabular}{p{3cm}p{10cm}}
\toprule
\textbf{Device} & \textbf{Recorded Data Modalities} \\
\midrule
Pupil Labs Neon Glasses~\cite{baumann2023neon} & First-person RGB video, gaze (position, azimuth), pupil data, IMU (accelerometer, gyroscope, quaternion), audio, recording events \\
\addlinespace
BioVital S2 Wristband & PPG (green 4-channel, red, infrared), heart rate, respiratory rate, EDA/GSR, SpO\textsubscript{2}, skin temperature, device temperature, ambient temperature and humidity, atmospheric pressure, IMU (accelerometer, gyroscope), 3-axis magnetometer \\
\addlinespace
Apple Watch & IMU (accelerometer, gyroscope, quaternion, gravity, magnetic field) at $\sim$30\,Hz; heart rate, step count, altitude, activity, GPS, volume at $\sim$1\,Hz; SpO\textsubscript{2}, respiratory rate, body temperature, HRV at $\sim$1/60\,Hz; \\
\addlinespace
Smart Ring~\cite{tang2025dataset} & PPG (green, red, infrared), IMU (accelerometer, gyroscope), temperature (3 channels) \\
\bottomrule
\end{tabular}
\caption{Devices and Recorded Data Modalities}
\label{tab:devices_modalities}
\end{table}

\subsection{Demographics}
Our dataset comprises a diverse group of 60 participants, with ages ranging from 15 to 48 years ($\mu = 23.63$, $Mdn = 23$). In terms of gender distribution, the cohort is relatively balanced, consisting of 33 females (55\%) and 27 males (45\%). The participants are internationally diverse, representing 14 different nationalities. While Chinese participants form the majority ($n=41$), the dataset also includes representation from Indonesia ($n=3$), Spain ($n=2$), Italy ($n=2$), Nepal ($n=2$), and one participant each from Bulgaria, France, Mexico, Kazakhstan, Ghana, and the Arabian region. This variety in age, gender, and cultural background ensures a broad spectrum of perspectives and behaviors within the collected data.

\begin{figure}[htbp]
    \centering
    \scriptsize
    
    \begin{minipage}{0.32\textwidth}
        \centering
        \begin{tikzpicture}[scale=0.7]
            \foreach \a/\b/\c in {0/198/mPink, 198/360/mBlue}
                \fill[\c, draw=white, thick] (0,0) -- (\a:2) arc (\a:\b:2) -- cycle;
            
            \node at (100:1.1) {\textbf{55\%}};
            \node at (280:1.1) {\textbf{45\%}};
            
            \begin{scope}[yshift=-2.6cm]
                \node at (0,0) {
                    \begin{tabular}{ll}
                    \textcolor{mPink}{$\blacksquare$} Female & \textcolor{mBlue}{$\blacksquare$} Male
                    \end{tabular}
                };
            \end{scope}
            \node[font=\bfseries] at (0,-3.4) {Gender};
        \end{tikzpicture}
    \end{minipage}
    \hfill
    \begin{minipage}{0.32\textwidth}
        \centering
        \begin{tikzpicture}[scale=0.7]
            \foreach \a/\b/\c in {0/42/mGreen, 42/258/mYellow, 258/342/mPurple, 342/360/mOrange}
                \fill[\c, draw=white, thick] (0,0) -- (\a:2) arc (\a:\b:2) -- cycle;
                
            \node at (21:2.5) {11.7\%};
            \node at (150:1.1) {\textbf{60\%}};
            \node at (300:1.1) {23\%};
            \node at (351:2.5) {5\%}; 
            
            \begin{scope}[yshift=-2.8cm]
                \node at (0,0) {
                    \begin{tabular}{ll}
                    \textcolor{mGreen}{$\blacksquare$} 15-19 & \textcolor{mYellow}{$\blacksquare$} 20-24 \\
                    \textcolor{mPurple}{$\blacksquare$} 25-29 & \textcolor{mOrange}{$\blacksquare$} 30+
                    \end{tabular}
                };
            \end{scope}
            \node[font=\bfseries] at (0,-4.0) {Age Group};
        \end{tikzpicture}
    \end{minipage}
    \hfill
    \begin{minipage}{0.32\textwidth}
        \centering
        \begin{tikzpicture}[scale=0.7]
            \begin{scope}[rotate=45]
                \foreach \a/\b/\c in {0/246/mYellow, 246/264/mBlue, 264/306/mGreen, 306/360/mPink}
                    \fill[\c, draw=white, thick] (0,0) -- (\a:2) arc (\a:\b:2) -- cycle;
                
                \node at (123:1.1) {\textbf{68\%}};
                \node[blue!70!black] at (255:2.5) {5\%}; 
                \node[green!50!black] at (285:2.5) {12\%}; 
                \node at (333:1.1) {15\%}; 
            \end{scope}
            
            \begin{scope}[yshift=-3.2cm]
                \node at (0,0) {
                    \begin{tabular}{ll}
                    \textcolor{mYellow}{$\blacksquare$} CN & \textcolor{mBlue}{$\blacksquare$} ID \\
                    \textcolor{mGreen}{$\blacksquare$} EU & \textcolor{mPink}{$\blacksquare$} Others
                    \end{tabular}
                };
            \end{scope}
            \node[font=\bfseries] at (0,-4.4) {Nationality};
        \end{tikzpicture}
    \end{minipage}

    \vspace{5.5em} 
    \caption{Demographic statistics of the 60 participants.}
    \label{fig:demographics}
\end{figure}
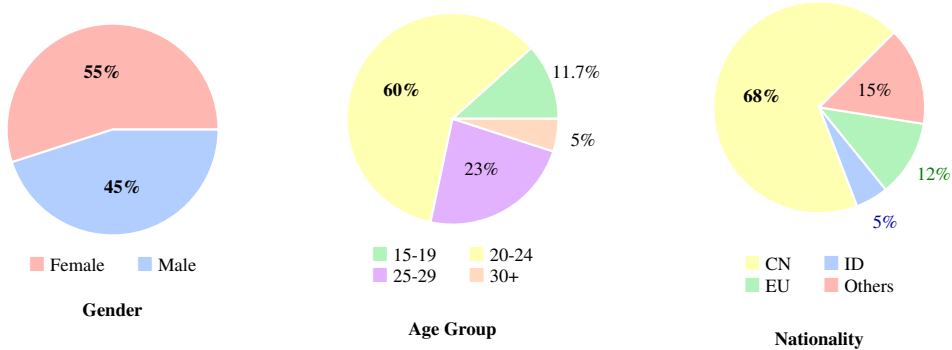

\subsection{Scenario Coverage}

 Rather than recording isolated tasks, each participant contributed a single, continuous long-form video session lasting between 2 to 4 hours. This continuous recording protocol was designed to capture the natural transitions between different environments and daily activities. Specifically, each long-form session covers 2 to 4 distinct scenario categories, resulting in a total of 192 annotated activity segments across 14 diverse categories. 

The distribution of these scenarios is detailed in Table~\ref{tab:scenario_stats} and illustrated in Fig.~\ref{Fig: scenario coverage}. Our dataset encompasses a broad spectrum of human activities, ranging from structured professional tasks to spontaneous social interactions and high-dynamic physical movements.

\begin{table}[htbp]
\centering
\caption{Statistics of Scenario Categories and Activity Examples.}
\label{tab:scenario_stats}
\resizebox{\textwidth}{!}{
\begin{tabular}{lp{7cm}c}
\toprule
\textbf{Scenario Category} & \textbf{Activity Examples} & \textbf{Segments} \\ \midrule
Social Interaction & Multi-person conversations, group activities, caregiving & 33 \\
Eating & Dining in canteens, restaurants, cafés, and bakeries & 27 \\
Leisure \& Special Events & Park walking, art exhibitions, dog walking, live performances & 22 \\
Shopping & Malls, department stores, supermarkets, and open markets & 16 \\
Sports & Light strength training, jogging, skateboarding & 14 \\
Transportation/Navigation & Navigating public transit, airports, and guide environments & 12 \\
Entertainment & Arcade centers, board game rooms, watching sports matches & 10 \\
Travelling \& Visiting & Cultural/historical sites and natural attractions & 10 \\
Personal Living (Home) & Cooking, household tidying, and deep cleaning & 9 \\
Tools, Assembly, Manufacture & Pottery, LEGO building, woodworking, and metalworking & 9 \\
Outdoors \& Gardening & Cycling, hiking, and gardening & 9 \\
Business/Service Processing & Parcel pickup, service counter interactions & 8 \\
Creative Arts \& Performance & Instrument practice, drawing, calligraphy, and knitting & 7 \\
Work and Study & Laboratory work, clinical shadowing, and electrical operations & 6 \\ \midrule
\textbf{Total} & & \textbf{192} \\ \bottomrule
\end{tabular}
}
\end{table}

The data collection process adhered to strict ethical and privacy guidelines. For scenarios involving high privacy sensitivity, such as \textit{Personal Living} or \textit{Social Interaction}, explicit consent for public data sharing was obtained from both the participants and their identifiable social partners. To ensure data quality and relevance for egocentric vision research, we intentionally avoided purely stationary desk-based activities (e.g., simple reading or computer work) and strictly prohibited the recording of copyrighted content. This collection strategy ensures a rich and ethically-compliant representation of dynamic, real-world human behavior.

\begin{figure}[htbp]
    \caption{Distribution of scenario categories across the 192 annotated activity segments. Each segment was extracted from 60 long-form egocentric video sessions, covering 14 diverse real-world contexts.}
    \label{Fig: scenario coverage}
    \centering
    \begin{tikzpicture}[scale=1.3]
        \foreach \a/\b/\c/\p in {
            0/61.9/mPink/17.2,           
            61.9/112.5/mBlue/14.1,        
            112.5/153.8/mGreen/11.5,      
            153.8/183.8/mYellow/8.3,      
            183.8/210.0/mPurple/7.3,      
            210.0/232.5/mOrange/6.3,      
            232.5/251.3/mMint/5.2,        
            251.3/270.0/mLavender/5.2,    
            270.0/286.9/mRose/4.7,        
            286.9/303.8/mPeach/4.7,       
            303.8/320.7/mLemon/4.7,       
            320.7/335.7/mSky/4.2,         
            335.7/348.8/mSage/3.6,        
            348.8/360.0/mGrey/3.1         
        } {
            \fill[\c, draw=white, thick] (0,0) -- (\a:3) arc (\a:\b:3) -- cycle;
            
            \node[font=\footnotesize\bfseries, text=black!70] at ({(\a+\b)/2}:2.1) {\p\%};
        }

        \begin{scope}[yshift=-5.2cm]
            \node[font=\small] at (0,0) {
                \begin{tabular}{ll}
                \textcolor{mPink}{$\blacksquare$} Social Interaction (33) & \textcolor{mLavender}{$\blacksquare$} Travelling (10) \\
                \textcolor{mBlue}{$\blacksquare$} Eating (27) & \textcolor{mRose}{$\blacksquare$} Personal Living (9) \\
                \textcolor{mGreen}{$\blacksquare$} Leisure \& Events (22) & \textcolor{mPeach}{$\blacksquare$} Tools \& Assembly (9) \\
                \textcolor{mYellow}{$\blacksquare$} Shopping (16) & \textcolor{mLemon}{$\blacksquare$} Outdoors \& Gardening (9) \\
                \textcolor{mPurple}{$\blacksquare$} Sports (14) & \textcolor{mSky}{$\blacksquare$} Business/Service (8) \\
                \textcolor{mOrange}{$\blacksquare$} Transportation (12) & \textcolor{mSage}{$\blacksquare$} Creative Arts (7) \\
                \textcolor{mMint}{$\blacksquare$} Entertainment (10) & \textcolor{mGrey}{$\blacksquare$} Work \& Study (6) \\
                \end{tabular}
            };
        \end{scope}

        \node[font=\bfseries] at (0,-6.8) {Distribution of 14 collection scenarios};
    \end{tikzpicture}
\end{figure}
\subsection{Dataset Split}

To evaluate model generalization while supporting future model development, we partition the dataset based on \textbf{unique participants}. This user-independent splitting strategy ensures that the test participants are never seen during model development, reducing the risk of participant-level data leakage.

Our current dataset contains 60 participants. We reserve a fixed set of 15 participants as the \textbf{Test Set} for final benchmark evaluation and cross-model comparison. The remaining 45 participants are used for development and are split into a \textbf{Training Set} and a \textbf{Validation Set} following an 8:2 ratio, resulting in 36 training participants and 9 validation participants. The training set is intended for model weight optimization, while the validation set is used for hyperparameter tuning, prompt selection, and model selection. All test participants are strictly excluded from the development phase.

As data collection is ongoing, newly collected participants will be assigned to the training and validation sets following the same 8:2 ratio, while the fixed test set will remain unchanged to provide a stable benchmark for future comparison. The current 60-participant split is summarized in Table~\ref{tab:data_split}.

\begin{table}[htbp]
\centering
\caption{Dataset partition based on the current 60 participants.}
\label{tab:data_split}
\begin{tabular}{lccc}
\toprule
\textbf{Split} & \textbf{Participants} & \textbf{Percentage} & \textbf{Role} \\ 
\midrule
Training       & 36 & 60\% & Model Development \\
Validation     & 9  & 15\% & Prompt/Model Selection \\
Test           & 15 & 25\% & Final Evaluation \\ 
\midrule
\textbf{Total} & \textbf{60}           & \textbf{100\%}      & -- \\ 
\bottomrule
\end{tabular}
\end{table}
\subsection{Benchmark Video Rendering}
\label{sec:benchmark_video_rendering}

\begin{figure}[htbp]
    \centering
    \begin{minipage}[t]{0.31\textwidth}
        \centering
        \includegraphics[width=0.92\linewidth,height=0.18\textheight,keepaspectratio]{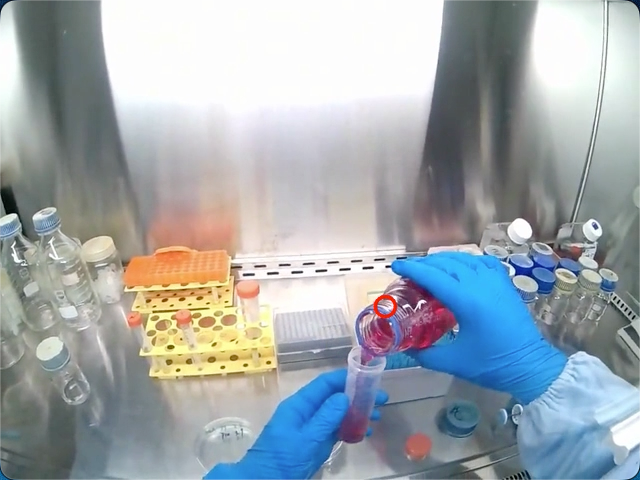}
        \vspace{2pt}
        \centerline{\small (a) Valid fixation}
    \end{minipage}
    \hfill
    \begin{minipage}[t]{0.31\textwidth}
        \centering
        \includegraphics[width=0.92\linewidth,height=0.18\textheight,keepaspectratio]{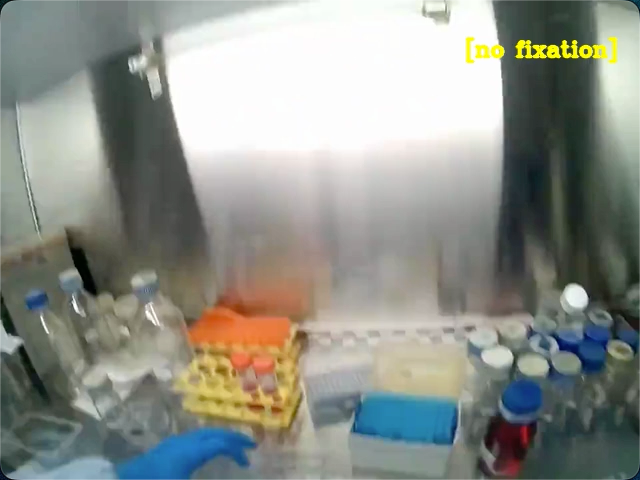}
        \vspace{2pt}
        \centerline{\small (b) No fixation}
    \end{minipage}
    \hfill
    \begin{minipage}[t]{0.31\textwidth}
        \centering
        \includegraphics[width=0.92\linewidth,height=0.18\textheight,keepaspectratio]{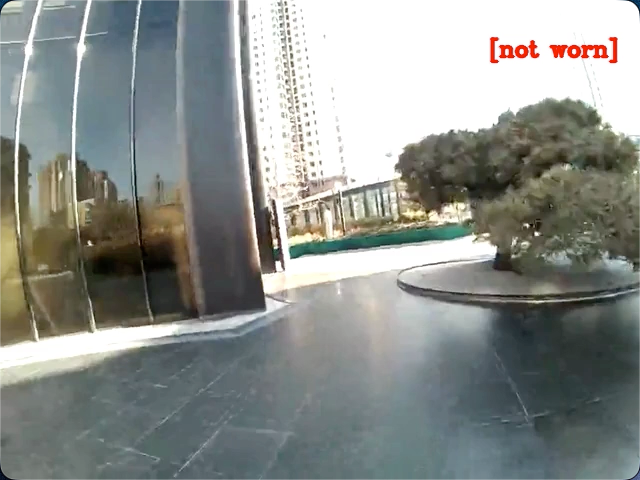}
        \vspace{2pt}
        \centerline{\small (c) Not worn}
    \end{minipage}
    \caption{Examples of gaze-state overlays in video-based benchmark inputs. The shared renderer exposes whether a frame contains an active fixation marker, has valid gaze tracking but no detected fixation, or corresponds to a period in which the glasses are not worn or gaze tracking is unavailable.}
    \label{fig:benchmark_fixation_overlay_states}
\end{figure}

All video-based benchmark tasks use a shared fixation-overlay renderer to standardize model inputs. For each task instance, the source egocentric recording is trimmed to the task-specific temporal window, compressed for inference, and augmented with visual cues that expose the state of the eye-tracking signal. As shown in Figure~\ref{fig:benchmark_fixation_overlay_states}, frames with an active fixation display a red circular marker at the fixation location. If gaze tracking is available but no fixation is detected at that frame, the video displays \texttt{[no fixation]}; if the gaze log indicates that the glasses are not worn, or that gaze tracking is unavailable, it displays \texttt{[not worn]}. This representation separates three visually similar cases---attended gaze, missing fixation, and unavailable gaze---so models can reason about the scene without treating absent fixation markers as implicit evidence of user attention. For evaluations that use overlaid video, the system prompt defines the red marker and both status labels before presenting the task-specific question.

\section{Dataset Open-Source}
\subsection{Open-Source Plan}

We plan to release our dataset in two complementary forms: benchmark data and the full dataset. The benchmark data will be publicly released to support direct reproduction of the benchmark results reported in this paper. This release will include all processed benchmark instances used in our evaluation, including the request-centered video clips, benchmark JSON files, task prompts, candidate answers, ground-truth labels, and evaluation scripts. The released benchmark data will be organized to work directly with our benchmark codebase, enabling future researchers to evaluate new models under the same task definitions and input settings.

In addition to the benchmark data, we will release the full dataset through a controlled-access process. The full dataset includes the continuous egocentric recordings, time-aligned multimodal signals, and participants' original retrospective annotations across all task categories. Since these recordings contain first-person video and may include bystanders, private spaces, screens, speech, and other sensitive contextual information, access to the full dataset will require a data-use agreement and an approval process. Similar to the access model used by large-scale egocentric video datasets such as Ego4D~\cite{grauman2022ego4d}, approved users will receive access credentials for downloading the controlled-access data.

Our current collection contains 60 participants. The benchmark results reported in this paper are computed on the 15-participant test set, while the remaining 45-participant training set will be released as part of the controlled-access full dataset. We plan to make the full dataset available around the conference release time. Since this project is designed as an ongoing data collection effort, we will continue to expand the controlled-access release with newly collected and processed data over time. Through this staged release plan, we aim to make the benchmark fully reproducible while preserving appropriate privacy protection for the broader first-person dataset.

\subsection{De-identification Process}

We will adopt a de-identification process similar to Ego4D~\cite{grauman2022ego4d}, combining automatic detection with manual review before releasing any public-access data. Since almost all of our recordings include outdoor or semi-public scenarios, the videos may contain bystanders who did not provide explicit consent for identity disclosure. Therefore, all public benchmark clips will be reviewed and de-identified before release.

The primary de-identification target is human faces. We will first apply an automatic face-detection and blurring pipeline to all video clips. After automatic processing, trained reviewers will manually inspect the results to correct missed detections and remove false positives. This manual verification step is necessary because first-person videos often contain challenging cases, such as motion blur, partial faces, side views, reflections, occlusions, or small faces in the background. Beyond faces, we will also manually review the videos for other sensitive information. In our data, we observed that participants may occasionally capture highly private content during daily activities, such as password entry, payment interfaces, addresses, or other personally identifiable information. These regions will be manually marked and blurred before public release. If a clip contains sensitive content that cannot be safely de-identified without substantially damaging the task content, it will be excluded from the public-access benchmark release.

We will also remove direct identifiers from released metadata, filenames, transcripts, and annotation files. Participant names and session identifiers will be replaced with randomized IDs, and any personally identifying text in transcripts or participant-written annotations will be removed or generalized. This process ensures that all publicly released benchmark data undergoes de-identification while preserving the information necessary for model evaluation.

\section{Benchmark Task1.1 - Moment of Recording}

The benchmark for "Moment of Recording" evaluates the ability of AI models to predict and interpret user-initiated recording actions in egocentric video data.

\subsection{Benchmark Question Definition and Curation Pipeline.}
 This task includes two primary question types:
\begin{enumerate}
    \item \textbf{The Recording Test}, which assesses whether a given video segment contains content that would prompt a user to initiate recording.
    \item \textbf{The Photo Test}, which identifies the specific object or scene that a user chose to photograph from a set of options.
\end{enumerate}

These questions probe the model's ability to infer user intent, visual appeal, and attentional focus from a first-person perspective.

The curation pipeline starts with an annotated egocentric video dataset, where human annotators identify moments when users initiate recording actions (e.g., taking photos). Each annotation includes the temporal boundaries of the recording event, a textual description of the user's intent, and the objects of interest in the scene. 

For the \textbf{Recording Test}, positive windows are constructed by extending 30 seconds before and after each user-initiated recording. This provides additional context to help the model understand the user's intent while ensuring that the recording windows are over one minute in length, as many participants tend to select recording moments that are just a few seconds long. Negative windows are sampled from non-overlapping video segments, and we include a greater number of negative windows to better simulate real-life scenarios, where users most often do not record anything. This also helps to avoid bias by ensuring that the model is not overly conditioned to expect a recording in every segment. The input consists of the scene context, requirements for the task, and the positive or negative segmented video. The output is a yes or no answer for whether recording should happen or not.

For the \textbf{Photo Test}, the pipeline also includes video segments where a photo was explicitly taken, being extended in a similar way with 30 seconds before and after each photo moment. The ground-truth target is determined by the annotated objects in the scene. To create a multiple-choice question, three additional distractor objects are selected randomly from a 10-second window around the exact moment the photo was taken. This encourages the model to evaluate which of the objects in the scene was most likely to have been selected for the photo. We also ensure that distractor objects are not repetitive across test groups, fostering a more challenging task for the model. The input consists of the scene context, requirements for the task, and the segmented video around the photo action, and 4 photos. The output is a choice of A/B/C/D corresponding ot the photo that the user should have recorded.

It is important to note that the video segments given to the models do not include audio in order to prevent the model from understanding if the recording was initiated through the voice command of the user.
\subsection{Benchmark Statistic.}
The performance of models in both the \textbf{Recording} and \textbf{Photo Tests} is primarily evaluated using accuracy, which measures the proportion of correct predictions to the total number of predictions.

\textbf{Recording Test:}
For the Recording Test, the model is tasked with predicting whether a given video segment contains content that would prompt a user to initiate recording. The expected output is a binary classification: "yes" if the content in the segment is likely to prompt recording, and "no" if it is not. The goal is to assess how accurately the model identifies moments in which a user might initiate recording based on the visual content.

\textbf{Photo Test:}
For the Photo Test, the model is asked to predict the object or scene that a user would choose to photograph from a set of options. The expected output is a multiple-choice selection of the correct object from a set of four candidate image crops extracted from the video. The goal is to evaluate the model’s accuracy in identifying the object the user intends to photograph.

\subsection{Benchmark Results.}
\label{app:1.1_result}
\paragraph{Dataset overview.}
The recording test (Subtask~1a) comprises \textbf{744} fixed-duration video windows
drawn from 15 participants: 253 positive windows (34.0\%), each centred
$\pm$30\,s around an annotated recording command (\texttt{record~a~video} or
\texttt{make~a~photo}), and 491 negative windows (66.0\%), randomly
sampled from non-recording intervals and matched in duration to their positive
counterparts.
These window counts are fixed and model-independent; all evaluated systems
face the same 744 instances.
Positive windows decompose into 189 originating from \texttt{make~a~photo}
commands and 63 from \texttt{record~a~video} commands; the latter carry
additional annotations for segment type
(Information Moment, Visual Aesthetic Curiousity, Task/Goal Oriented,
Social Interaction, Emotional Reaction),
attention level (Low / Normal / High),
and intention (Spontaneous / On-the-spot Plan / Pre Planned),
enabling a fine-grained diagnostic analysis.
The photo-identification test (Subtask~1b) contains \textbf{181} four-choice
multiple-choice questions, one per \texttt{make~a~photo} command, where the
model must identify the intended capture target among four candidate object
crops extracted from the same video segment.
Random chance performance is 25\%; the correct-answer distribution across
positions is balanced (A:~53, B:~38, C:~50, D:~40), ruling out positional bias.
Human annotators achieve \textbf{77\%} accuracy on the recording test and
\textbf{83\%} on the photo test.

\begin{table}[htbp]
\centering
\caption{Task 1.1 benchmark results across all evaluated models. Recording
test: 744 windows (binary yes/no). Photo test: 181 four-choice MCQ instances
(random baseline 25\%). All values in \%.}
\label{tab:task11_main}
\scriptsize
\setlength{\tabcolsep}{2pt}
\begin{tabularx}{\textwidth}{@{} l *{5}{Y} @{}}
\toprule
Models & \multicolumn{2}{c}{Recording Test} & \multicolumn{2}{c}{Photo Test} \\
\cmidrule(lr){2-3} \cmidrule(lr){4-5}
 & Acc. & Macro-F1 & Acc. & Macro-F1 \\
\midrule
\multicolumn{5}{@{}l}{Closed-source models} \\
Gemini-3-flash-preview          & 58.24 & 57.99 & \textbf{60.92} & 60.95 \\
Gemini-3.1-pro-preview          & 67.70 & 63.65 & 54.80 & 54.36 \\
GPT-4o                          & 68.15 & 63.53 & 24.86 & 25.98 \\
Kimi-K2.5                       & 63.58 & 62.83 & 29.28 & 21.83 \\
Kimi-K2.5(thinking mode)                       & 53.22 & 53.12 & 21.54 & 18.17 \\
Qwen3.5-397B-A17B               & 56.32 & 57.34 & 24.31 & 20.18 \\
Qwen3.5-Omni-Plus               & 56.85 & 57.11 & 26.52 & 20.42 \\
Qwen3.6-Plus                    & 56.59 & 56.66 & 24.86 & 21.26 \\
\midrule
\multicolumn{5}{@{}l}{Open-source models} \\
InternVL2.5-8B                  & 57.66 & 57.34 & 42.54 & 40.14 \\
InternVL3-8B                    & 65.46 & 62.37 & 46.41 & 44.26 \\
InternVL3.5-8B-Instruct         & \textbf{68.41} & 62.62 & 45.30 & 42.45 \\
InternVL3.5-8B-Thinking         & 58.47 & 57.37 & 43.09 & 43.18 \\
MiniCPM-V-4.5                   & 48.32 & 46.91 & 24.58 & 17.65 \\
Qwen2.5-VL-7B-Instruct          & 66.62 & 41.84 & 48.60 & 49.59 \\
Qwen3-Omni-30B-A3B-Instruct     & 52.49 & 52.70 & 60.89 & 59.86 \\
Qwen3-Omni-30B-A3B-Thinking     & 50.07 & 49.72 & 54.75 & 54.47 \\
Qwen3-VL-8B-Instruct            & 57.74 & 57.74 & 52.51 & 52.07 \\
Qwen3-VL-8B-Thinking            & 40.51 & 35.81 & 58.66 & 58.31 \\
LongVU-Qwen2-7B                 & 65.99 & 39.76 & 22.65 & 12.11 \\
LongVA-7B-DPO                   & 48.25 & 48.23 & 45.86 & 45.72 \\
LLaVA-OneVision-Qwen2-7B        & 58.60 & 57.53 & 28.73 & 11.56 \\
LLaVA-Video-7B-Qwen2            & 45.03 & 42.40 & 29.28 & 23.43 \\
\midrule
Human performance               & 77.00 & 65.00 & 83.00 & 91.00 \\
\bottomrule
\end{tabularx}
\end{table}

\subsubsection{Recording Test: GPT-4o vs.\ InternVL3.5-8B-Instruct Analysis}

We select GPT-4o and InternVL3.5-8B-Instruct for a detailed error analysis because InternVL3.5-8B-Instruct is the top-performing model overall (68.41\% accuracy), narrowly surpassing GPT-4o (68.15\%) while being a compact 8B open-source model.
The comparison reveals that comparable aggregate accuracy can mask meaningful differences in where and why each model succeeds.

\paragraph{Shared conservative bias, different precision-recall balance.}
Table~\ref{tab:task11_recording_cm} shows that both models adopt a conservative strategy, preferring to abstain from predicting a recording moment rather than risk a false alarm.
Both achieve specificity above 81\%, while recall remains below 46\%. GPT-4o has a slight edge in recall (45.6\% vs.\ 42.7\%), recovering six more true positives (114 vs.\ 108), whereas InternVL3.5 is marginally more conservative with higher specificity (81.7\% vs.\ 81.5\%) and fewer false alarms
(90 vs.\ 98). The result is a near-identical positive-class F1 (50.3\% vs.\ 47.9\%), confirming that despite the open-source size advantage, neither model has found a substantially better operating point for this task.

\begin{table}[h]
\centering
\caption{Recording test --- positive class diagnostic metrics for GPT-4o and  InternVL3.5-8B-Instruct (744 windows; 253 positive, 491 negative).}
\label{tab:task11_recording_cm}
\small
\begin{tabular}{lcccccc}
\toprule
\textbf{Model} & \textbf{Acc.} & \textbf{Prec.} & \textbf{Recall} &
\textbf{Spec.} & \textbf{F1\textsubscript{pos}} & \textbf{TP\,/\,FN} \\
\midrule
GPT-4o                 & 68.1 & 56.2 & 45.6 & 81.5 & 50.3 & 114\,/\,136 \\
InternVL3.5-8B-Instruct & 68.4 & 54.5 & 42.7 & 81.7 & 47.9 & 108\,/\,145 \\
\bottomrule
\end{tabular}
\end{table}

\paragraph{Command-kind gap.}
Table~\ref{tab:task11_kind} shows that both models follow the same pattern:
recall on \texttt{make~a~photo} windows is roughly twice that on
\texttt{record~a~video} windows.
This consistent gap across models suggests that photo-command moments carry a visually clearer, more localised trigger --- a salient foreground object highlighted by the gaze fixation --- whereas video-recording moments are often motivated by broader temporal or ambient context that a short clip alone does
not reliably convey. InternVL3.5 trails GPT-4o slightly on both command kinds (49.7\% vs.\ 51.3\%
on photo; 22.2\% vs.\ 27.0\% on video), indicating GPT-4o retains a narrow advantage in detecting these positive moments despite the overall accuracy parity.

\begin{table}[h]
\centering
\caption{Recording test --- recall (\%) on positive windows by command kind.}
\label{tab:task11_kind}
\small
\begin{tabular}{lccc}
\toprule
\textbf{Command kind} & \textbf{$n$} & \textbf{GPT-4o} & \textbf{InternVL3.5} \\
\midrule
\texttt{make a photo}   & 189 & 51.3 & 49.7 \\
\texttt{record a video} &  63 & 27.0 & 22.2 \\
\bottomrule
\end{tabular}
\end{table}

\paragraph{Complementary strengths across segment types and intentions.}
Despite their overall similarity, the two models show complementary
strengths when broken down by annotation category
(Table~\ref{tab:task11_breakdown}).
GPT-4o leads on Information Moment (42.9\% vs.\ 23.8\%), performing well when a legible object or informative text is centered in the field of view, but nearly fails on Task/Goal Oriented (7.1\%) and Visual Aesthetic Curiosity (14.3\%), types where recording intent has no single immediately detectable
visual anchor. InternVL3.5 exhibits the reverse strength: it is best on Task/Goal Oriented
(28.6\%) and more balanced across types (21--29\%), though uniformly lower than GPT-4o on Information Moment.
Neither model detects Emotional Reaction moments (both 0\%), confirming that affectively motivated recording is currently beyond the reach of both systems.

The intention dimension reveals \textbf{GPT-4o's most critical failure}: On-the-spot
Plan moments yield only 3.3\% recall (one out of 30), while Spontaneous recordings reach 58.8\%. This gap implies the model depends almost entirely on immediate visual triggers and is blind to short-horizon deliberation. InternVL3.5 avoids this extreme collapse (On-the-spot Plan: 16.7\%; Spontaneous: 23.5\%; Pre Planned: 36.4\%), though its uniformly lower recall
means it does not offer a practical advantage on any individual intention class.

\begin{table}[h]
\centering
\caption{Recording test --- recall (\%) by segment type and intention
         (\texttt{record~a~video} windows only, $n$ as labelled;
         G4o\,=\,GPT-4o, IVL\,=\,InternVL3.5-8B-Instruct).}
\label{tab:task11_breakdown}
\small
\begin{tabular}{lc rr | lc rr}
\toprule
\multicolumn{4}{c|}{\textbf{Segment type}} &
\multicolumn{4}{c}{\textbf{Intention}} \\
\textbf{Type} & $n$ & \textbf{G4o} & \textbf{IVL} &
\textbf{Intention} & $n$ & \textbf{G4o} & \textbf{IVL} \\
\midrule
Information Moment         & 21 & 42.9 & 23.8 & Spontaneous      & 17 & 58.8 & 23.5 \\
Task/Goal Oriented         & 14 &  7.1 & 28.6 & Pre Planned      & 11 & 27.3 & 36.4 \\
Vis.\ Aesthetic Curiousity & 14 & 14.3 & 21.4 & On-the-spot Plan & 30 &  3.3 & 16.7 \\
Social Interaction         &  6 & 33.3 & 16.7 & & & & \\
Emotional Reaction         &  3 &  0.0 &  0.0 & & & & \\
\bottomrule
\end{tabular}
\end{table}

\paragraph{Participant-level variance.}
Both models show extreme participant-level inconsistency with largely overlapping failure patterns.
GPT-4o achieves 95.0\% recall for P2 yet 0\% for P28, P32, P44, and P5. InternVL3.5 improves on several of GPT-4o's zero-recall participants (P13: 87.5\%, P33: 81.0\%, P52: 66.7\%) but introduces new failures of its own (P49: 0\%, P45: 8.3\%, P5: 11.1\%) and notably still fails entirely on P28. This large inter-participant variance ($\sigma_{\text{recall}} > 0.30$ for
both models) indicates that participant-specific scene characteristics --- recording context, environment, and individual behaviour patterns --- dominate over cross-participant model capability, and that current VLMs lack the
user-adaptive grounding required for consistent proactive recording.

\subsubsection{Photo Test: Insights and Discussion}

Subtask~1b asks the model to watch the egocentric video clip and select, from four candidate object crops, the specific object the user intended to photograph. Unlike the binary recording test, this task demands fine-grained understanding of where the user's attention is directed and what, among visually competing
objects, they considered capture-worthy.

\paragraph{Near-chance performance of most models.}
Table~\ref{tab:task11_main} reveals that the majority of closed-source models
cluster at or below the \textbf{25\% random baseline}. 
GPT-4o scores 24.86\%, and Kimi-K2.5, Qwen3.5-397B, Qwen3.5-Omni-Plus, and Qwen3.6-Plus all fall in the 24--29\% range. This near-chance performance, combined with GPT-4o producing 17 invalid responses out of 181 (9.4\%), suggests these models cannot reliably connect gaze-indicated attention to a specific object, or default to positional guessing when fine-grained visual grounding fails.

\paragraph{Reversed model ranking.}
The photo test induces a markedly different model ranking than the recording test. Gemini-3-flash-preview, the weakest closed-source model on recording (58.24\%), becomes the strongest overall on photo identification (60.92\%), matched only by Qwen3-Omni-30B-Instruct (60.89\%) and approaching the human ceiling of 83\%. Conversely, GPT-4o, the top performer on recording, drops to essentially random performance on the photo test.
This reversal suggests the two subtasks draw on complementary capabilities. Recording prediction benefits from holistic scene-level saliency assessment, while photo identification requires fine-grained, object-level visual grounding conditioned on inferred user intent.

\paragraph{Open-source models are competitive.}
On the photo test, multiple open-source models substantially outperform most
closed-source entries.
The InternVL family shows consistent improvement with architecture version: InternVL2.5-8B (42.54\%), InternVL3-8B (46.41\%), InternVL3.5-8B (45.30\%), all well above GPT-4o (24.86\%). Qwen3-VL-8B-Instruct (52.51\%), Qwen2.5-VL-7B-Instruct (48.60\%), and Qwen3-VL-8B-Thinking (58.66\%) further demonstrate that 7--8B open models can rival or exceed mid-range closed-source systems on this task. Thinking variants of Qwen3-VL and Qwen3-Omni consistently outperform their Instruct counterparts on the photo test (e.g., Qwen3-VL-Thinking 58.66\% vs.\ Instruct 52.51\%), consistent with the expectation that explicit chain-of-thought reasoning benefits the object-identification step.

\paragraph{Persistent human gap.}
Despite Gemini-3-flash reaching 60.92\%, a gap of over \textbf{22 percentage points} to human performance (83\%) remains.
The difficulty appears concentrated in cases where the target object is identified not by visual prominence alone but by the user's momentary intent, encoded in subtle gaze patterns, proximity, and contextual familiarity—factors that current VLMs cannot reliably infer from a short egocentric clip.

\subsubsection{Ablation Study}

\begin{table}[htbp]
    \centering
    \caption{Task~1.1 ablation over input conditioning for representative closed- and open-weight models (same decoding as Table~\ref{tab:task11_main}). Scores are percentages ($\times 100$). ICL denotes a one-shot in-context example in the benchmark prompt from the same participant, to act as their past interaction history. \textbf{Bold} marks the best score in each numeric column within the Closed-source/API and Open-source/local blocks.}
    \scriptsize
    \setlength{\tabcolsep}{3pt}
    \setlength{\tabcolsep}{2pt}
    \begin{tabularx}{\textwidth}{@{} l >{\raggedright\arraybackslash}X *{6}{Y} @{}} 
        \toprule
        Model                                 & Inputs                          & \multicolumn{2}{c}{Recording} & \multicolumn{2}{c}{Photo}                                                                     \\
        \cmidrule(lr){3-4} \cmidrule(lr){5-6}
                                              &                                 & acc                     & macro f1                & acc            & macro f1       \\
        \midrule
        \multicolumn{6}{@{}l}{\textit{Closed-source/API models}}                                                                                                                                        \\
        \multirow{4}{*}{kimi-k2.5}            & video                           & 44.22                   & 51.21                   & \textbf{32.59}          & \textbf{23.52}            \\
                                              & video + gaze                    & 63.57                   & 62.83                   & 29.28          & 21.83           \\
                                              & video + gaze + ICL & \textbf{70.00}          & \textbf{66.16}          & 20.48 & 20.30     \\
        \midrule
        \multicolumn{6}{@{}l}{\textit{Open-source/local models}}                                                                                                                                        \\
        \multirow{4}{*}{Qwen3-VL-8B-Instruct} & video                           & 52.01                   & 51.85                   & \textbf{55.42}          & \textbf{55.15}           \\
                                              & video + gaze                    & \textbf{57.73}                   & \textbf{57.73}                   & 52.51          & 52.07                  \\
                                              & video + gaze + ICL & 52.01          & 51.18                   & 52.97 & 52.40           \\
        \bottomrule
    \end{tabularx}
    \label{tab:task11_ablation}
\end{table}
We conduct two ablation conditions—gaze removal and in-context learning (ICL)—on both subtasks using Kimi-K2.5 and Qwen3-VL-8B-Instruct as representative models.
In the no-gaze condition, the fixation overlay is omitted, and models receive the raw video without any gaze indicator.
In the ICL condition, a single task-relevant demonstration (question--answer pair) is prepended to the prompt; samples without a pre-assigned example are excluded, leaving 730 recording windows and 166 photo samples.
Table~\ref{tab:abl} summarizes all conditions.

\begin{table}[h]
\centering
\caption{Ablation results for Kimi-K2.5 and Qwen3-VL-8B-Instruct (accuracy / macro-F1, \%).
         Baseline = standard gaze overlay, no ICL.}
\label{tab:abl}
\small
\begin{tabular}{l cc cc}
\toprule
& \multicolumn{2}{c}{\textbf{Recording test}} & \multicolumn{2}{c}{\textbf{Photo test}} \\
\cmidrule(lr){2-3} \cmidrule(lr){4-5}
\textbf{Condition} & \textbf{Kimi} & \textbf{Qwen3-VL} & \textbf{Kimi} & \textbf{Qwen3-VL} \\
\midrule
Baseline (w/ gaze) & 63.57\,/\,62.83 & 57.73\,/\,57.73 & 29.28\,/\,21.83 & 52.51\,/\,52.07 \\
No gaze            & 44.22\,/\,51.21 & 52.01\,/\,51.85 & 32.59\,/\,23.52 & 55.42\,/\,55.15 \\
ICL (one-shot)     & 70.00\,/\,66.16 & 52.01\,/\,51.18 & 20.48\,/\,20.30 & 52.97\,/\,52.40 \\
\bottomrule
\end{tabular}
\end{table}

\paragraph{Gaze is a critical signal for the recording test.}
Removing the gaze overlay causes Kimi-K2.5 to drop \textbf{19.4 points in accuracy} (63.57\% to 44.22\%). Qwen3-VL-8B-Instruct is less sensitive, losing 5.7 points (57.73\% to 52.02\%), suggesting it relies more on holistic video content and less on the fixation signal.
Overall, the gaze overlay provides a substantial grounding benefit for the recording task, particularly for models that attend closely to the attentional cue.

\paragraph{Gaze slightly hurts photo identification.}
Counterintuitively, removing gaze modestly improves photo-test accuracy for both models: Kimi gains 3.3 points (29.28\% to 32.59\%) and Qwen3-VL gains 2.9 points (52.51\% to 55.42\%).
A likely explanation is that the circular fixation marker introduces visual clutter that interferes with fine-grained object matching against the four candidate crops, whereas the raw frame provides a cleaner signal for visual comparison. This finding suggests that the optimal gaze representation may differ between the two subtasks.

\paragraph{ICL helps Kimi on recording but hurts on the photo test.}
One-shot ICL raises Kimi's recording accuracy by \textbf{6.4 points} (63.57\% to 70.00\%) and improves macro-F1 by 3.3 points, indicating that a concrete example of the yes/no judgement meaningfully guides Kimi's decision boundary.
For Qwen3-VL, ICL provides no benefit on recording (57.73\% to 52.02\%), suggesting the model's baseline instruction-following is already well-calibrated for this task, and the example may introduce a confounding bias. On the photo test, ICL substantially hurts Kimi (29.28\% to 20.48\%, a drop of
\textbf{8.8 points}), potentially because the single-shot example biases the model toward a particular answer format or object category that misaligns with most test instances.
Qwen3-VL remains stable on the photo test under ICL (52.51\% to 52.97\%), showing no sensitivity to the demonstration in either direction.

\section{Benchmark Task1.2 - Emotion Analysis}
\subsection{Benchmark Question Definition and Curation Pipeline.}
We construct the Task~1.2 benchmark from participant-annotated emotion segments and instantiate three question types that probe affect understanding at progressively increasing levels of difficulty: emotion-label prediction (EI-label), emotion-label prediction with reason grounding (EI-label+reason), and relative emotion-intensity comparison (EIR).

\textbf{EI-label.} EI-label evaluates whether a model can identify the participant's emotion category from a single egocentric video segment. Let $\mathcal{E}=\{e_1,\dots,e_9\}$ denote the predefined emotion taxonomy, consisting of \textit{Joy}, \textit{Surprise}, \textit{Stress}, \textit{Social Connection}, \textit{Achievement}, \textit{Disappointment}, \textit{Awkward}, \textit{Boring}, and \textit{Relaxation}. For an annotated segment with ground-truth label $e_i^*$, we define the candidate set as $\mathcal{C}_i=\mathcal{E}$ and require the model to select one option $\hat{y}_i \in \mathcal{C}_i$ given the multimodal context $\mathcal{X}_i$. Although the original annotation protocol allows an \textit{Other} field, we adopt a fixed-label benchmark design and retain only segments that can be assigned to one of these nine categories.

\textbf{EI-label+reason.} EI-label+reason extends EI-label by requiring the model to jointly predict the participant's emotion and the reason underlying that emotion. For the same segment, let the correct option be $c_i^+=(e_i^*, r_i^*)$, where $e_i^*$ is the participant-annotated emotion label and $r_i^*$ is the corresponding reason description. We then construct a nine-way candidate set $\mathcal{C}_i=\{c_i^+, c_{i,1}^-, \dots, c_{i,8}^-\}$, where each distractor $c_{i,j}^-=(e_j, r_{i,j}^-)$ pairs an alternative emotion category with a Gemini-3.1-pro-generated reason conditioned on the video content, the participant's annotated reason, and the target emotion category. The model is required to select one option $\hat{y}_i \in \mathcal{C}_i$. This formulation is more demanding than category prediction alone, as it requires the model to align the observed scene with both the reported affect and its attributed cause.

\textbf{EIR.} Emotion Intensity Recognition evaluates whether a model can compare the relative strength of emotional experiences rather than merely identify their category. For a target participant and emotion category, we construct a four-choice candidate set $\mathcal{C}_i = \{c_{i,1}, c_{i,2}, c_{i,3}, c_{i,4}\}$, where each candidate $c_{i,j}=(\mathcal{X}_{i,j}, e_{i,j}^*, r_{i,j}^*)$ contains the available multimodal context, the annotated emotion label, and the corresponding reason description for one clip. The model is then required to identify either the strongest or the weakest emotional experience among the four candidates, with the ground-truth targets defined by the participant-rated intensity scores:
\[
    y_i^{\max} = \arg\max_{j \in \{1,\dots,4\}} e_{i,j}^{I,*}, \qquad
    y_i^{\min} = \arg\min_{j \in \{1,\dots,4\}} e_{i,j}^{I,*}.
\]
When fewer than four valid clips are available for a target participant-emotion group, the remaining slots are filled with clips from the same participant that do not belong to the target emotion and do not share the same reason description, thereby preserving within-user comparison while reducing shortcut cues from duplicated causes.

\subsection{Benchmark Statistic.}
Task~1.2 contains three benchmark variants constructed from 15 participants. The two emotion-identification variants, EI-label and EI-label+reason, share the same 281 benchmark instances and differ only in the candidate answer format, while EIR contains 88 four-choice intensity-comparison questions. To ensure consistent temporal context, all benchmark clips are uniformly cropped to 180 seconds. The EI benchmark spans all nine predefined emotion categories, with a moderately imbalanced label distribution: Joy (98), Surprise (64), Stress (34), Social Connection (28), Relaxation (17), Awkward (11), Disappointment (11), Achievement (9), and Boring (9). EIR is balanced across the two question variants, containing 44 highest-intensity questions and 44 lowest-intensity questions.

\begin{table}[htbp]
    \centering
    \caption{Task1.2 benchmark statistics.}
    \scriptsize
    \setlength{\tabcolsep}{2pt}
    \begin{tabularx}{\textwidth}{@{} l *{4}{Y} @{}}
        \toprule
        Question Type & \# Instances & \# Participants & Answer Space & Chance \\
        \midrule
        EI-label & 281 & 15 & 9 labels & 11.11 \\
        EI-label+reason & 281 & 15 & 9 emotion-reason pairs & 11.11 \\
        EIR & 88 & 15 & 4 candidates & 25.00 \\
        \bottomrule
    \end{tabularx}
    \label{tab:task12_statistics}
\end{table}

\subsection{Benchmark Results.}
\subsubsection{Baseline Results}
\textbf{Evaluation protocol.} We evaluate the same set of models across all three benchmark variants. For the main benchmark setting, EI-label and EI-label+reason are evaluated under the full multimodal input condition with video, gaze, transcript, and physiological signals, while EIR is evaluated under video, gaze, and transcript inputs. Human performance is measured on a separate subset comprising roughly 10\% of the benchmark data using the same input formats and answer constraints as the model evaluation. We report accuracy for all three variants and additionally report macro-F1 to account for label imbalance and option-level bias in the multiple-choice outputs.

\begin{table}[htbp]
\centering
\caption{Task~1.2 baseline benchmark results. Scores are percentages ($\times 100$).}
\label{tab:task12_results}
\scriptsize
\setlength{\tabcolsep}{2pt}
\begin{tabularx}{\textwidth}{@{} l *{6}{Y} @{}}
\toprule
Models & \multicolumn{2}{c}{EI-label} & \multicolumn{2}{c}{EI-label+reason} & \multicolumn{2}{c}{EIR} \\
\cmidrule(lr){2-3}\cmidrule(lr){4-5}\cmidrule(lr){6-7}
 & acc & macro f1 & acc & macro f1 & acc & macro f1 \\
\midrule
\multicolumn{7}{@{}l}{\textit{Closed-source models}} \\
Gemini-3-flash-preview          & \textbf{62.28} & 61.92 & 62.63 & 60.21 & \textbf{55.68} & \textbf{51.34} \\
Gemini-3.1-pro-preview          & 58.71 & 59.84 & \textbf{66.90} & \textbf{65.37} & 54.55 & 48.92 \\
gpt-4o                          & 59.79 & 58.66 & 63.35 & 62.47 & 50.00 & 46.87 \\
kimi-k2.5                       & 53.38 & 52.11 & 54.09 & 51.83 & 39.77 & 41.26 \\
kimi-k2.5(thinking mode)        & 54.40 & 54.92 & 53.74 & 52.95 & 47.73 & 44.18 \\
qwen3.5-397b-a17b               & 51.60 & 50.28 & 52.67 & 49.96 & 44.32 & 42.07 \\
qwen3.5-omni-plus               & 54.09 & 53.41 & 54.44 & 53.18 & 46.60 & 43.95 \\ \midrule
\multicolumn{7}{@{}l}{\textit{Open-source models}} \\
InternVL2\_5-8B                 & 27.41 & 26.58 & 29.97 & 28.91 & \textbf{43.18} & 36.27 \\
InternVL3-8B                    & 22.42 & 21.36 & 27.47 & 26.84 & 34.09 & 33.18 \\
InternVL3\_5-8B-Instruct        & 25.98 & 24.73 & 35.68 & 34.92 & 39.77 & 35.64 \\
InternVL3\_5-8B-Thinking        & 29.54 & 28.11 & 32.47 & 31.08 & 36.36 & 34.05 \\
MiniCPM-V-4\_5                  & 14.95 & 13.42 & 38.18 & 37.25 & 38.64 & 34.72 \\
Qwen2.5-VL-7B-Instruct          & 38.08 & 36.47 & 31.39 & 30.14 & 29.55 & 32.46 \\
Qwen3-Omni-30B-A3B-Instruct     & 30.25 & 29.06 & 37.47 & 36.55 & 37.50 & 35.11 \\
Qwen3-Omni-30B-A3B-Thinking     & 36.30 & 35.48 & 35.68 & 34.87 & 38.64 & 36.02 \\
Qwen3-VL-8B-Instruct            & 30.25 & 28.94 & 31.75 & 30.66 & 42.05 & \textbf{37.58} \\
Qwen3-VL-8B-Thinking            & 37.37 & 36.02 & 33.54 & 32.41 & 32.95 & 33.47 \\
LongVU\_Qwen2\_7B               & 39.50 & 37.26 & \textbf{43.06} & \textbf{41.72} & 31.82 & 30.84 \\
LongVA-7B-DPO                   & \textbf{42.00} & \textbf{40.33} & 40.21 & 38.96 & 32.95 & 32.27 \\
llava-onevision-qwen2-7b-ov-hf  & 19.22 & 17.88 & 24.97 & 23.41 & 40.91 & 34.93 \\
LLaVA-Video-7B-Qwen2            & 28.47 & 27.16 & 28.83 & 27.95 & 35.22 & 31.76 \\
\midrule
human performance               & 78.57 & 74.80 & 89.29 & 86.10 & 60.00 & 55.20 \\
\bottomrule
\end{tabularx}
\end{table}

Table~\ref{tab:task12_results} shows a consistent closed-source advantage across all three Task~1.2 variants. On EI-label, Gemini-3-flash-preview achieves the best score ($62.28$ accuracy, $61.92$ macro-F1), while the strongest open-source result is LongVA-7B-DPO ($42.00$, $40.33$). On EI-label+reason, Gemini-3.1-pro-preview is strongest ($66.90$, $65.37$), whereas the best open-source score is LongVU\_Qwen2\_7B ($43.06$, $41.72$). The enlarged gap in EI-label+reason indicates that jointly modeling emotion and explanation remains substantially harder than label-only prediction for open-weight models.

For EIR, Gemini-3-flash-preview remains the top performer among closed-source models ($55.68$, $51.34$). The best open-source EIR accuracy is obtained by InternVL2\_5-8B ($43.18$), while the best open-source EIR macro-F1 is achieved by Qwen3-VL-8B-Instruct ($37.58$). Compared with EI tasks, EIR macro-F1 is generally lower, reflecting the additional difficulty of relative intensity comparison over multiple clips. Overall, Task~1.2 suggests a persistent capability gap in user-centric affect reasoning, especially when moving from emotion labeling to reason-grounded and intensity-comparison settings.

\subsubsection{Ablation Results}
Table~\ref{tab:task12_ablation} reports ablation results for representative closed-source and open-source models. For EI-label and EI-label+reason, we compare four input settings: \textit{video}, \textit{video+gaze}, \textit{video+gaze+transcript}, and \textit{video+gaze+transcript+physiology}. For EIR, we compare \textit{video}, \textit{video+gaze}, \textit{video+gaze+transcript}, and \textit{video+gaze+transcript+ICL}.

\begin{table}[htbp]
\centering
\caption{Task~1.2 ablation over input conditioning. Scores are percentages ($\times 100$).}
\label{tab:task12_ablation}
\scriptsize
\setlength{\tabcolsep}{2pt}
\begin{tabularx}{\textwidth}{@{} l >{\raggedright\arraybackslash\hsize=2\hsize}X *{6}{>{\centering\arraybackslash\hsize=\dimexpr 5\hsize/6\relax}X} @{}}
\toprule
Model & Inputs & \multicolumn{2}{c}{EI-label} & \multicolumn{2}{c}{EI-label+reason} & \multicolumn{2}{c}{EIR} \\
\cmidrule(lr){3-4}\cmidrule(lr){5-6}\cmidrule(lr){7-8}
 &  & acc & macro f1 & acc & macro f1 & acc & macro f1 \\
\midrule
\multicolumn{8}{@{}l}{\textit{Closed-source/API models}} \\
kimi-k2.5 & video & 53.38 & 52.11 & 54.09 & 51.83 & 39.77 & 41.26 \\
kimi-k2.5 & video + gaze & 52.67 & 50.94 & 53.38 & 50.72 & 38.64 & 39.88 \\
kimi-k2.5 & video + gaze + transcript & \textbf{56.58} & \textbf{55.31} & \textbf{58.01} & \textbf{56.27} & 43.18 & 43.62 \\
kimi-k2.5 & EI: V+G+T+P / EIR: V+G+T+ICL & 56.23 & 54.68 & 57.65 & 55.94 & \textbf{44.32} & \textbf{44.05} \\
\midrule
\multicolumn{8}{@{}l}{\textit{Open-source/local models}} \\
Qwen3-VL-8B-Instruct & video & 30.25 & 28.94 & 31.67 & 30.66 & 42.05 & 37.58 \\
Qwen3-VL-8B-Instruct & video + gaze & 31.32 & 29.37 & 31.32 & 30.11 & 40.91 & 36.92 \\
Qwen3-VL-8B-Instruct & video + gaze + transcript & 35.59 & 33.08 & \textbf{36.65} & \textbf{35.24} & \textbf{47.73} & \textbf{40.17} \\
Qwen3-VL-8B-Instruct & EI: V+G+T+P / EIR: V+G+T+ICL & \textbf{36.30} & \textbf{33.42} & 36.30 & 34.96 & 46.59 & 39.48 \\
\bottomrule
\end{tabularx}
\end{table}

The ablation results show that transcript is the most consistently useful additional signal for both representative models, while gaze-only input can be neutral or slightly harmful. For kimi-k2.5, adding gaze to video reduces EI-label ($53.38/52.11 \rightarrow 52.67/50.94$) and EI-label+reason ($54.09/51.83 \rightarrow 53.38/50.72$), whereas adding transcript yields the largest gains (EI-label $56.58/55.31$, EI-label+reason $58.01/56.27$). The final EI setting with physiology does not further improve over the transcript setting ($56.23/54.68$ and $57.65/55.94$), indicating limited incremental benefit from physiology in this configuration.

Qwen3-VL-8B-Instruct shows a similar but not identical pattern. Gaze slightly improves EI-label ($30.25/28.94 \rightarrow 31.32/29.37$) but slightly hurts EI-label+reason and EIR ($31.67/30.66 \rightarrow 31.32/30.11$, $42.05/37.58 \rightarrow 40.91/36.92$). Transcript provides the largest gain for EI-label+reason and EIR ($36.65/35.24$ and $47.73/40.17$), while the final EI setting with physiology further improves EI-label to $36.30/33.42$. By contrast, the same final row slightly underperforms transcript-only on EI-label+reason and EIR ($36.30/34.96$, $46.59/39.48$), indicating that physiology and ICL provide task-dependent gains rather than uniformly improving all subtasks.

\section{Benchmark Task2.1 - User-Initiated Contextual Request}
\subsection{Benchmark Question Definition and Curation Pipeline.}
We constructed the benchmark for User-Initiated Contextual Request from the participant-verified request annotations described above. Each annotated request segment contains the corrected temporal boundary, the verified spoken request $u_i^*$, the participant-provided expected response requirement $\rho_i^*$, and optional visual referent annotations. Based on these annotations, we constructed two question types: \textit{Tool Selection (TS)}, which evaluates whether the model can identify the external capabilities needed to fulfill a request, and \textit{Request Prediction (RP)}, which evaluates whether the model can infer the user's intended request from the surrounding egocentric context.

\subsubsection{Tool Selection (TS)}

\textbf{Stage1: Initial tool requirement annotation.}
For each participant-verified request segment, we used an LLM to infer which external tools would be needed to satisfy the request. The input to this step included the spoken request $u_i^*$, the expected response requirement $\rho_i^*$, the segment timestamps, and any associated visual referent annotations. The LLM produced two tool lists: \textit{required tools}, which are necessary to answer the request, and \textit{helpful tools}, which may improve the response but are not strictly required. For each selected tool, the LLM also generated example tool calls, including function names and query arguments.

\textbf{Stage2: Tool name normalization and tool-universe construction.}
Since each request was processed independently, semantically equivalent tools could be assigned different surface names, such as \texttt{map\_search}, \texttt{maps\_api}, or \texttt{location\_search}. We therefore aggregated all tool names across participants and used an LLM to propose a normalization mapping from raw tool names to canonical tool names. This mapping was then manually reviewed by the researchers to correct two types of errors: over-merging, where functionally different tools were mapped to the same canonical name, and under-merging, where equivalent tools remained separated.

After normalization, we rebuilt the final tool universe $\mathcal{T}$ using the canonical tool names, consisting of 33 tools in total. For each tool, we summarized its usage examples and generated a short natural-language description. We then updated each participant's tool annotations by replacing raw tool names with canonical names. When a tool name changed during normalization, we regenerated its corresponding example calls so that the function names and query arguments remained consistent with the canonical tool semantics.

\textbf{Stage3: Video clip, transcripts, and benchmark instance construction.}
For each request segment, we extracted an egocentric video clip spanning 30 seconds before the request onset to 30 seconds after the request offset. We overlaid gaze fixations as red circles and added an ``Asking Question'' label during the time interval in which the participant was speaking the request. For participants with multiple recording sessions, we first processed and concatenated the session videos with gaze overlays, then extracted request-centered clips from the merged video.

Each \textit{TS} instance presents the model with the request-centered video clip and a complete list of available tools. The candidate set consists of all canonical tools in $\mathcal{T}$ plus a ``No tool needed'' option. Unlike a four-option multiple-choice question, \textit{TS} is formulated as a multi-select task: the model must output all tools needed to fulfill the request. We store the normalized required-tool list and helpful-tool list as ground truth. To support models with different input capabilities, we construct two prompt versions: one using the original user question text, and another that additionally includes an automatic transcript of the full video clip. We further transcribed the full request-centered clip, including both the spoken request and the audio context, using automatic speech recognition: Xunfei LFASR~\cite{iflytek_asr_llm_2026} for Chinese recordings and ElevenLabs~\cite{elevenlabs_speech_to_text} for English recordings.

\textbf{Correctness Criterion for Tool Selection.} Because each TS instance contains both \textit{required} and \textit{helpful} tool annotations, we use a hierarchical correctness criterion rather than a simple exact-match rule. For each instance $i$, let $R_i$ denote the set of required tools, $H_i$ denote the set of helpful-but-not-required tools, $\hat{P}_i$ denote the set of tools predicted by the model, and $n$ denote the special option \textit{No tool needed}. We define the strict correctness indicator as:\[\mathrm{Correct}(i)=\begin{cases}1, & \text{if } R_i \neq \emptyset \text{ and } R_i \subseteq \hat{P}_i \subseteq R_i \cup H_i, \\1, & \text{if } R_i=\emptyset, H_i \neq \emptyset, \text{ and } \left(\hat{P}_i=\{n\} \text{ or } \emptyset \neq \hat{P}_i \subseteq H_i \right), \\1, & \text{if } R_i=\emptyset, H_i=\emptyset, \text{ and } \hat{P}_i=\{n\}, \\0, & \text{otherwise.}\end{cases}\]Thus, the model must select all required tools and may additionally select helpful tools, but any irrelevant tool makes the answer incorrect. 

To further characterize model behavior beyond strict accuracy, we report two auxiliary metrics for \textit{TS}. \textit{Required Recall} measures whether the model identifies all tools that are strictly necessary for fulfilling the request. It is computed only on instances with non-empty required-tool annotations and does not penalize the model for selecting additional helpful or irrelevant tools. This metric captures whether the model recognizes the core external capabilities needed by the request.

We also report \textit{Distractor Rate}, which measures how often the model selects tools outside the annotated acceptable set. For each instance, we treat the union of required and helpful tools as the set of acceptable tools, while \textit{No tool needed} is treated as a special option rather than a real tool. A prediction is counted as containing a distractor if it includes any real tool that is neither required nor helpful. This metric reflects the model's tendency to over-call irrelevant tools.

\subsubsection{Request Prediction (RP)}

\textbf{Stage1: Partially muted video clip construction.}
RP evaluates whether the model can infer what the participant intended to ask, rather than simply recognizing the spoken words. We therefore used the same request-centered video construction as TS, including the $\pm30$ second temporal context, gaze fixation overlays, and the ``Asking Question'' label. However, for RP we muted the audio during the request interval $[t_0,t_1]$, so that models with audio understanding could not directly hear the answer. Audio outside the request interval was preserved, since surrounding environmental sounds or conversations may provide useful context.

\textbf{Stage2: Candidate request generation.}
For each muted request clip, we used Gemini to generate a four-option candidate set. The model was given the muted video together with the participant annotation, including the original request $u_i^*$, the expected response requirement $\rho_i^*$, and any visual referent annotations. Gemini then generated one \textit{correct request}, which reformulates the participant's actual request and expected response requirement into a complete request description, and three \textit{wrong requests}, which are plausible in the same egocentric context but do not match the participant's true intent. The original request text was used only to help generate high-quality options and is not exposed in the final benchmark prompt.

\textbf{Stage3: Benchmark instance construction.}
We randomly shuffled the correct request and three distractor requests using a fixed seed, assigned them to options A--D. We generated transcripts for the muted clips using automatic speech recognition. Because the request interval itself was muted, the transcript only captures the surrounding audio context rather than the participant's spoken request. Each \textit{RP} instance asks the model to select the option that best matches the participant's actual request. The user prompt contains only the four candidate requests, and does not include the original spoken request. As with \textit{TS}, we provide two benchmark versions: one without transcript and one with a transcript block containing only the surrounding audio context.

\subsection{Benchmark Statistics.}

Task 2.1 contains 587 benchmark instances from 16 participants, including 298 \textit{Tool Selection (TS)} instances and 289 \textit{Request Prediction (RP)} instances. Each instance is paired with a request-centered egocentric video clip, with a median clip duration of 66.55 seconds and a median spoken-query duration of 6.55 seconds. All records passed our basic quality checks, with no missing fields, invalid fields, or duplicated record keys.

\begin{table}[t]
\centering
\small
\begin{tabular}{lcc}
\toprule
Question Type & \# Instances & Key Statistics \\
\midrule
\textit{Tool Selection (TS)} & 298 & 33 tools; 96 no-tool cases; 120 multi-tool cases \\
\textit{Request Prediction (RP)} & 289 & 4 options per instance; all clips muted during the query interval \\
\bottomrule
\end{tabular}
\caption{Summary statistics of the Task 2.1 benchmark.}
\label{tab:task21_statistics}
\end{table}

\subsection{Benchmark Results.}

\begin{table}[htbp]
\centering
\caption{Benchmark Results for Request Intent - Task 2.1}
\scriptsize 
\label{tab:append_benchmark_results_task2.1}
\setlength{\tabcolsep}{3pt} 
\setlength{\tabcolsep}{2pt} 
\begin{tabularx}{\textwidth}{@{} l *{5}{Y} @{}} 
\toprule
Models & \multicolumn{3}{c}{Tool Selection} & \multicolumn{2}{c}{Request Prediction}  \\ 
\cmidrule(lr){2-4} \cmidrule(lr){5-6} 
 & acc & required recall & distractor rate & acc & macro-F1  \\ 

\midrule

\multicolumn{6}{@{}l}{\textit{Closed-source models}} \\ 
Gemini-3-flash-preview & 34.46 & \textbf{72.33} & 62.50 & \textbf{65.31} & \textbf{63.53} \\
Gemini-3.1-pro-preview          & \textbf{52.70} & 60.13 & \textbf{38.18} &  &  \\
gpt-4o                          & 41.81 & 53.55 & 49.13 & 39.14 & 38.04\\
kimi-k2.5                       & 43.96 & 64.15 & 45.97 & 48.10 & 46.45 \\
kimi-k2.5(thinking mode)        & 50.34 & 59.12 & 39.60 & 49.48 & 47.16 \\
qwen3.5-397b-a17b               & 50.67 & 71.70 & 43.29 & 50.17 & 48.26  \\
qwen3.5-omni-plus               & 45.64 & 68.55 & 46.31 & 59.86 & 57.42 \\ \midrule

\multicolumn{6}{@{}l}{\textit{Open-source models}} \\ 
InternVL2\_5-8B                 & 32.89 & 39.62 & 53.69 & 35.64 & 34.58 \\
InternVL3-8B                    & 27.52 & 57.86 & 67.11 & 35.99 & 34.69  \\
InternVL3\_5-8B-Instruct        & \textbf{51.34} & 60.38 & 39.93 & 38.06 & 36.92 \\
InternVL3\_5-8B-Thinking        & 40.27 & 59.75 & 48.66 & 40.14 & 39.37  \\
MiniCPM-V-4\_5                  & 21.81 & 64.46 & 67.22 & 41.87 & 40.14 \\
Qwen2.5-VL-7B-Instruct          & 24.83 & 59.12 & 69.46 & 39.79 &  37.58 \\
Qwen3-Omni-30B-A3B-Instruct     & 45.30 & 54.09 & 43.29 & \textbf{42.91} & \textbf{40.63}\\
Qwen3-Omni-30B-A3B-Thinking     & 40.27 & 49.06 & 48.66 & 42.21 &  40.26 \\
Qwen3-VL-8B-Instruct            & 26.85 & \textbf{67.92} & 70.47 & 40.83 &  39.22 \\
Qwen3-VL-8B-Thinking            & 39.93 & 49.37 & 48.31 & 41.87 & 39.95 \\
LongVU\_Qwen2\_7B               & 36.91 & 17.61 & 32.89 & 28.72 &  28.53\\
LongVA-7B-DPO                   & 37.58 & 23.27 & \textbf{30.41} & 35.64 & 34.28\\
llava-onevision-qwen2-7b-ov-hf  & 6.04 & 61.19 & 89.54 & 34.60 &  32.70\\
LLaVA-Video-7B-Qwen2            & 7.05 & 32.69 & 90.78 & 35.29 &  34.11\\
\midrule
human performance               & 66.67 & 73.33 & 26.67 & 79.31 & 76.79 \\
\bottomrule
\end{tabularx}
\end{table}

\begin{table}[htbp]
    \centering
    \caption{Task~2.1 ablation over input conditioning for representative closed-source/API and open-source/local models.}
    \scriptsize
    \setlength{\tabcolsep}{2pt}
    \begin{tabularx}{\textwidth}{
        @{}
        l
        >{\raggedright\arraybackslash\hsize=2\hsize}X
        *{5}{>{\centering\arraybackslash\hsize=\dimexpr 4\hsize/5\relax}X}
        @{}
    }
        \toprule
        Model & Inputs
        & \multicolumn{3}{c}{Tool Selection}
        & \multicolumn{2}{c}{Request Prediction} \\
        \cmidrule(lr){3-5}
        \cmidrule(lr){6-7}
        &
        & Acc.
        & Required Recall
        & Distractor Rate
        & Acc.
        & Macro-F1 \\
        \midrule

        \multicolumn{7}{@{}l}{\textit{Closed-source/API models}} \\

        \multirow{4}{*}{kimi-k2.5}
        & video                           & 51.68 & 59.12 & 36.58 & 43.25 & 42.02 \\
        & video + gaze                    & 50.00 & 59.12 & 38.93 & 45.33 & 44.48 \\
        & video + gaze + transcript       & 43.96 & 64.15 & 45.97 & 48.10 & 46.45 \\
        & video + gaze + transcript + ICL & --    & --    & --    & 50.36 & 48.49 \\

        \midrule
        \multicolumn{7}{@{}l}{\textit{Open-source/local models}} \\

        \multirow{4}{*}{Qwen3-VL-8B-Instruct}
        & video                           & 27.18 & 57.86 & 66.11 & 30.45 & 29.61 \\
        & video + gaze                    & 27.52 & 59.75 & 66.11 & 30.10 & 29.05 \\
        & video + gaze + transcript       & 26.85 & 67.92 & 70.47 & 40.83 & 39.22 \\
        & video + gaze + transcript + ICL & --    & --    & --    & 39.78 & 38.77 \\

        \bottomrule
    \end{tabularx}
    \label{tab:append_benchmark_ablation_task2.1}
\end{table}
Overall, as shown in Table~\ref{tab:append_benchmark_results_task2.1}, closed-source/API models generally perform better than open-source/local models on Task~2.1, especially for \textit{Request Prediction}. For \textit{Tool Selection}, the best strict accuracy is achieved by Gemini-3.1-pro-preview among closed-source models and InternVL3.5-8B-Instruct among open-source models. However, high strict accuracy does not always align with high required recall: some models identify many required tools but also select many irrelevant tools, leading to a high distractor rate. This reflects the difficulty of the task, where models must not only recognize the necessary external capabilities but also avoid over-calling unrelated tools.

For \textit{Request Prediction}, Gemini-3-flash-preview achieves the best performance, substantially outperforming both open-source models and other API models. This suggests that inferring the user's intended request from egocentric context remains challenging, especially when the spoken request itself is removed from the audio. Compared with \textit{Tool Selection}, this task benefits more clearly from additional contextual signals, since the model must rely on visual context, gaze, and surrounding audio cues to infer what the user was trying to ask.

The ablation results in Table~\ref{tab:append_benchmark_ablation_task2.1} further show different effects of input conditioning across the two question types. For \textit{Tool Selection}, adding gaze and transcript does not consistently improve performance and may even reduce strict accuracy or increase distractor rate. One possible reason is that the explicit user request is already provided in the prompt, so additional contextual signals may introduce noise or encourage models to over-select tools. In contrast, \textit{Request Prediction} consistently benefits from richer context. For kimi-k2.5, performance improves from video-only to video+gaze, then further improves with transcript and in-context examples. A similar pattern appears for Qwen3-VL-8B-Instruct, where adding transcript leads to a large gain in both accuracy and macro-F1. These results indicate that gaze and surrounding audio are especially useful when the model must infer the latent request rather than reason from an explicit user query.
\section{Benchmark Task2.2 - Proactive Request Recommendations}
\subsection{Benchmark Question Definition and Curation Pipeline.}

As introduced in Section~\ref{sec:Proactive request rec}, Task2.2 evaluates proactive assistance in egocentric streams from two complementary perspectives: \textbf{when} should the assistant interrupt the user and \textbf{what} proactive interaction would be valuable and helpful if an intervention is allowed. We instantiate these perspectives as two benchmark question types.

\paragraph{Proactive Timing Judgment (PTJ).} PTJ is a binary classification task that asks whether the user is interruptible during a video segment. Each sample contains a fixation-overlaid egocentric video clip, the participant profile, and, when available, the audio transcript. The model must output exactly one label:
\[
    y \in \{\texttt{True}, \texttt{False}\},
\]
where \texttt{True} means that a proactive recommendation can be shown, and \texttt{False} means that the segment should not be interrupted.

To construct PTJ, we convert each participant's retrospective interruptibility annotation into a temporally grounded binary supervision signal. For a recording of duration $T$, participant-marked forbidden intervals are first clipped to $[0,T]$; after merging overlapping spans, the resulting non-interruptible set is denoted by $\widetilde{\mathcal{B}}$. Its complement,
\[
    \mathcal{A}=[0,T]\setminus\bigcup_{[s,e]\in\widetilde{\mathcal{B}}}[s,e],
\]
defines the candidate interruptible intervals. To keep model inputs within a comparable temporal scale, intervals from both $\widetilde{\mathcal{B}}$ and $\mathcal{A}$ are partitioned into clips of at most 300 seconds, while very short fragments below 15 seconds are discarded. Clips derived from $\widetilde{\mathcal{B}}$ receive the \texttt{False} label, while clips derived from $\mathcal{A}$ receive the \texttt{True} label. Each retained clip is then materialized as a fixation-overlaid egocentric clip and paired with the participant profile, transcript, temporal metadata, and the binary prompt. Participant-provided explanations for forbidden intervals are preserved as metadata but are not exposed as target answers. Thus, the PTJ input is an egocentric video clip, and the output is one binary interruptibility label.

\paragraph{Valuable Interaction Identification (VII).} VII is a four-choice multiple-choice task that asks which candidate recommendation should be proactively offered in a given segment. Each sample contains a fixation-overlaid video clip, participant profile, scene context, previous-segment summary, optional transcript, and four recommendation options:
\[
    \mathcal{C}_i = \{c_A, c_B, c_C, c_D\}.
\]
The model outputs a single option key $\hat{y}_i \in \{A,B,C,D\}$. The correct option is a participant-accepted recommendation; the other three options are distractors sampled from recommendations generated for the same recording segment but not accepted by the participant.

To construct VII, we start from the merged proactive-recommendation annotations, which combine the generated candidate pool with participants' retrospective acceptance decisions. For segment $k$, let $\mathcal{P}_k$ be its generated recommendation pool and $\mathcal{A}_k$ be the participant-accepted subset. Each accepted recommendation $a_i\in\mathcal{A}_k$ becomes one benchmark instance and serves as the positive option. When a participant edited the accepted recommendation, the edited text is used as the ground-truth option; Chinese edits are translated into English to keep the benchmark prompt language consistent. The negative options pool excludes all accepted IDs from the same segment,
\[
    \mathcal{N}_{k,i}=\{r\in\mathcal{P}_k \mid \operatorname{id}(r)\notin \operatorname{id}(\mathcal{A}_k)\},
\]
from which three negative options are sampled without replacement using a fixed seed. The four options are then shuffled with the same seeded random generator to avoid positional bias, and the resulting correct option key is recorded. The visual input is centered on the participant-annotated presentation window, for an accepted interval $[u_i^s,u_i^e]$, the clip interval is
\[
    [\max(u_i^s-60,0),\,u_i^e],
\]
where $[u_i^s,u_i^e]$ is first replaced by the full source segment interval when the annotation is marked as $[-1,-1]$. Each benchmark instance retains the accepted recommendation, negative option IDs, option strings, correct key, scene context, previous-segment summary, transcript field, and fixation-overlaid clip path. Thus, the VII input is an egocentric video clip plus four candidate proactive requests, and the output is one option key.

Table~\ref{tab:task22_benchmark_prompts} summarizes the benchmark prompt templates used for PTJ and VII. The prompt fields are instantiated from each sample's participant profile, transcript when available, scene context, previous-segment summary, and recommendation options. Task2.2 clips follow the shared benchmark video rendering described in Section~\ref{sec:benchmark_video_rendering}; they are generated from the original 1600$\times$1200 recording as 640$\times$480, 500k-bitrate, 1 FPS videos. Transcripts are generated for models without native audio input, using automatic speech recognition with Xunfei LFASR~\cite{iflytek_asr_llm_2026} for Chinese recordings and ElevenLabs Speech to Text~\cite{elevenlabs_speech_to_text} for English recordings.

\begin{table}[htbp]
    \centering
    \caption{Prompt templates used for Task2.2 benchmark questions.}
    \scriptsize
    \setlength{\tabcolsep}{2pt}
    \begin{tabularx}{\textwidth}{@{} l >{\raggedright\arraybackslash}X >{\raggedright\arraybackslash}X @{}}
        \toprule
        Question Type & Prompt Inputs                                                                                                                                                                                                                                                                       & Output Constraint                                                                                                                                                                                \\
        \midrule
        PTJ           & Fixation-overlaid clip; participant profile; transcript when available; interruptibility criteria covering safety-critical activity, focused work, deep social engagement, eating/drinking, appreciation or entertainment, transactions, privacy, rest, and device-boundary states. & Decide whether the user can be proactively interrupted during the segment. Output exactly one label: \texttt{True} if interruptible, or \texttt{False} if the segment should not be interrupted. \\
        VII           & Fixation-overlaid clip; participant profile; scene context; previous-segment summary; four recommendation options with intent categories.                                                                                                                                           & Select the recommendation that should be proactively offered during the segment. Output exactly one option key from \texttt{A}/\texttt{B}/\texttt{C}/\texttt{D}, with no explanation.            \\
        \bottomrule
    \end{tabularx}
    \label{tab:task22_benchmark_prompts}
\end{table}

\subsection{Benchmark Statistic.}

Table~\ref{tab:task22_stats_overview} summarizes the scale of the two Task2.2 benchmark question types. PTJ contains 269 binary samples from 269 unique fixation-overlaid clips across 15 participants. VII contains 1,161 four-choice samples from 742 unique fixation-overlaid clips across 15 participants, with one sample constructed for each participant-accepted proactive recommendation. Table~\ref{tab:task22_original_annotation_stats} reports the annotation statistics before benchmark instance construction.

\begin{table}[htbp]
    \centering
    \caption{Task2.2 benchmark statistics.}
    \scriptsize
    \setlength{\tabcolsep}{2pt}
    \begin{tabularx}{\textwidth}{@{} >{\raggedright\arraybackslash}X *{5}{Y} @{}}
        \toprule
        Question Type & \#Samples & \#Participants & \#Unique Clips & Answer Space  & Chance \\
        \midrule
        PTJ           & 269       & 15             & 269            & True / False  & 0.50   \\
        VII           & 1,161     & 15             & 742            & A / B / C / D & 0.25   \\
        \bottomrule
    \end{tabularx}
    \label{tab:task22_stats_overview}
\end{table}

\begin{table}[htbp]
    \centering
    \caption{Original annotation statistics used to construct Task2.2 benchmarks.}
    \scriptsize
    \setlength{\tabcolsep}{2pt}
    \begin{tabularx}{\textwidth}{@{} >{\raggedright\arraybackslash}X Y Y @{}}
        \toprule
        Statistic                                & Scope           & Value             \\
        \midrule
        \multicolumn{3}{@{}l}{\textit{PTJ original forbidden-segment annotations}}     \\
        Total annotated forbidden segments       & 15 participants & 74                \\
        Avg. forbidden segments / participant    & 15 participants & 4.93              \\
        Avg. forbidden-segment length            & 74 segments     & 345.2 sec.        \\
        \midrule
        \multicolumn{3}{@{}l}{\textit{VII original merged recommendation annotations}} \\
        Total annotated segments                 & 15 participants & 519               \\
        Avg. segments / participant              & 15 participants & 34.6              \\
        Avg. generated recommendations / segment & 519 segments    & 25.9              \\
        Avg. accepted recommendations / segment  & 519 segments    & 2.24              \\
        \bottomrule
    \end{tabularx}
    \label{tab:task22_original_annotation_stats}
\end{table}

PTJ has a moderately imbalanced binary label distribution (Table~\ref{tab:task22_ptj_stats}), with interruptible intervals forming the majority class. The benchmark preserves this imbalance to test whether proactive systems can avoid disturbing users during participant-marked forbidden contexts. PTJ clips range from 19.0 to 300.0 seconds (mean 240.7 seconds; median 300.0 seconds). The transcript-bearing prompt version contains non-empty transcripts for 255/269 samples (94.8\%) and is used for models without native audio understanding; native-audio models can instead consume the original clip audio directly.

\begin{table}[htbp]
    \centering
    \caption{PTJ label distribution.}
    \scriptsize
    \setlength{\tabcolsep}{2pt}
    \begin{tabularx}{\textwidth}{@{} l Y Y @{}}
        \toprule
        Label                    & \#Samples & Percent \\
        \midrule
        True (interruptible)     & 162       & 60.2\%  \\
        False (do not interrupt) & 107       & 39.8\%  \\
        \bottomrule
    \end{tabularx}
    \label{tab:task22_ptj_stats}
\end{table}

VII has a multi-class distribution over accepted recommendation types (Table~\ref{tab:task22_vii_intent_distribution}) and a separately shuffled answer-key distribution (Table~\ref{tab:task22_vii_option_distribution}). Because several participant-accepted recommendations can occur within the same source segment, the 1,161 VII samples reuse 742 unique fixation-overlaid clips; among reused clips, each clip appears between 1 and 10 times (mean 1.56 samples per clip). Participant-annotated recommendation presentation time windows range from 0.8 to 300.0 seconds (mean 115.1 seconds; median 52.8 seconds). Trimmed clips range from 60.8 to 360.0 seconds (mean 175.1 seconds; median 112.8 seconds), reflecting the additional 60-second context prepended before each accepted recommendation. The transcript version contains non-empty transcripts for 1,129/1,161 samples (97.2\%).

\begin{table}[htbp]
    \centering
    \caption{VII accepted-recommendation type distribution. Each sample corresponds to one participant-accepted proactive recommendation.}
    \scriptsize
    \setlength{\tabcolsep}{2pt}
    \begin{tabularx}{\textwidth}{@{} >{\raggedright\arraybackslash}X Y Y >{\raggedright\arraybackslash}X Y Y @{}}
        \toprule
        Type                                  & \#Samples & Percent & Type                                     & \#Samples & Percent \\
        \midrule
        Context-Aware Recommendation          & 252       & 21.7\%  & Computation and Estimation               & 108       & 9.3\%   \\
        Task Assistance                       & 185       & 15.9\%  & Decision Support                         & 106       & 9.1\%   \\
        Engaging Interaction                  & 152       & 13.1\%  & Contextual Memory                        & 88        & 7.6\%   \\
        Procedural Guidance                   & 123       & 10.6\%  & Translation, Text, and Audio Recognition & 33        & 2.8\%   \\
        Object Identification and Recognition & 114       & 9.8\%   & --                                       & --        & --      \\
        \bottomrule
    \end{tabularx}
    \label{tab:task22_vii_intent_distribution}
\end{table}

\begin{table}[htbp]
    \centering
    \caption{VII correct option-key distribution.}
    \scriptsize
    \setlength{\tabcolsep}{2pt}
    \begin{tabularx}{\textwidth}{@{} l Y Y l Y Y @{}}
        \toprule
        Correct Key & \#Samples & Percent & Correct Key & \#Samples & Percent \\
        \midrule
        A           & 280       & 24.1\%  & C           & 326       & 28.1\%  \\
        B           & 266       & 22.9\%  & D           & 289       & 24.9\%  \\
        \bottomrule
    \end{tabularx}
    \label{tab:task22_vii_option_distribution}
\end{table}

\subsection{Benchmark Results.}

For reproducibility, Table~\ref{tab:task22_inference_settings} reports the main inference settings used for Task2.2. Vision-only and transcript-dependent video models are evaluated with the transcript-bearing prompt; native audio-video models are evaluated with the paired no-transcript prompt and consume the clip audio directly. The same decoding settings are used for both question types. Most non-thinking models use a maximum of 5 new tokens; qwen3.5-omni-plus requires at least 10 new tokens. Temperature is 0.0 for all models except InternVL3.5-8B-Instruct (thinking), which requires temperature 0.6 to enable thinking mode.

\begin{table}[htbp]
    \centering
    \caption{Task2.2 inference settings. ``Frames/input'' describes the visual/audio inputs supplied per clip; native-video API models handle frame selection internally. Gemini runs use full video and audio through the Files API without an explicit generation-token cap. Kimi and Qwen API runs start from the full video and dynamically reduce FPS/resolution to fit the API byte/token limit (20MB).}
    \scriptsize
    \setlength{\tabcolsep}{3pt}
    \setlength{\tabcolsep}{2pt}
    \begin{tabularx}{\textwidth}{@{} l >{\raggedright\arraybackslash}X >{\raggedright\arraybackslash}X Y Y @{}}
        \toprule
        Models                                            & Frames/input                    & Transcript or audio & Temp. & Max new tokens \\
        \midrule
        \multicolumn{5}{@{}l}{\textit{Closed-source/API models}}                                                                           \\
        gemini-3-flash-preview; gemini-3.1-pro-preview    & native video + audio            & audio               & 0.0   & --             \\
        qwen3.5-omni-plus                                 & native video + audio            & audio               & 0.0   & 10             \\
        kimi-k2.5                                         & native video                    & transcript          & 0.0   & 5              \\
        kimi-k2.5 (thinking); qwen3.5-397b-a17b           & native video                    & transcript          & 0.0   & 8192           \\
        gpt-4o                                            & 50 frames                       & transcript          & 0.0   & 5              \\
        \midrule
        \multicolumn{5}{@{}l}{\textit{Open-source/local models}}                                                                           \\
        Qwen3-Omni-30B-A3B-Instruct                       & native video + audio            & audio               & 0.0   & 5              \\
        Qwen3-Omni-30B-A3B-Thinking                       & native video + audio            & audio               & 0.0   & 8192           \\
        MiniCPM-V-4\_5                                    & native video ($\leq$180 frames) & transcript          & 0.0   & 5              \\
        Qwen2.5-VL-7B-Instruct; Qwen3-VL-8B-Instruct      & native video                    & transcript          & 0.0   & 5              \\
        Qwen3-VL-8B-Thinking                              & native video                    & transcript          & 0.0   & 8192           \\
        LongVA-7B-DPO                                     & 360 frames                      & transcript          & 0.0   & 5              \\
        InternVL3-8B; InternVL3\_5-8B-Instruct            & 135 frames                      & transcript          & 0.0   & 5              \\
        InternVL3\_5-8B-Instruct (thinking)               & 135 frames                      & transcript          & 0.6   & 8192           \\
        llava-onevision-qwen2-7b-ov-hf; LongVU\_Qwen2\_7B & 64 frames                       & transcript          & 0.0   & 5              \\
        InternVL2\_5-8B; LLaVA-Video-7B-Qwen2             & 45 frames                       & transcript          & 0.0   & 5              \\
        \bottomrule
    \end{tabularx}
    \label{tab:task22_inference_settings}
\end{table}

\subsubsection{Baseline Results}
\begin{table}[htbp]
    \centering
    \caption{Task~2.2 benchmark results under the transcript/native-audio setting. PTJ: random chance $50.00\%$. VII: random chance $25.00\%$. Scores are percentages ($\times 100$). \textbf{Bold} marks the best score in each numeric column within the Closed-source/API and Open-source/local blocks.}
    \scriptsize
    \setlength{\tabcolsep}{3pt}
    \setlength{\tabcolsep}{2pt}
    \begin{tabularx}{\textwidth}{@{} l *{6}{Y} @{}} 
        \toprule
        Models                              & \multicolumn{4}{c}{PTJ} & \multicolumn{2}{c}{VII}                                                                     \\
        \cmidrule(lr){2-5} \cmidrule(lr){6-7}
                                            & acc                     & \textbf{macro f1}       & f1 (true)      & f1 (false)     & \textbf{acc}   & macro f1       \\
        \midrule
        \multicolumn{7}{@{}l}{\textit{Closed-source/API models}}                                                                                                    \\
        gemini-3.1-pro-preview              & 57.25                   & 57.13                   & 54.90          & 59.36          & \textbf{55.81} & \textbf{55.76} \\
        gemini-3-flash-preview              & 57.62                   & 57.41                   & 54.40          & 60.42          & 52.80          & 52.66          \\
        kimi-k2.5                           & \textbf{59.85}          & \textbf{59.78}          & \textbf{61.43} & 58.14          & 54.78          & 54.66          \\
        kimi-k2.5 (thinking)                & 54.28                   & 54.21                   & 52.51          & 55.91          & 52.37          & 52.08          \\
        qwen3.5-397b-a17b                   & 58.36                   & 58.36                   & 58.52          & 58.21          & 51.42          & 51.28          \\
        qwen3.5-omni-plus                   & 54.65                   & 53.22                   & 45.05          & \textbf{61.39} & 49.96          & 49.44          \\
        gpt-4o$^{\dagger}$                  & 37.92                   & 29.55                   & 1.22           & 57.88          & 43.84          & 44.94          \\
        \midrule
        \multicolumn{7}{@{}l}{\textit{Open-source/local models}}                                                                                                    \\
        InternVL2\_5-8B                     & 66.54                   & 59.06                   & 76.56          & 41.56          & 43.58          & 43.27          \\
        InternVL3-8B                        & 60.59                   & 56.22                   & 70.06          & 42.39          & 42.89          & 42.34          \\
        InternVL3\_5-8B-Instruct            & 53.53                   & 52.78                   & 46.81          & \textbf{58.75} & 41.69          & 41.02          \\
        InternVL3\_5-8B-Instruct (thinking) & 46.84                   & 46.70                   & 43.92          & 49.47          & 41.26          & 41.03          \\
        MiniCPM-V-4\_5                      & 60.22                   & 60.00                   & 62.98          & 57.03          & 43.32          & 43.13          \\
        Qwen2.5-VL-7B-Instruct              & 39.41                   & 28.27                   & 0.00           & 56.53          & 41.09          & 39.33          \\
        Qwen3-Omni-30B-A3B-Instruct         & \textbf{66.91}          & 57.62                   & \textbf{77.47} & 37.76          & 46.34          & 45.36          \\
        Qwen3-Omni-30B-A3B-Thinking         & 60.97                   & 59.92                   & 66.88          & 52.97          & 47.29          & 46.62          \\
        Qwen3-VL-8B-Instruct                & 65.80                   & \textbf{60.69}          & 74.86          & 46.51          & \textbf{47.63} & \textbf{47.16} \\
        Qwen3-VL-8B-Thinking                & 61.71                   & 57.57                   & 70.82          & 44.32          & 46.17          & 45.33          \\
        LongVU\_Qwen2\_7B                   & 60.22                   & 37.59                   & 75.17          & 0.00           & 37.98          & 35.97          \\
        LLaVA-Video-7B-Qwen2                & 60.97                   & 39.60                   & 75.52          & 3.67           & 40.65          & 37.69          \\
        llava-onevision-qwen2-7b-ov-hf      & 57.62                   & 56.97                   & 62.25          & 51.69          & 38.50          & 36.72          \\
        LongVA-7B-DPO                       & 60.22                   & 38.45                   & 75.06          & 1.83           & 36.09          & 34.86          \\
        \midrule
        human performance                   & 68.97                   & 68.36                   & 72.73          & 64.00          & 68.00          & 70.34          \\
        \bottomrule
    \end{tabularx}
    \label{tab:task22_results}
    {\scriptsize \raggedright $^{\dagger}$gpt-4o produced 25 invalid PTJ predictions (9.3\%) and 89 invalid VII predictions (7.7\%) due to content-safety refusals.\par}
\end{table}

PTJ is evaluated primarily with macro F1 because the binary labels are moderately imbalanced and a useful proactive assistant must recognize both interruptible and non-interruptible contexts. Among closed-source/API models, kimi-k2.5 obtains the best macro F1 ($59.78$), while qwen3.5-omni-plus achieves the highest F1 on the non-interruptible class but at the cost of lower F1 on interruptible clips. Among open-source/local models, Qwen3-VL-8B-Instruct reaches the best macro F1 ($60.69$), whereas Qwen3-Omni-30B-A3B-Instruct obtains the highest accuracy and F1 on interruptible clips. The per-class scores reveal different failure modes: LongVU, LLaVA-Video, and LongVA strongly favor the majority \texttt{True} label and miss many participant-marked forbidden contexts, while Qwen2.5-VL-7B-Instruct collapses toward \texttt{False}, producing zero F1 on interruptible clips. MiniCPM-V-4.5 is notable for a more balanced PTJ profile despite lower absolute accuracy.

VII is evaluated primarily with accuracy because the shuffled multiple-choice labels are approximately balanced. This semantic selection task requires the model to distinguish the participant-accepted proactive request from plausible alternatives generated for the same segment. Gemini-3.1-pro-preview performs best overall ($55.81$ accuracy), followed by kimi-k2.5 and gemini-3-flash-preview among closed-source/API models. The best open-source/local result is Qwen3-VL-8B-Instruct ($47.63$ accuracy), leaving a substantial gap to the strongest API models. Native audio does not uniformly improve results. Audio-capable omni models are competitive but do not consistently outperform transcript-conditioned MLLMs, suggesting that explicit transcripts can capture much of the audio information needed for this task. Thinking modes are also not consistently beneficial under the constrained answer format, likely because both PTJ and VII require short-form decisions rather than long free-form reasoning.

We additionally conducted a human prediction study to test whether the participant-provided Task2.2 labels are predictable from the same inputs given to models and can be approximated through human annotation. For each question type, we sampled a label-stratified and participant-balanced subset from the full test split, targeting around 10\% of the data with a maximum of 50 examples per task. This yielded 29 PTJ examples and 50 VII examples, evaluated using the same inputs and answer formats as the model benchmark. Because the ground truth reflects the original participant's retrospective interruption judgments and recommendation preferences, the study should not be read as an absolute human upper bound. Instead, it measures cross-person predictability from the available video, dialogue, participant profile, and candidate recommendations. Human annotators substantially outperform the strongest models on the primary metrics, reaching $68.36$ macro F1 on PTJ and $68.00$ accuracy on VII. These results support that Task2.2 labels are not arbitrary despite their subjective origin. With careful inspection of the scene, dialogue, and participant profile, humans can often infer when interruption is acceptable and which proactive request the participant would prefer. The remaining gap between human and perfect performance still reflects genuine personalization difficulty, but the human advantage over MLLMs indicates that the benchmark captures predictable user-aligned judgments rather than unknowable preferences.

\subsubsection{Error Analysis}

We analyze errors for kimi-k2.5 and Qwen3-VL-8B-Instruct, which are the strongest representative closed-source/API and open-source/local models under the transcript setting.

\begin{figure}[htbp]
    \centering
    \includegraphics[width=0.82\textwidth]{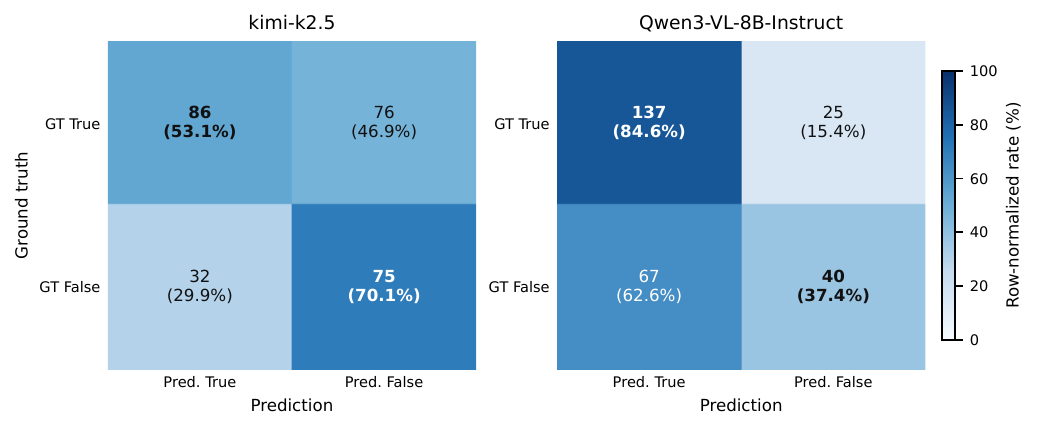}
    \caption{PTJ confusion heatmaps for the two representative models. Each cell reports count and row-normalized percentage. The \texttt{True} label denotes an interruptible segment, and \texttt{False} denotes a participant-marked non-interruptible segment.}
    \label{fig:task22_ptj_error_confusion}
\end{figure}

\paragraph{PTJ errors reflect different interruption biases.}
The two models reach similar PTJ macro F1, but make qualitatively different mistakes, as shown in Figure~\ref{fig:task22_ptj_error_confusion}. Kimi-k2.5 predicts \texttt{False} more often than \texttt{True}, with 151/269 predictions ($56.1\%$) versus 118/269 predictions ($43.9\%$), correctly identifying 75 of 107 non-interruptible segments ($70.1\%$) but missing 76 of 162 interruptible segments ($46.9\%$). This pattern suggests a conservative bias that avoids interruption in many ambiguous cases. Qwen3-VL-8B-Instruct shows the opposite tendency: it predicts \texttt{True} in 204 of 269 cases ($75.8\%$), correctly recovering 137 of 162 interruptible segments ($84.6\%$) but incorrectly interrupting on 67 of 107 non-interruptible segments ($62.6\%$). Thus, the best open-source/local model obtains higher PTJ accuracy by capturing the majority interruptible label, while kimi-k2.5 is more balanced in recognizing participant-marked forbidden contexts. The models are jointly wrong on only 49 PTJ segments ($18.2\%$), indicating that their errors are partly complementary rather than identical.

\begin{figure}[htbp]
    \centering
    \includegraphics[width=\textwidth]{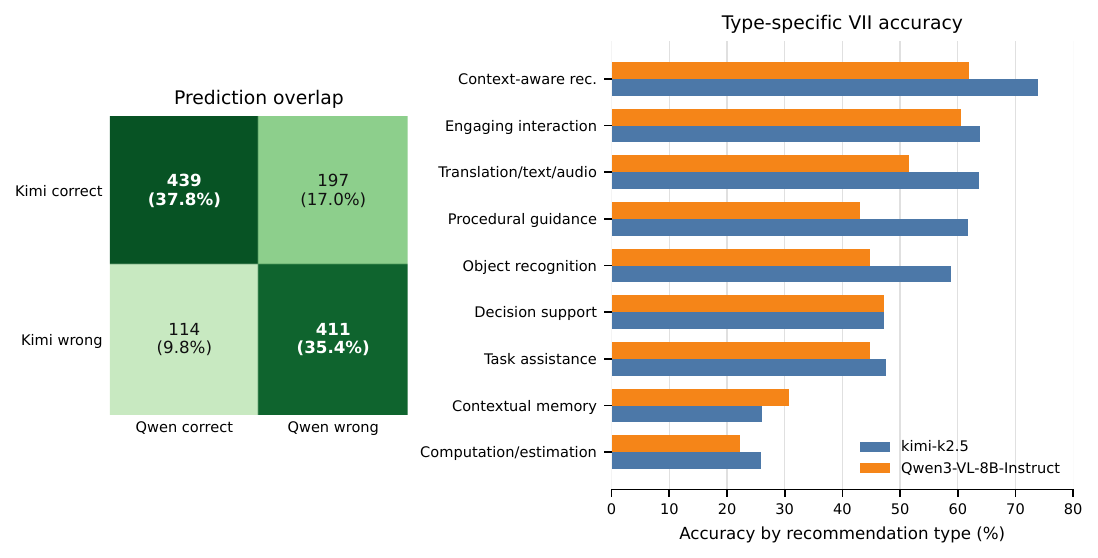}
    \caption{VII error diagnostics for kimi-k2.5 and Qwen3-VL-8B-Instruct. The left heatmap shows whether the two models succeed or fail on the same multiple-choice instances. The right panel compares accuracy across recommendation types, revealing which preference distinctions remain difficult even for the strongest representative models.}
    \label{fig:task22_vii_error_diagnostics}
\end{figure}

\paragraph{VII errors expose hard preference distinctions.}
For VII, kimi-k2.5 is more accurate than Qwen3-VL-8B-Instruct (636 vs. 553 correct predictions out of 1,161), but the two models share many difficult cases. Figure~\ref{fig:task22_vii_error_diagnostics} summarizes their prediction overlap and type-specific accuracy. They are both correct on 439 instances and both wrong on 411 instances; among the shared errors, they choose the same wrong option in 296 cases ($72.0\%$). This suggests that many VII failures arise from highly plausible distractors generated for the same segment, not only from random answer-key bias. This suggests that many VII failures arise from highly plausible distractors generated for the same segment. This pattern is expected from the construction of VII, since distractors are rejected recommendations generated for the same segment, so they are often visually and contextually plausible but fail to match the participant's preference. The shuffled ground-truth positions are approximately balanced, but Qwen3-VL-8B-Instruct over-selects option D (394/1{,}161, $33.9\%$), whereas kimi-k2.5 remains closer to the balanced key distribution. Errors also vary by recommendation type. Both models perform best on context-aware recommendation and engaging interaction examples, with $73.8\%$ and $63.8\%$ accuracy for kimi-k2.5 and $61.9\%$ and $60.5\%$ for Qwen3-VL-8B-Instruct. In contrast, computation/estimation and contextual memory are among the hardest categories, with $25.9\%$ and $26.1\%$ accuracy for kimi-k2.5 and $22.2\%$ and $30.7\%$ for Qwen3-VL-8B-Instruct. These harder cases require the model to infer a participant's specific informational need or remembered context, which is often less visually explicit than recognizing an immediately relevant scene-based suggestion.

\begin{figure}[htbp]
    \centering
    \includegraphics[width=\textwidth]{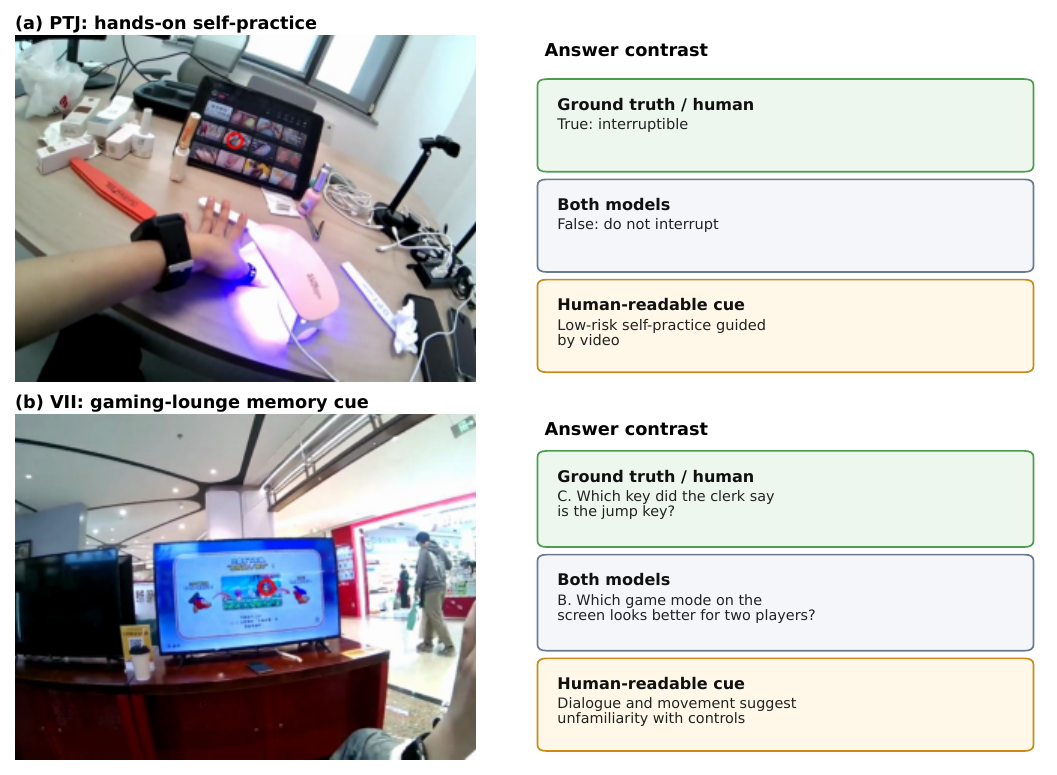}
    \caption{Representative human-correct/model-wrong Task2.2 examples. Each panel shows a rendered benchmark frame and the answer contrast between human annotators, ground truth, and the two representative models. The examples illustrate that humans can use broader activity context and participant intent to resolve cases where both models select conservative or plausible but preference-misaligned answers.}
    \label{fig:task22_human_correct_model_wrong_examples}
\end{figure}

\paragraph{Human-correct cases reveal missing contextual judgment.}
We further inspect cases in the transcript-based human prediction subset where human annotators answer correctly but both representative models fail. This occurs in 4 of 29 PTJ examples and 8 of 50 VII examples. Figure~\ref{fig:task22_human_correct_model_wrong_examples} shows one example from each question type. In the PTJ example, both models classify the segment as non-interruptible, likely because the user is engaged in a hands-on activity. However, the broader context suggests low-risk self-practice guided by video, making interruption acceptable. In the VII example, both models choose a plausible screen-level decision about game mode, but the participant-accepted answer asks which key the shopkeeper said was for jumping. Human annotators can reason from the user's dialogue and movement that the participant is still unfamiliar with the console, controls, and game, making it more useful to recall the clerk's recent instruction than to recommend a game-mode choice. The failure reflects preference alignment and contextual memory rather than scene understanding alone: the rejected option is relevant to the visible game screen, but the prior dialogue makes the remembered control instruction the more useful intervention.

\subsubsection{Ablation Study}

Table~\ref{tab:task22_ablation} isolates the contribution of the main input sources used in Task2.2. We compare plain video, fixation-overlaid video, fixation-overlaid video with transcript, and fixation-overlaid video with transcript and in-context learning (ICL). The ICL condition serves as a lightweight personalization signal rather than a generic few-shot prompt: for VII, each participant is represented by one randomly selected example of their accepted recommendation behavior; for PTJ, each participant contributes one interruptible and one non-interruptible example when both are available. These examples are removed from the evaluated ICL split, so the comparison measures whether user-specific history improves prediction on the remaining samples.

\begin{table}[htbp]
    \centering
    \caption{Task~2.2 ablation over input conditioning for representative closed-source/API and open-source/local models (same decoding as Table~\ref{tab:task22_results}). Scores are percentages ($\times 100$). ICL denotes same-participant in-context examples used as past interaction history. \textbf{Bold} marks the best score in each numeric column within each model block.}
    \scriptsize
    \setlength{\tabcolsep}{2pt}
    \begin{tabularx}{\textwidth}{@{} l >{\raggedright\arraybackslash\hsize=2\hsize}X *{6}{>{\centering\arraybackslash\hsize=\dimexpr 5\hsize/6\relax}X} @{}}
        \toprule
        Model                                 & Inputs                          & \multicolumn{4}{c}{PTJ} & \multicolumn{2}{c}{VII}                                                                                             \\
        \cmidrule(lr){3-6} \cmidrule(lr){7-8}
                                              &                                 & acc                     & \textbf{macro f1}                & f1 (true)      & f1 (false)     & \textbf{acc}                        & macro f1                   \\
        \midrule
        \multicolumn{8}{@{}l}{\textit{Closed-source/API models}}                                                                                                                                                                \\
        \multirow{4}{*}{kimi-k2.5}            & video                           & 60.22                   & 59.90                   & 63.48          & 56.33          & 47.29                      & 47.15                      \\
                                              & video + gaze                    & 57.99                   & 57.96                   & 59.21          & 56.70          & 46.25                      & 46.20                      \\
                                              & video + gaze + transcript       & 59.85                   & 59.78                   & 61.43          & 58.14          & 54.78                      & 54.66                      \\
                                              & video + gaze + transcript + ICL & \textbf{68.33}          & \textbf{67.83}          & \textbf{71.85} & \textbf{63.83} & $^{\dagger}$\textbf{54.89} & $^{\dagger}$\textbf{54.85} \\
        \midrule
        \multicolumn{8}{@{}l}{\textit{Open-source/local models}}                                                                                                                                                                \\
        \multirow{4}{*}{Qwen3-VL-8B-Instruct} & video                           & 66.54                   & 59.37                   & 76.44          & 42.31          & 42.72                      & 41.80                      \\
                                              & video + gaze                    & 61.71                   & 56.38                   & 71.63          & 41.14          & 42.29                      & 41.42                      \\
                                              & video + gaze + transcript       & 65.80                   & 60.69                   & 74.86          & \textbf{46.51} & 47.63                      & 47.16                      \\
                                              & video + gaze + transcript + ICL & \textbf{68.33}          & \textbf{61.35}          & \textbf{77.78} & 44.93          & \textbf{48.95}             & \textbf{48.10}             \\
        \bottomrule
    \end{tabularx}
    \vspace{2pt}
    {\scriptsize \raggedright $^{\dagger}$The VII score for kimi-k2.5 uses a multi-turn dialogue format for the ICL example, which performed better for this model than the standard single-turn prompt format used by the remaining ICL runs.\par}
    \label{tab:task22_ablation}
\end{table}

\paragraph{Transcripts provide the clearest contribution.}
Task2.2 benefits most consistently from linguistic context. Adding transcripts to fixation-overlaid video produces large gains for VII, increasing accuracy by \textbf{8.5 points} for kimi-k2.5 ($46.25$ to $54.78$) and \textbf{5.3 points} for Qwen3-VL-8B-Instruct ($42.29$ to $47.63$). It also improves PTJ macro F1 relative to the gaze-only setting, especially for Qwen3-VL-8B-Instruct ($56.38$ to $60.69$). These gains indicate that proactive request prediction depends strongly on audio-derived information, including the participant's speech and surrounding sounds, especially for models without native audio understanding.
 
\paragraph{Same-participant ICL improves personalization.}
Adding ICL further improves both representative models. The largest effect appears on PTJ for kimi-k2.5, where macro F1 rises by \textbf{8.1 points} ($59.78$ to $67.83$). Qwen3-VL-8B-Instruct shows smaller but consistent gains, improving from $60.69$ to $61.35$ macro F1 on PTJ and from $47.63$ to $48.95$ accuracy on VII. Because the examples come from the same participant, these gains support the value of user-specific history: past judgments help the model estimate interruption tolerance and preferred proactive assistance beyond what can be inferred from the current segment alone.

\paragraph{Fixation overlays alone do not reliably help current MLLMs.}
Although fixation cues should intuitively help models reason about the user's visual attention, the gaze-only comparison is more nuanced. For both representative models, fixation-overlaid video slightly underperforms plain video before transcripts are added: kimi-k2.5 drops from $59.90$ to $57.96$ macro F1 on PTJ and from $47.29$ to $46.25$ accuracy on VII, while Qwen3-VL-8B-Instruct drops from $59.37$ to $56.38$ macro F1 on PTJ and from $42.72$ to $42.29$ accuracy on VII. Prior work shows that red-circle markings can steer VLM attention in static grounding tasks~\cite{shtedritski2023redcircle}, and EgoGazeVQA demonstrates that gaze-guided prompting can improve MLLM performance on egocentric video questions requiring precise spatial reasoning and intent interpretation~\cite{peng2025eye}. However, EgoGazeVQA also reports that gaze cues can fail under heavy camera/body motion and that rapid gaze transitions may direct models toward irrelevant objects. These observations align with our setting, where the fixation marker is a temporally varying attention signal sampled in a few-minute egocentric clip rather than a single object-level visual prompt, and the target labels are participant-specific judgments of interruptibility or recommendation value rather than direct object localization. At the 1 FPS benchmark sampling rate, fixation cues may be too sparse or visually distracting for models to aggregate reliably, and many ground-truth decisions may depend more on speech, activity context, or personal preference than on the exact fixation point. Thus, the ablation suggests that fixation overlays alone are not sufficient for current general-purpose MLLMs to exploit gaze robustly in Task2.2. Future work may require higher-rate temporal gaze modeling, gaze-aware fine-tuning, or architectures explicitly trained to interpret fixation overlays as user-attention signals.

\section{Benchmark Task3.1 - Memory Recall Prediction}
\subsection{Benchmark Question Definition and Curation Pipeline.}
As introduced in Section~\ref{sec:Memory Reachability}, the MRP task evaluates a model's capacity to simulate human memory patterns by assessing its ability to identify details that are most likely to be clearly recalled by a human observer.

The MRP is formulated as a four-choice multiple-choice question. Each sample comprises a fixation-overlaid video clip with audio track, scene context, a transcript (or audio), and a set of four options: $\mathcal{C}_i = \{c_A, c_B, c_C, c_D\}$. The model must predict a single option key $\hat{y}_i \in \{A, B, C, D\}$. The correct option is derived from a participant-annotated vivid detail, while the distractors consist of other items or events that appear within the same video segment

The construction of MRP follows a rigorous multi-stage pipeline:
\begin{itemize}[leftmargin=10pt]
\item \textbf{Textual Refinement:} We begin with the original participant accounts. Accounts that are overly dense, excessively colloquial, or too lengthy are manually refined into a formal, concise format suitable for standardized options.
\item \textbf{Segment Sampling and Generation:} For each vivid detail, we programmatically extract a 3-minute video clip centered around the relevant timestamp. We then utilize Gemini to generate three distractors based on other elements present in the clip. To ensure linguistic consistency, the model is prompted to match the style of the original vivid detail. Furthermore, the description of the vivid detail is subtly adjusted to align with the phrasing and complexity of the generated distractors.\item \textbf{Quality Control:} Experimenters manually curate the candidate set to maintain an appropriate difficulty balance. We eliminate distractors that are either too trivial (e.g., common road signs) or excessively obscure (e.g., objects that are not visually salient). We also conduct a final verification to ensure that the semantic integrity of the correct option is preserved after LLM-based adjustment.\item \textbf{Collision Avoidance:} In cases where vivid details are temporally close and 3-minute clips inevitably overlap, we explicitly instruct the API to exclude other annotated vivid details from the distractor set. These instances are further cross-checked manually by experimenters.\item \textbf{Temporal Shuffling}:To ensure that models rely on true multimodal understanding rather than temporal heuristics, the candidate options are randomly shuffled. This ensures that the order of options does not correspond to the chronological sequence of events in the video, thereby requiring the model to engage deeply with the visual and narrative context.\end{itemize}

Table ~\ref{tab:task31_benchmark_prompts} summarizes the benchmark prompt templates used for MRP. The prompt field are instantiated from each sample's scene contex,transcripts when available and option choices.  transcripts are only provided for models cannot deal with audio.

\begin{table}[htbp]
    \centering
    \caption{Prompt templates used for MRP benchmark questions.}
    \scriptsize
    \setlength{\tabcolsep}{2pt}
    \begin{tabularx}{\textwidth}{@{} l >{\raggedright\arraybackslash}X >{\raggedright\arraybackslash}X @{}}
        \toprule
        Question Type & Prompt Inputs                                                                                                                                                                                                                                                                       & Output Constraint                                                                                                                                                                                \\
        \midrule
        MRP           & Fixation-overlaid clip; scene context; transcript when available, four detail options .                                                                                                                                           & Select the detail that is most recallable and most likely to stay vivid in memory after the event. Output exactly one option key from \texttt{A}/\texttt{B}/\texttt{C}/\texttt{D}, with no explanation.            \\
        \bottomrule
    \end{tabularx}
    \label{tab:task31_benchmark_prompts}
\end{table}

\subsection{Benchmark Statistic.}
The MRP dataset consists of 200 samples derived from 200 distinct video clips across 15 participants, where each sample corresponds to a single annotated vivid detail. The distribution of annotations per participant ranges from 3 to 23, with a mean ($\mu$) of 13.33 and a median of 12. To maintain consistency in temporal context, all video clips are standardized to a duration of 180 seconds.
\subsection{Benchmark Results.}
For reproducibility, Table~\ref{tab:task31_inference_settings} reports the main inference settings used for Task3.1. Vision-only and transcript-dependent video models are evaluated with the transcript-bearing benchmark files; native audio-video models are evaluated with the paired no-transcript files and consume the clip audio directly. The same decoding settings are used for both question types. Most non-thinking models use a maximum of 5 new tokens; qwen3.5-omni-plus requires at least 10 new tokens. Temperature is 0.0 for all models except InternVL3.5-8B-Instruct (thinking), which requires temperature 0.6 to enable thinking mode.

\begin{table}[htbp]
    \centering
    \caption{Task3.1 inference settings. ``Frames/input'' describes the visual/audio inputs supplied per clip; native-video API models handle frame selection internally. Gemini runs use full video and audio through the Files API without an explicit generation-token cap. Kimi and Qwen API runs start from the full video and dynamically reduce FPS/resolution to fit the API byte/token limit (20MB). }
    \scriptsize
    \setlength{\tabcolsep}{3pt}
    \setlength{\tabcolsep}{2pt}
    \begin{tabularx}{\textwidth}{@{} l >{\raggedright\arraybackslash}X >{\raggedright\arraybackslash}X Y Y @{}}
        \toprule
        Models                                            & Frames/input                    & Transcript or audio & Temp. & Max new tokens \\
        \midrule
        \multicolumn{5}{@{}l}{\textit{Closed-source/API models}}                                                                           \\
        gemini-3-flash-preview; gemini-3.1-pro-preview    & native video + audio            & audio               & 0.0   & --             \\
        qwen3.5-omni-plus                                 & native video + audio            & audio               & 0.0   & 10             \\
        kimi-k2.5                                         & native video                    & transcript          & 0.0   & 5              \\
        kimi-k2.5 (thinking); qwen3.5-397b-a17b           & native video                    & transcript          & 0.0   & 8192           \\
        gpt-4o                                            & 50 frames                       & transcript          & 0.0   & 5              \\
        \midrule
        \multicolumn{5}{@{}l}{\textit{Open-source/local models}}                                                                           \\
        Qwen3-Omni-30B-A3B-Instruct                       & native video + audio & audio  & 0.0   & 5              \\
        Qwen3-Omni-30B-A3B-Thinking                       & native video + audio  & audio  & 0.6   & 8192           \\
        MiniCPM-V-4\_5                                    & native video ($\leq$180 frames) & transcript & 0.0   & 32             \\
        Qwen2.5-VL-7B-Instruct                            & native video    & transcript          & 0.0   & 5              \\
        Qwen3-VL-8B-Instruct                              & native video  & transcript & 0.0   & 32             \\
        Qwen3-VL-8B-Thinking                              & native video  & transcript & 0.0   & 2048           \\
        LongVA-7B-DPO                                     & 360 frames       & transcript          & 0.0   & 5              \\
        InternVL3-8B; InternVL3\_5-8B-Instruct            & 150 frames       & transcript          & 0.0   & 5              \\
        InternVL3\_5-8B-Instruct (thinking)               & 150 frames       & transcript          & 0.6   & 2048           \\
        llava-onevision-qwen2-7b-ov-hf; LongVU\_Qwen2\_7B & 64 frames        & transcript          & 0.0   & 5              \\
        InternVL2\_5-8B                                   & 58 frames        & transcript          & 0.0   & 5              \\
        LLaVA-Video-7B-Qwen2                              & 64 frames        & transcript          & 0.0   & 5              \\
        \bottomrule
    \end{tabularx}
    \label{tab:task31_inference_settings}
\end{table}
\subsubsection{Baseline Results}
As a semantic selection task, Memory Recall Prediction (MRP) requires models to distinguish vivid details recalled by participants from plausible distractors derived from the same segments. Detailed in Table~\ref{tab:results for task3}, our experimental results show that among closed-source and API-based models, \textit{gpt-4o} achieves the best overall performance with an accuracy of $51.50\%$ and a macro F1 of $44.01$, closely followed by \textit{qwen3.5-omni-plus} ($52.00$ accuracy, $42.85$ macro F1). In the open-source and local category, \textit{Qwen3-Omni-30B-A3B-Instruct} delivers the highest accuracy at $55.50\%$, though its macro F1 ($38.27$) indicates a more skewed prediction profile. Notably, reasoning-heavy "thinking" models frequently underperform compared to their instruction-tuned counterparts; for instance, \textit{Qwen3-Omni-30B-A3B-Thinking} shows a significant drop in accuracy ($47.50\%$) relative to its base instruct version. This performance discrepancy likely stems from the intuitive nature of vivid human memory; excessive deliberation within these reasoning-heavy architectures may inadvertently obscure the subtle, salient nuances prioritized by human recall, leading to suboptimal selection.

Regarding the \textbf{human expert baseline}, as indicated in Table~\ref{tab:results for task3}, there remains a substantial performance gap between the most advanced MLLMs and human experts across almost all cognitive dimensions. Human experts achieve an accuracy of $76.67\%$ in the MRP task, significantly outperforming the best-performing model, \textit{Qwen3-Omni-30B-A3B-Instruct}, by over $21\%$.

\begin{table}[htbp]
\centering
\caption{Benchmark Results for Cognitive Memory}
\label{tab:results for task3}
\scriptsize 
\setlength{\tabcolsep}{2pt} 
\begin{tabularx}{\textwidth}{@{} l *{10}{X} @{}} 
\toprule
Models & \multicolumn{2}{c}{MRP} & \multicolumn{2}{c}{MIM-item-text} & \multicolumn{2}{c}{MIM-item-photo} & \multicolumn{4}{c}{MIM-Lifespan} \\
\cmidrule(l){2-3}\cmidrule(l){4-5}\cmidrule(l){6-7}\cmidrule(l){8-11}
 & acc & macro f1 & acc &macro f1& acc & macro f1&acc&macro f1&f1($\leq 24h$)&f1($> 24h$) \\ 
\midrule

\multicolumn{7}{@{}l}{\textit{Closed-source models}} \\ 
Gemini-3.1-pro-preview          & 50.00 & 41.06 & 47.06 & 43.03 & \textbf{64.10} & \textbf{63.21} & \textbf{85.39} & \textbf{82.34} & \textbf{75.00} & \textbf{89.68} \\ 
Gemini-3-flash-preview          & 47.00 & 38.71 & \textbf{53.85} & \textbf{51.39} & 43.14 & 38.55 & 79.76 & 74.33 & 62.50 & 86.15 \\ 
gpt-4o                          & 51.50 & \textbf{44.01} & 47.06 & 43.55 & 57.89 & 56.75 & 73.03 & 71.75 & 63.49 & 80.0 \\
kimi-k2.5                       & 47.50 & 37.03 & 49.02 & 45.53 & 51.28 & 51.65 & 78.09 & 74.89 & 62.96 & 84.04 \\
kimi-k2.5(thinking)             & 48.50 & 38.61 & 45.10 & 40.15 & 48.72 & 48.33 & 82.02 & 76.55 & 65.22 & 88.21 \\
qwen3.5-397b-a17b               & 49.00 & 40.33 & 39.22 & 34.51 & 51.28 & 49.87 & 79.21 & 74.60 & 63.36 & 85.83 \\
qwen3.5-omni-plus               & \textbf{52.00} & 42.85 & 45.10 & 38.32 & 61.54 & 57.65 & 82.02 & 78.60 & 69.90 & 87.30 \\ 
\midrule

\multicolumn{7}{@{}l}{\textit{Open-source models}} \\ 
InternVL2\_5-8B                 & 43.00 & 35.95 & 40.00 & 36.46 & 46.15 & \textbf{41.38} & 53.37 & 52.71 & 47.13 & 58.29 \\
InternVL3-8B                    & 40.50 & 34.05 & 29.41 & 28.86 & 43.59 & 38.70 & \textbf{75.84} & \textbf{71.86} & \textbf{61.26} & \textbf{84.44} \\
InternVL3\_5-8B-Instruct        & 35.50 & 28.74 & 43.14 & 38.97 & 41.03 & 38.67 & 48.31 & 48.29 & 49.45 & 47.13 \\
InternVL3\_5-8B-Thinking        & 34.50 & 28.47 & \textbf{56.86} & \textbf{52.83} & 35.90 & 35.39 & 50.56 & 55.55 & 50.00 & 51.11 \\
MiniCPM-V-4\_5                  & 50.00 & 39.44 & 31.37 & 29.40 & 15.38 & 10.26 & 69.66 & 67.06 & 57.81 & 76.32 \\
Qwen2.5-VL-7B-Instruct          & 38.00 & 29.91 & 33.33 & 28.93 & 33.33 & 33.11 & 48.31 & 48.29 & 47.14 & 49.45 \\
Qwen3-Omni-30B-A3B-Instruct     & \textbf{55.50} & 38.27 & 29.41 & 25.36 & 35.90 & 36.30 & 61.80 & 60.52 & 53.42 & 67.62 \\
Qwen3-Omni-30B-A3B-Thinking     & 47.50 & 35.99 & 29.41 & 26.78 & 41.03 & 38.19 & 75.28 & 71.06 & 60.00 & 82.11 \\
Qwen3-VL-8B-Instruct            & 47.50 & 39.49 & 47.06 & 41.99 & 38.46 & 31.63 & 47.78 & 47.75 & 46.75 & 48.75 \\
Qwen3-VL-8B-Thinking            & 43.00 & 34.23 & 37.25 & 36.11 & 31.58 & 31.15 & 52.25 & 52.23 & 51.43 & 53.04 \\
LongVU\_Qwen2\_7B               & 36.00 & 31.75 & 27.45 & 22.64 & \textbf{48.72} & 24.83 & 13.48 & 22.45 & 25.97 & 18.92 \\
LongVA-7B-DPO                   & 49.00 & \textbf{41.43} & 37.25 & 32.16 & 25.64 & 15.48 & 24.16 & 19.13 & 39.27 & 0.00 \\
llava-onevision-qwen2-7b-ov-hf  & 47.00 & 40.25 & 29.41 & 27.14 & 21.62 & 14.38 & 25.84 & 20.54 & 41.07 & 0.00 \\
LLaVA-Video-7B-Qwen2            & 42.50 & 35.37 & 25.49 & 20.87 & 23.08 & 9.57 & 42.14 & 42.12 & 43.09 & 41.14 \\
\midrule
human performance               & 76.67 & 40.11 & 60.00 & 61.11 & 75.00 & 66.67&88.89 & 87.50 & 83.33& 91.67 \\
\bottomrule
\end{tabularx}
\end{table}

\begin{table*}[t]
\centering
\caption{Ablation results for MRP. Bold indicates the best performance within each model category.}
\label{tab:ablation results for task3.1}
\scriptsize
\begin{tabularx}{\textwidth}{l l Y Y}
\toprule
\multirow{2}{*}{Model} & \multirow{2}{*}{Inputs} & \multicolumn{2}{c}{MRP} \\
 \cmidrule(lr){3-4}
& & acc & macro f1 \\
\midrule
\multicolumn{4}{l}{\textit{Closed-source/API models}} \\ 
\multirow{4}{*}{kimi-k2.5} & video  & 45.00 & 37.81 \\
& video + gaze & 43.00 & 35.33 \\
& video + gaze + transcript & 47.50 & 37.03 \\
& video + gaze + transcript + ICL & \textbf{56.22} & \textbf{45.83} \\
\midrule
\multicolumn{4}{l}{\textit{Open-source/local models}} \\ 
\multirow{4}{*}{Qwen3-VL-8B-Instruct} & video & 37.50 & 31.48  \\
& video + gaze & 35.00 & 29.16  \\
& video + gaze + transcript & 47.50 & \textbf{39.49}   \\
& video + gaze + transcript + ICL & \textbf{48.65} & 37.54 \\
\bottomrule
\end{tabularx}
\end{table*}

\subsubsection{Ablation Study}

Table \ref{tab:ablation results for task3.1} presents the ablation study results for the \textit{Memory Recall Prediction} (MRP) task, evaluating the performance of both closed-source (\textit{Kimi-k2.5}) and open-source (\textit{Qwen3-VL-8B-Instruct}) models under various input configurations. The analysis provides several critical insights into how multimodal cues and prompting strategies contribute to modeling memory reachability:

\begin{itemize}[leftmargin=15pt]
    \item \textbf{The Pivotal Role of Transcripts:} The integration of \textit{transcripts}---textual transcriptions of the video's audio track---emerges as a primary driver of performance gains. For \textit{Qwen3-VL-8B-Instruct}, adding transcripts to the video and gaze baseline yields a significant improvement, increasing Accuracy from 35.00\% to 47.50\% and Macro F1 from 29.16 to 39.49. This suggests that linguistic information provides essential semantic context that helps the model ground vivid details within the associated past events.
    
    \item \textbf{Synergy between Transcripts and ICL:} The highest overall performance is consistently achieved by integrating all modalities with \textit{In-Context Learning} (ICL). Specifically, \textit{Kimi-k2.5} reaches its peak performance (56.22\% Accuracy and 45.83 Macro F1) under the full configuration. These results suggest that while transcripts provide the essential factual grounding, ICL serves as a powerful calibration mechanism. By presenting one-shot example of a user's recall, ICL enables the model to characterize the individual's unique \textbf{memory patterns}---specifically how they prioritize certain vivid details while others succumb to \textbf{memory decay}. This allows the model to better navigate the complex mapping between multimodal cues and the subjective nature of human memory retrieval.
    
      \item \textbf{Observations on Gaze Information:} Interestingly, the isolated addition of gaze signals $g_i$ to the video-only baseline leads to a marginal performance dip for both models. For instance, \textit{Kimi-k2.5}'s accuracy decreases from 45.00\% to 43.00\% when gaze is introduced without textual context. This suggests that raw gaze duration may be a misleading indicator of memory salience; for example, a prolonged fixation on complex text may reflect high cognitive load rather than successful encoding, whereas a transient glance at a distinct object might result in a lasting memory.
    
    \item \textbf{Model Comparison:} Across all configurations, the closed-source \textit{Kimi-k2.5} generally maintains a performance lead over the open-source \textit{Qwen3-VL-8B-Instruct}, particularly in its ability to leverage one-shot examples through ICL to resolve the most challenging memory recall queries.
\end{itemize}

In summary, the results underscore that transcripts are indispensable for bridging the gap between raw egocentric video $v_i$ and high-level cognitive recall, while ICL serves as a powerful mechanism to refine the model's predictive accuracy in the MRP task.

\section{Benchmark Task 3.2 - Memory Intent Modeling}
\subsection{Benchmark Question Definition and Curation Pipeline.}
As described in section~\ref{sec:Memory Intent Modeling}, MIM evaluates the model's understanding of memory intent from two aspects: whether it can tell the item a user explicitly intends to preserve for future utility and whether it can understand the temporal relevance of information. We instantiate these perspectives as two benchmark question types.

\textbf{Memory Assistance Recognition(MAR)}. MAR is a four choice multi-choice question task that asks the model select the item the user want to help remember. Each sample comprises a fixation-overlaid video clip with audio track, scene context, a transcript (or audio), and a set of four options: $\mathcal{C}_i = \{c_A, c_B, c_C, c_D\}$. The model must predict a single option key $\hat{y}_i \in \{A, B, C, D\}$. The correct option is derived from a participant-annotated memory content, while the distractors consist of other items or events that appear within the same video segment.

The construction of MAR follows a rigorous multi-stage pipeline:
\begin{itemize}[leftmargin=10pt]
  \item \textbf{Type-Aware Segment Construction:} Because memory intent is instantiated as three sub-types---short-term memory (STM), long-term event (LTE), and long-term object (LTO)---we extract a clip whose temporal extent reflects the granularity of each type: STM uses the annotated cue interval (extended to a minimum of $12$\,s), LTE uses the full event interval, and LTO uses a $\pm 15$\,s window around the annotation timestamp. Any segment exceeding $45$\,s is re-centered on the interval midpoint and truncated to a $45$\,s window, keeping the cue moment at the center of the clip. MAR does not retain audio tracks that explicitly reveal the answer or whose memory target is itself audio; all other audio cues are removed to prevent shortcut solutions while keeping the task focused on visual and contextual reasoning.
  \item \textbf{Textual Refinement:} Raw participant accounts are frequently elliptical and rely on private context. We apply a type-specific LLM rewriting pass that resolves deictic references and normalizes phrasing: STM/LTE rewrites preserve the ``item-to-remember'' framing while expanding vague references using the annotator-provided memory reason, whereas LTO rewrites collapse the account to the target object itself, stripping the motivation for remembering it.
  \item \textbf{Segment Sampling and Generation:} Distractors are grounded in the same recording session as the target. For each question we sample three non-overlapping sub-clips from a local window around the target timestamp, enforcing a minimum temporal separation to guarantee visual diversity. Each sub-clip is then converted into one option: for text-form questions, Gemini is prompted to generate a textual description of the sub-clip conditioned on scene context (location, participants, main activities, emotion), so that the distractor text mimics the density and framing of the correct option; for image-form questions, a reference frame is extracted from the sub-clip and used directly as the option visual.
  \item \textbf{Quality Control and Collision Avoidance:} Experimenters manually curate the candidate set to remove options that are trivially identifiable or not visually salient, and verify that LLM-based rewriting preserves the semantic identity of the correct option. For image-form questions, experimenters additionally draw a bounding box around the candidate object on each reference frame, so that the visual referent of every option is unambiguous. When multiple annotated memory items fall within the same local window, their intervals are explicitly excluded from the distractor candidate region, and a greedy non-overlap placement procedure iteratively shrinks the sampling windows so that no distractor source overlaps another annotated memory item; these cases are further cross-checked manually.
  \item \textbf{Temporal Shuffling:} The four options are randomly shuffled before export, so that the position of the correct answer carries no information about its source and models cannot rely on positional or chronological heuristics.
  
\end{itemize}

Table~\ref{tab:task322_benchmark_prompts} summarizes the benchmark prompt templates used for MAR. The prompt fields are instantiated from each sample's scene context, transcripts when available, and option choices.

\begin{table}[htbp]
    \centering  
    \caption{Prompt templates used for MAR benchmark questions.}
    \scriptsize
    \setlength{\tabcolsep}{2pt}
    \begin{tabularx}{\textwidth}{@{} l >{\raggedright\arraybackslash}X >{\raggedright\arraybackslash}X @{}}
        \toprule
        Question Type & Prompt Inputs & Output Constraint \\
        \midrule
        MAR & Fixation-overlaid clip with audio; scene context; four memory-item options. & Select the item the user most likely intends to preserve for future utility. Output exactly one option key from \texttt{A}/\texttt{B}/\texttt{C}/\texttt{D}, with no explanation. \\
        \bottomrule
    \end{tabularx}
    \label{tab:task322_benchmark_prompts}
\end{table}

\textbf{Memory Lifespan Identification (MLI)}.
MLI is a binary classification task designed to evaluate a model's ability to determine the temporal relevance of a memory item. Each sample consists of a fixation-overlaid video clip (without audio) and a classification guideline. Models are required to employ Chain-of-Thought (CoT) prompting, first outputting their reasoning before concluding with a definitive label: \textit{within\_24h} or \textit{more\_than\_24h}.

The video segments for MLI are derived from user-annotated memory content through a rigorous curation process:
\begin{itemize}[leftmargin=10pt]
\item \textbf{Standardization and Truncation:} For annotated segments exceeding 15 seconds, experimenters manually verify the content. If the core memory information remains consistent throughout, the segment is truncated into a 15-second clip. Any remaining fragments shorter than 15 seconds are discarded.

\item \textbf{Information Integrity:} A primary constraint during curation is the preservation of "memory needs." If truncating a segment to 15 seconds results in severe information loss or renders the memory content unrecognizable, the original length is maintained to ensure sufficient context.

\item \textbf{Context Extension:} For segments originally shorter than 15 seconds, experimenters manually review and extend the temporal boundaries to include necessary environmental context, ensuring the visual data is interpretable for the model.

\item \textbf{Audio Redaction:} Because users performed \textit{in-situ labeling} for the MIM dataset, the audio tracks often contain verbal cues or reasoning that explicitly reveal the information's lifespan. To prevent data leakage and ensure the model relies solely on visual and contextual cues, all audio tracks are removed.
\end{itemize}
The labels for this task are derived directly from the users' original temporal annotations.

Table~\ref{tab:mli_benchmark_prompts} summarizes the benchmark prompt templates used for the Memory Lifespan Identification (MLI) task. The prompt configurations include specific classification examples designed to calibrate the model's understanding of user intent across various scenarios, such as utility tracking and knowledge acquisition. To guide the model through the complex cognitive process of judging memory utility, we implement a Chain-of-Thought (CoT) analysis protocol defined as follows:

\begin{itemize}[leftmargin=10pt]
    \item \textbf{Step 1: [WHAT]} Identify the specific content or intent the user is capturing in the egocentric video.
    \item \textbf{Step 2: [CONTEXT]} Determine the underlying motivation for capturing this information based on the established Scenario Guidelines.
    \item \textbf{Step 3: [REUSABILITY]} Assess whether the visual information retains long-term value for the user's future self or personal narrative once the immediate task concludes.
\end{itemize}

\begin{table}[htbp]
    \centering
    \caption{Prompt templates and output constraints used for the MLI benchmark questions.}
    \label{tab:mli_benchmark_prompts}
    \footnotesize
    \renewcommand{\arraystretch}{1.5} 
    \begin{tabularx}{\textwidth}{@{} l X X @{}} 
        \toprule
        \textbf{Task} & \textbf{Prompt Inputs} & \textbf{Output Constraint} \\
        \midrule
        MLI & Fixation-overlaid video clip; classification examples covering utility tracking, step-by-step execution, temporary status, journeys, knowledge acquisition, and sensory appreciation. & Provide reasoning following the three-step CoT guideline; output exactly \texttt{within\_24h} or \texttt{more\_than\_24h} on the final line. \\
        \bottomrule
    \end{tabularx}
\end{table}
    
\subsection{Benchmark Statistic.}
\textbf{Memory Assistance Recognition (MAR).} The MAR benchmark comprises 90 samples collected from  
  15 participants. Following the type-aware curation pipeline, each sample is presented in one of two  
  question forms: \textit{text} questions, where the four options are pure text strings, and           
  \textit{image} questions, where each option is additionally anchored by an extracted reference frame 
  with a human-drawn bounding box. Of the 90 samples, 51 are \textit{text} questions and 39 are
  \textit{image} questions (detailed in Table~\ref{tab:task32_MAR_labels}). To provide sufficient
  context for identifying the user's memory intent, the fixation-overlaid clips range from 119.4\,s to
  337.0\,s, centered around a mean of 234.1\,s and a median of 237.2\,s. Detailed dataset
  specifications, including participant-wise annotation density and temporal metrics, are summarized in
   Table~\ref{tab:mar_specs}.
  \begin{table}[htbp]
      \centering
      \caption{MAR question-form distribution.}
      \scriptsize
      \setlength{\tabcolsep}{2pt}
      \begin{tabularx}{\textwidth}{@{} l Y Y @{}}
          \toprule
          Question Form & \#Samples & Percent \\
          \midrule
          text  & 51 & 56.7\% \\
          image & 39 & 43.3\% \\
          \bottomrule
      \end{tabularx}
      \label{tab:task32_MAR_labels}
  \end{table}
  \begin{table}[htbp]
  \centering
  \caption{Detailed statistics for MAR video clips and annotations.}
  \label{tab:mar_specs}
  \scriptsize
  \begin{tabularx}{\columnwidth}{@{} l Y Y Y Y @{}}
  \toprule
  \textbf{Metric} & \textbf{Mean ($\mu$)} & \textbf{Median} & \textbf{Min} & \textbf{Max} \\
  \midrule
  Annotations per Participant & 6.0   & 4.0   & 1     & 17    \\
  Video Clip Duration (s)     & 234.1 & 237.2 & 119.4 & 337.0 \\
  \bottomrule
  \end{tabularx}
  \end{table}

  For reproducibility, Table~\ref{tab:task32_inference_settings} reports the main inference settings used for MAR. Prompts require the model to select a single option key from \{A,B,C,D\} with no explanation. API models use native video where supported; local vLLM video models sample the clip per each runner's \texttt{video\_fps} / frame recipe.

  \begin{table}[htbp]
      \centering
      \caption{MAR inference settings. ``Frames/input'' summarizes visual inputs; native-video API
  models stream the clip directly, while local vLLM models sample frames at the listed
  \texttt{video\_fps} or fixed count. For image-form MAR questions, option reference frames are
  appended as additional visual inputs. Token budgets are tight because MAR outputs a single letter;
  thinking variants retain a larger budget for the hidden reasoning trace that is stripped before
  choice extraction.}
      \scriptsize
      \setlength{\tabcolsep}{4pt}
      \begin{tabularx}{\textwidth}{@{} l >{\raggedright\arraybackslash}X Y Y @{}}
          \toprule
          Models                                            & Frames / video input
                                               & Temp. & Max new tokens \\
          \midrule
          \multicolumn{4}{@{}l}{\textit{Closed-source / API models}} \\
          gemini-3-flash-preview; gemini-3.1-pro-preview    & native video (Files API)
                                               & 0.0   & --   \\
          qwen3.5-omni-plus                                 & native video
                                               & 0.0   & 8192 \\
          kimi-k2.5                                         & frames mode (default); \texttt{video}
  mode available                                  & 0.0   & 8192 \\
          kimi-k2.5 (thinking)                              & frames mode (default); \texttt{video}
  mode available                                  & 0.0   & 8192 \\
          gpt-4o                                            & 32 frames (JPEG, uniform)
                                                & 0.0   & 512  \\
          \midrule
          \multicolumn{4}{@{}l}{\textit{Open-source / local models}} \\
          Qwen3-Omni-30B-A3B-Instruct                       & native video; \texttt{video\_fps}=0.25;
  audio muted                                   & 0.0   & 8192 \\
          Qwen3-Omni-30B-A3B-Thinking                       & native video; \texttt{video\_fps}=0.25;
  audio muted                                   & 0.6   & 8192 \\
          MiniCPM-V-4\_5                                    & 32 frames cap; \texttt{video\_fps}=0.25
                                               & 0.0   & 512  \\
          Qwen2.5-VL-7B-Instruct                            & native video; \texttt{video\_fps}=0.5
                                               & 0.0   & 5    \\
          Qwen3-VL-8B-Instruct                              & native video; \texttt{video\_fps}=0.5
                                               & 0.0   & 5    \\
          Qwen3-VL-8B-Thinking                              & native video; \texttt{video\_fps}=0.5
                                               & 0.0   & 8192 \\
          LongVA-7B-DPO                                     & 96 frames
                                               & 0.0   & 512  \\
          InternVL3-8B                                      & 96 frames (pre-resized $448{\times}448$)
                                               & 0.0   & 5    \\
          InternVL3\_5-8B-Instruct                          & 150 frames (pre-resized $448{\times}448$)
                                               & 0.0   & 5    \\
          InternVL3\_5-8B-Instruct (thinking)               & 150 frames (pre-resized $448{\times}448$)
                                               & 0.6   & 5    \\
          llava-onevision-qwen2-7b-ov-hf                    & 48 frames
                                               & 0.0   & 512  \\
          LongVU\_Qwen2\_7B                                 & 48 frames
                                               & 0.0   & 512  \\
          InternVL2\_5-8B                                   & 40 frames (tiled via
  \texttt{max\_dynamic\_patch})                                    & 0.0   & 5    \\
          LLaVA-Video-7B-Qwen2                              & 48 frames
                                               & 0.0   & 512  \\
          \bottomrule
    \end{tabularx}
      \label{tab:task322_inference_settings}
  \end{table}

\textbf{Memory Lifespan Identification(MLI)}. The MLI benchmark comprises 178 samples collected from 15 participants. The dataset exhibits a natural imbalance reflecting real-world memory patterns: 132 samples are labeled as \textit{more\_than\_24h}, while 46 samples are labeled as \textit{within\_24h} (detailed in Table~\ref{tab:task32_MLI_labels}).To ensure sufficient context for temporal reasoning, the video clips vary in duration, ranging from 5.6s to 31.4s. The distribution is centered around a mean of 14.79s and a median of 15.0s, aligning with our 15-second curation pipeline. Detailed dataset specifications, including participant-wise annotation density and temporal metrics, are summarized in Table~\ref{tab:mli_specs}.
\begin{table}[htbp]
    \centering
    \caption{MLI label distribution.}
    \scriptsize
    \setlength{\tabcolsep}{2pt}
    \begin{tabularx}{\textwidth}{@{} l Y Y @{}}
        \toprule
        Label                    & \#Samples & Percent \\
        \midrule
        within\_24h     & 46       & 25.8\%  \\
        more\_than\_24h & 132       & 74.2\%  \\
        \bottomrule
    \end{tabularx}
    \label{tab:task32_MLI_labels}
\end{table}
\begin{table}[htbp]
\centering
\caption{Detailed statistics for MLI video clips and annotations.}
\label{tab:mli_specs}
\scriptsize
\begin{tabularx}{\columnwidth}{@{} l Y Y Y Y @{}}
\toprule
\textbf{Metric} & \textbf{Mean ($\mu$)} & \textbf{Median} & \textbf{Min} & \textbf{Max} \\ 
\midrule
Annotations per Participant & 11.87 & 12.0 & 0   & 44   \\ 
Video Clip Duration (s)     & 14.79 & 15.0 & 5.6 & 31.4 \\
\bottomrule
\end{tabularx}
\end{table}

For reproducibility, Table~\ref{tab:task32_inference_settings} reports the main inference settings used for LFI. Prompts require chain-of-thought before the final label. API models use native video where supported; local vLLM video models sample the clip per each runner’s \texttt{video\_fps} / frame recipe.

\begin{table}[htbp]
    \centering
    \caption{MLI inference settings. ``Frames/input'' summarizes visual inputs; native-video vLLM models sample internally (e.g.\ Qwen2.5/3-VL). Gemini uses full video via the Files API without a fixed generation-token cap in the wrapper. Kimi defaults to \texttt{mode=video} (full video); \texttt{n\_frames} applies only in \texttt{frames} mode.}
    \scriptsize
    \setlength{\tabcolsep}{4pt} 
    \begin{tabularx}{\textwidth}{@{} l >{\raggedright\arraybackslash}X Y Y @{}}
        \toprule
        Models                                            & Frames / video input                         & Temp. & Max new tokens \\
        \midrule
        \multicolumn{4}{@{}l}{\textit{Closed-source / API models}} \\
        gemini-3-flash-preview; gemini-3.1-pro-preview    & native video                                 & 0.0   & -- \\
        qwen3.5-omni-plus                  & native video           & 0.0   & 4096 ($\max(\cdot,10)$ in code) \\
        kimi-k2.5                                         & native video           & 0.0   & 8192 \\
        kimi-k2.5 (thinking)            & native video                                 & 0.0   & 8192 \\
        gpt-4o                                    & 32 frames (JPEG)                             & 0.0   & 4096 \\
        \midrule
        \multicolumn{4}{@{}l}{\textit{Open-source / local models}} \\
        Qwen3-Omni-30B-A3B-Instruct                       & native video                & 0.0   & 4096 \\
        Qwen3-Omni-30B-A3B-Thinking                       & native video                        & 0.6   & 8192 \\
        MiniCPM-V-4\_5                                    & native video; up to 180 frames / slot, 3 slots; \texttt{video\_fps}=1 & 0.0   & 4096 \\
        Qwen2.5-VL-7B-Instruct                            & native video; \texttt{video\_fps}=1.0        & 0.0   & 4096 \\
        Qwen3-VL-8B-Instruct                              & native video; \texttt{video\_fps} default    & 0.0   & 4096 \\
        Qwen3-VL-8B-Thinking                              & native video; \texttt{video\_fps} default    & 0.0   & 8192 \\
        LongVA-7B-DPO                                     & 360 frames                                   & 0.0   & 4096 \\
        InternVL3-8B                                      & 150 frames                                   & 0.0   & 4096 \\
        InternVL3\_5-8B-Instruct                          & 96 frames                & 0.0   & 4096 \\
        InternVL3\_5-8B-Instruct (thinking)               & 96 frames               & 0.6   & 8192 \\
        llava-onevision-qwen2-7b-ov-hf                    & 64 frames                                    & 0.0   & 4096 \\
        LongVU\_Qwen2\_7B                                 & 64 frames; early stop on final label line   & 0.0   & 2048\\
        InternVL2\_5-8B                                   & 58 frames                                    & 0.0   & 4096 \\
        LLaVA-Video-7B-Qwen2                              & 64 frames                                    & 0.0   & 4096 \\
        \bottomrule
    \end{tabularx}
    \label{tab:task32_inference_settings}
\end{table}

\subsection{Benchmark Results.}
\subsubsection{Baseline Results}
MAR is evaluated with accuracy and macro F1 across two question forms---MIM-item-text (four purely textual options, primarily from short-term cues) and MIM-item-photo (four bbox-anchored reference-frame options, primarily for long-term object recognition)---because the two forms probe distinct capabilities: verbal/event-level inference for text questions and fine-grained visual grounding for image questions. Among closed-source/API models, Gemini-3-flash-preview obtains the best text-form macro F1 ($51.39$) and the best text-form accuracy ($53.85$), while Gemini-3.1-pro-preview leads on the image form with the best macro F1 ($63.21$) and accuracy ($64.10$); the two Gemini variants thus trade off between the two MAR forms, suggesting that text-form MAR benefits from faster decoding and image-form MAR benefits from stronger visual grounding rather than from a single dominant model. Among open-source/local models, InternVL3\_5-8B-Thinking reaches the best text-form macro F1 ($52.83$) and accuracy ($56.86$)---in fact surpassing every closed-source entry on the text form---whereas InternVL2\_5-8B attains the best image-form macro F1 ($41.38$) and accuracy ($46.15$). The per-form scores also reveal a systematic weakness on image-form MAR for several local video models: MiniCPM-V-4\_5 ($10.26$), LLaVA-Video-7B-Qwen2 ($9.57$), LongVA-7B-DPO ($15.48$), and llava-onevision-qwen2-7b-ov-hf ($14.38$) all fall well below the $25\%$ random-choice level in macro F1, indicating that appending bbox-anchored option frames as extra visual inputs is not natively supported by these architectures. 

Despite Gemini-3.1-pro-preview reaching $63.21$ Macro F1 on image-form MAR and Gemini-3-flash-preview reaching $51.39$ on text-form, gaps of roughly $3.5$ and $9.7$ points to human performance ($66.67$ / $61.11$) remain, widening further in accuracy ($10.9$ points on image-form, $6.2$ points on text-form). The difficulty concentrates in cases where the item worth remembering is identified not by visual prominence or verbal explicitness alone but by the user's momentary intent---encoded in subtle gaze dwell, object-proximity cues, and task context---signals that current VLMs cannot reliably infer from a short egocentric clip with a fixation overlay. For all results, see Table~\ref{tab:results for task3}.

MLI is evaluated primarily with macro F1 because the binary labels are moderately imbalanced and a useful memory assistant must recognize both short-term information and long-term ones. Among closed-source/API models, Gemini-3.1-pro-preview obtains the best macro F1 ($82.34$) and the best acc($85.39$). Among open-source/local models, InternVL3-8B reaches the best macro F1 ($71.86$)and the best acc($75.84$). The per-class scores reveal that LongVU, LLaVA-onevision, and LongVA strongly favor the \texttt{within\_24h} label. For all results, see Table ~\ref{tab:results for task3}

The evaluation of the human expert baseline on MLI reveals a profound gap between human cognitive precision and current MLLM capabilities. As shown in Table~\ref{tab:results for task3}, human experts achieve an exceptional \textbf{accuracy of 88.89\%} and a \textbf{macro F1-score of 87.50\%}, demonstrating a robust ability to perform long-term temporal grounding. A more granular analysis of the F1-scores indicates that human experts maintain high reliability across different time scales, scoring \textbf{83.33\% for short-term recall ($\leq 24h$)} and an impressive \textbf{91.67\% for long-term recall ($> 24h$)}. In stark contrast, even the top-performing closed-source models, such as \textit{Gemini-3.1-pro-preview}, exhibit a significant performance drop, particularly in short-term precision. 
\subsubsection{Ablation Study}
Table~\ref{tab:ablation_results_for_task3_2_2} presents the ablation study results for the \textit{Memory Assistance Recognition} (MAR) task, evaluating the performance of both closed-source (\textit{Kimi-k2.5}) and open-source (\textit{Qwen3-VL-8B-Instruct}) models under various input configurations. Because MAR contains two question forms with distinct input structures---\textit{text-form} questions (four textual candidates, primarily from short-term cues) and \textit{image-form} questions (four bbox-anchored reference-frame candidates, for long-term object memory)---we report each form separately. The analysis provides several critical insights into how multimodal cues and prompting strategies contribute to modeling memory-assistance intent:

\begin{itemize}[leftmargin=15pt]
    \item \textbf{The Pivotal Role of ICL on Image-form Questions:} The integration of \textit{In-Context Learning} (ICL)---where a single demonstration question is injected as a textual few-shot example, with one Gemini-2.5-Pro-generated sentence per bbox-anchored option---emerges as a primary driver of performance gains on image-form questions. For \textit{Kimi-k2.5}, adding ICL to the video and gaze baseline yields a significant improvement, increasing Accuracy from 29.03\% to 38.71\% and Macro F1 from 18.61 to 28.75. This suggests that a grounded textual description of each candidate object provides the semantic anchor the model needs to disambiguate visually similar distractors within the same scene.
    \item \textbf{Observations on Gaze Information:} The isolated addition of gaze signals $g_i$ to the video-only baseline has a form-dependent effect. On text-form questions, gaze consistently hurts both models: \textit{Kimi-k2.5}'s accuracy drops from 52.50\% to 47.50\% (Macro F1 48.08 $\rightarrow$ 42.02) and \textit{Qwen3-VL-8B-Instruct}'s from 40.00\% to 35.00\%. On image-form questions, in contrast, gaze is a genuine asset, raising \textit{Qwen3-VL-8B-Instruct}'s accuracy from 29.03\% to 32.26\% (Macro F1 25.83 $\rightarrow$ 28.61). This asymmetry suggests that raw gaze duration is informative when the task reduces to spatial localisation of an intended object, but acts as a visual distractor when the decision has to be made by linguistic reasoning over short-term cues.
    \item \textbf{Mixed Behaviour of ICL on Text-form Questions:} On text-form questions, the benefit of ICL is no longer uniform. \textit{Qwen3-VL-8B-Instruct} recovers to its video-only baseline under the full configuration (40.00\% Accuracy, 37.67 Macro F1), while \textit{Kimi-k2.5}'s accuracy keeps rising to 55.00\% but its Macro F1 collapses from 42.02 to 24.93, indicating that the model's predictions concentrate on a single dominant option letter rather than calibrating to the participant's memory pattern. This asymmetry suggests that when short-term memory cues are linguistically heterogeneous across a participant's items, a single demonstration can encourage shortcut mimicry instead of genuine generalisation, particularly in higher-capacity API models.
    \item \textbf{Model Comparison:} Across all configurations, the closed-source \textit{Kimi-k2.5} maintains a clear lead on text-form questions, peaking at 55.00\% Accuracy against the 40.00\% Accuracy of \textit{Qwen3-VL-8B-Instruct}. On image-form questions the two models are much closer once ICL is introduced (Macro F1 28.75 vs.\ 29.09), indicating that the fine-grained visual grounding required for image-form MAR is not a capability that scales trivially with model size and benefits comparably from an additional textual grounding cue.
\end{itemize}

In summary, the results underscore that ICL is indispensable for bridging the gap between raw egocentric video $v_i$ and object-level memory-assistance intent on image-form questions, while gaze signals offer selective value only when the underlying task is spatial rather than linguistic.

\newcolumntype{Y}{>{\centering\arraybackslash}X}

\begin{table*}[t]
\centering
\caption{Ablation results for MAR. All three input settings are evaluated on the same 71-question ICL-comparable subset ($40$ text-form + $31$ image-form); every model produces a valid answer on every question. Bold indicates the best performance within each model category.}
\label{tab:ablation_results_for_task3_2_2}
\scriptsize
\begin{tabularx}{\textwidth}{l l Y Y Y Y Y Y}
\toprule
\multirow{2}{*}{\textbf{Model}} & \multirow{2}{*}{\textbf{Inputs}} & \multicolumn{2}{c}{\textbf{Overall}} & \multicolumn{2}{c}{\textbf{Text-form}} & \multicolumn{2}{c}{\textbf{Image-form}} \\
\cmidrule(lr){3-4}\cmidrule(lr){5-6}\cmidrule(lr){7-8}
& & Acc & Macro F1 & Acc & Macro F1 & Acc & Macro F1 \\
\midrule
\multicolumn{8}{l}{\textit{Closed-source/API models}} \\
\multirow{3}{*}{Kimi-k2.5}
    & Video              & 40.85 & \textbf{37.21} & 52.50 & 48.08 & 25.81 & 19.68 \\
    & Video + Gaze       & 39.44 & 33.44 & 47.50 & 42.02 & 29.03 & 18.61 \\
    & Video + Gaze + ICL & \textbf{47.89} & 30.40 & \textbf{55.00} & 24.93 & \textbf{38.71} & \textbf{28.75} \\
\midrule
\multicolumn{8}{l}{\textit{Open-source/local models}} \\
\multirow{3}{*}{Qwen3-VL-8B-Instruct}
    & Video              & 35.21 & 33.30 & \textbf{40.00} & \textbf{38.36} & 29.03 & 25.83 \\
    & Video + Gaze       & 33.80 & 30.94 & 35.00 & 32.60 & 32.26 & 28.61 \\
    & Video + Gaze + ICL & \textbf{36.62} & \textbf{33.83} & \textbf{40.00} & 37.67 & \textbf{32.26} & \textbf{29.09} \\
\bottomrule
\end{tabularx}
\end{table*}

Table \ref{tab:ablation_results_for_task3_2_3} presents the performance of models in predicting the temporal utility of memory items, specifically distinguishing between short-term ($\mathcal{S}$, within 24 hours) and long-term ($\mathcal{L}$, beyond 24 hours) retention needs. The results reveal distinct behaviors in how closed-source and open-source models process physiological and visual cues for lifespan prediction:

\begin{itemize}[leftmargin=15pt]
    \item \textbf{Effective Multi-modal Integration in Kimi-k2.5:} For the \textit{Kimi-k2.5} model, the inclusion of gaze signals $g_i$ alongside video $v_i$ leads to a consistent performance improvement across all metrics. The Accuracy increases from 76.97\% to 78.09\%, with a notable gain in the Macro F1 score (from 71.99 to 74.89). This enhancement is particularly pronounced in the $\mathcal{S}$ class (F1 score rising from 60.19 to 65.49), suggesting that fine-grained physiological signals such as gaze assist the model in identifying the immediate, transient nature of certain user intents.
    
    \item \textbf{Performance Degradation in Open-source Models:} In contrast, \textit{Qwen3-VL-8B-Instruct} exhibits a significant performance drop when gaze information is introduced, with Accuracy falling from 59.55\% to 46.07\%. This discrepancy indicates that while open-source models possess basic visual recognition capabilities, they struggle to effectively align and interpret raw physiological signals for high-level cognitive tasks like lifespan identification, potentially due to the lack of specialized training on synchronized egocentric-physiological datasets.
    
    \item \textbf{Class-specific Bias and Information Importance:} Across both models, performance on long-term memory items ($\mathcal{L}$) is generally superior to that on short-term items ($\mathcal{S}$). This suggests that information requiring long-term retention---such as \textit{semantic knowledge} or \textit{meaningful experiences} that users intend to preserve---may possess more distinct contextual signatures than the ephemeral, state-based information in the $\mathcal{S}$ category. However, transient information often carries higher \textit{immediate utility} and mission-critical importance in daily life (e.g., the temporary location of an object). The observed performance gap underscores the need for more targeted training and refined feature extraction to better capture the subtle cues associated with short-term memory needs.
\end{itemize}

In conclusion, the results demonstrate that while predicting memory lifespan is a challenging cognitive task, the integration of gaze signals can provide valuable cues for identifying information utility, provided the model has sufficient capacity to fuse these heterogeneous modalities.
\newcolumntype{Y}{>{\centering\arraybackslash}X}

\begin{table*}[t]
\centering
\caption{Ablation results for MLI. Bold indicates the best performance within each model category.}
\label{tab:ablation_results_for_task3_2_3}
\scriptsize
\begin{tabularx}{\textwidth}{l l Y Y Y Y}
\toprule
\multirow{2}{*}{\textbf{Model}} & \multirow{2}{*}{\textbf{Inputs}} & \multicolumn{4}{c}{\textbf{MLI}} \\
\cmidrule(lr){3-6}
& & Acc & Macro F1 & F1($\leq 24h$) & F1($> 24h$) \\
\midrule
\multicolumn{6}{l}{\textit{Closed-source/API models}} \\
\multirow{2}{*}{Kimi-k2.5} 
    & Video        & 76.97 & 71.99 & 60.19 & 83.79 \\
    & Video + Gaze & \textbf{78.09} & \textbf{74.89} & \textbf{65.49} & \textbf{84.30} \\
\midrule
\multicolumn{6}{l}{\textit{Open-source/local models}} \\
\multirow{2}{*}{Qwen3-VL-8B-Instruct} 
    & Video        & \textbf{59.55} & \textbf{58.67} & \textbf{52.63} & \textbf{64.71} \\
    & Video + Gaze & 46.07 & 45.96 & 48.39 & 43.53 \\
\bottomrule
\end{tabularx}
\end{table*}
\section{Limitation}
\label{}
Despite the heavy investments, EgoIntrospect has several limitations. Although our dataset captures richer human-centered signals than prior egocentric benchmarks, the scale of participant diversity remains limited compared to internet-scale datasets. Compared to prior pioneer works such as Ego4D, the demographic diversity of our dataset remains relatively limited due to the substantial cost and logistical challenges associated with large-scale cross-national multisensory data collection and annotation.

On the modeling side, while we provide comprehensive analyses of existing models and explore simple strategies such as in-context calibration of individual differences, we do not propose a dedicated modeling framework specifically designed for human-centered internal-state understanding. Yet, we hope our work can motivate future research on architectures that better model subjective user states, long-term personalization, and individual behavioral patterns.



\end{document}